\RequirePackage{fix-cm}

\documentclass{svjour3}

\smartqed

\usepackage{url}
\usepackage{graphicx}
\usepackage{epstopdf}
\usepackage{booktabs}
\usepackage{commath}
\usepackage{subcaption}
\journalname{Medical \& Biological Engineering \& Computing}

\begin{document}

\title{A Fast Segmentation-free Fully Automated Approach to White Matter Injury Detection in Preterm Infants
}

\titlerunning{Segmentation-free WMI Detection in Preterms}

\author{Subhayan Mukherjee         \and
        Irene Cheng                \and
        Steven Miller              \and
        Jessie Guo                 \and
        Vann Chau                  \and
        Anup Basu
}

\institute{Subhayan Mukherjee, Irene Cheng \at
              Department of Computing Science, University of Alberta, Edmonton, Alberta, Canada
           \and
           Steven Miller, Jessie Guo, Vann Chau \at
              The Hospital for Sick Children and the University of Toronto, Toronto, Ontario, Canada
           \and
		   Anup Basu \at
              402 Athabasca Hall, Department of Computing Science, University of Alberta, Edmonton, Alberta, Canada T6G 2H1 \\
              Tel.: +1-780-492-2285\\
              Fax:  +1-780-492-1071\\
              \email{basu@ualberta.ca}
           \and
           Total number of words of the manuscript, including entire text from title page to figure legends: 9740\at
           Number of words of the abstract: 160
           \and
           Number of figures: 8\at
           Number of tables: 3
}

\date{Received: date / Accepted: date}

\maketitle

\begin{abstract}
White Matter Injury (WMI) is the most prevalent brain injury in the preterm neonate leading to developmental deficits. However, detecting WMI in Magnetic Resonance (MR) images of preterm neonate brains using traditional WM segmentation-based methods is difficult mainly due to lack of reliable preterm neonate brain atlases to guide segmentation. Hence, we propose a segmentation-free, fast, unsupervised, atlas-free WMI detection method. We detect the ventricles as blobs using a fast linear Maximally Stable Extremal Regions algorithm. A reference contour equidistant from the blobs and the brain-background boundary is used to identify tissue adjacent to the blobs. Assuming normal distribution of the gray-value intensity of this tissue, the outlier intensities in the entire brain region are identified as potential WMI candidates. Thereafter, false positives are discriminated using appropriate heuristics. Experiments using an expert-annotated dataset show that the proposed method runs 20 times faster than our earlier work which relied on time-consuming segmentation of the WM region, without compromising WMI detection accuracy.

\keywords{White Matter Injury \and segmentation \and Magnetic Resonance Imaging \and preterm newborn \and atlas-free}
\end{abstract}

\section{Introduction}
\label{intro}

Brain tissue segmentation generally refers to the separation of the brain into three functional components; namely, Grey Matter (GM), White Matter (WM) and Cerebro-Spinal Fluid (CSF). Segmentation is often performed as the first step in detection of physiological abnormalities, in volumetric study and diagnostic analysis \cite{Devi2015}. Seizures, strokes, brain infections and injuries are often hard to determine by manual examination of image scans, since significant features may not be obvious on a 2D DICOM slice. Also, manual examination can be subjective; and interpretation may vary from one expert to another. Even the same expert may make different decisions at different times. Thus, the commonly adopted gold standard in the clinical community is to obtain at least two consistent interpretations out of three expert opinions. However, getting datasets of preterm neonates with ground truth annotation by multiple clinical experts is not feasible in practice.

To address this issue, Computer Aided (or Assisted) Detection (CAD) has gained increasing attention in clinical research. For example, semi-automated approaches, like the Clusterize algorithm \cite{Clas201226}, have been applied to identify brain lesions in stroke victims and have been shown to significantly speed up lesion demarcation without loss of precision and reproducibility \cite{deHaan201569}. Computer programs designed for CAD \cite{Castellino2005} aim to detect potential abnormalities, such as Brain Tumors and Multiple Sclerosis brain lesions \cite{Bilello2013} by identifying suspected features on the image for further inspection by a radiologist. The benefit of CAD is two-fold: (1) When there is a backlog, e.g., in looking through X-ray films, due to an insufficient number of radiologists, CAD is used to pre-screen and thus reduce the workload of radiologists; and (2) when a large sequence of image scans need to be compared or examined as an integrated volume, CAD helps to detect patterns that are difficult if not impossible for human eyes to comprehend. CAD software can pin-point the areas of concern and then the radiologists can focus on these identified regions in order to arrive at the diagnosis.

Automated feature extraction and segmentation in adult brain images have been extensively studied over the past years \cite{Mekhmoukh2015,Roura2014,Ji2012}. However, the same is not true for neonates, whether preterm or term, because of several practical challenges in obtaining and analyzing MR images of neonate brains, including:
\begin{enumerate}
\item Lack of a reliable anatomical map, or ``atlas'' for an neonate brain, to guide the segmentation process in areas of low contrast and help distinguish tissues of similar intensities. Even when such atlases are available, they need to be registered onto the test MR image, which is a difficult process, because the neonate brain undergoes rapid structural and physiological changes during maturation.
\item Neonates tend to move during the MRI scan process, which is highly sensitive to patient movements. Motion artifacts, blurring, etc. degrade the quality of the MR images.
\item Neonate brains are small in size and the duration for which a neonate is scanned is also shorter than adults. This results in a low Contrast-to-Noise ratio (CNR), low Signal-to-Noise ratio (SNR) and low spatial resolution.
\item Contrast between grey matter (GM) and white matter (WM) in both T1- and T2-weighted images (T1w and T2w) is different from the adult brain. Most parts of the neonate brain are non-myelinated at birth, where WM appears less intense in T1 images and more intense in T2 images, whereas this trend is reversed for a fully myelinated adult brain (contrast inversion).
\end{enumerate}
Thus, a WMI detection method for pre-term infants should not be overly dependent on an atlas. It should be able to handle noisy, low-resolution images with motion artefacts, and it should be automated as much as possible (least human intervention), so that the process can integrate well with the CAD pipeline.

In preterm neonates, the characteristic of brain injury is multi-focal WMI in the first weeks after birth, where using T1 weighted (T1w) MR images of the neonate's brain for detection is more effective \cite{Chau2009,Miller2005}. Hence, in this work we adopt the atlas-free approach to analyze T1 weighted (T1w) MR images. We anticipate that this method will also be relevant to detecting WMI in elderly patients with leukoariosis \cite{Conklin2014}.

\subsection{Our Contribution}

Our most important advancement with respect to related work and our earlier pre-term WMI detection algorithm \cite{Cheng2015} (both described in the next section) is to eliminate the dependence of the WMI detection step on a pre-segmented white matter region. As explained earlier, this gives primarily two important benefits:

\begin{enumerate}
\item The execution time for WMI detection decreases drastically, since we do not perform the time-consuming segmentation of the white matter region. This is the main contribution of our present work from the computational perspective.
\item Since we bypass the segmentation step, no brain atlas is required. For pre-term infants, it is difficult to obtain reliable brain atlases.
\end{enumerate}

It should be noted that our proposed method is different from existing atlas-free segmentation methods in literature which use local contrast and geometric traits, brain morphology and tissue connectivity to guide segmentation. The strength of atlas-free methods is that they can accommodate changes in anatomy of the developing (neonate) brain, as they are not bound by constraints imposed by the atlas. However, their main weakness is the computational complexity of the segmentation process itself. We bypass segmentation to overcome this weakness.

\subsection{Differences with Related Work}

As described later, one of the steps in our proposed method involves the localization of ventricles as a collection of blobs detected using the Maximally Stable Extremal Regions (MSER) algorithm \cite{MSERoriginal} and optimized using Genetic Algorithms (GAs) \cite{Goldberg:1989:GAS:534133}. MSER or its modified forms have been used earlier to detect various retinopathy pathologies \cite{6460565}, segment ultrasound liver images \cite{Zhu2015}, localize cell nuclei in microscopic images \cite{Song2013}, isolate fetal brain tissues from maternal anatomy during fetal brain in-utero MR imaging \cite{Keraudren2014} and for 3D segmentation of simulated brain MR images \cite{1698834}. However, they have not been tested in preterm brain WMI detection. The main advantage of the MSER algorithm is that there is no need to specify an initial contour, which is necessary and often drawn manually in other algorithms. For example, brain tissue segmentation approaches based on Active Contour Models \cite{Qian2013,Li2005,Moreno2014,Sachdeva2012,6740480} require an initial contour. Furthermore, the region stability of MSER is constrained by local information obtained in the neighbourhood and can accommodate large intra-image variations \cite{Song2013}.

Medical images often have poor image contrast and are associated with artifacts that result in missing or diffuse organ/tissue boundaries. The resulting search space is therefore often noisy with a multitude of local optima. Genetic Algorithms (GAs) benefit medical image segmentation \cite{4798001} as they are less prone to get stuck in a local optima. GAs have been used in a learning-based approach to segment and label numerous neuroanatomic structures including left/right and third ventricles \cite{511748}. Their approach was based on observer-defined contours of neuroanatomic structures, which were used as a priori knowledge. However, in the context of preterm neonate brain WMI detection, it is not possible to obtain sufficient number of expert-annotated training images (for learning or validation). A variant of GAs called parallel genetic algorithms have been used earlier for volumetric segmentation of lateral ventricles \cite{YongFan2002} on simulated Brainweb images, but not on preterm brain images. Their strategy for choosing the initial population involves deriving an initial surface by segmenting the ventricle slice-by-slice (using a 2D method), and then solving an evolution equation (formulated using that initial surface) using a finite-difference method, whose result is used to generate the initial population for the GA. In contrast, our initial population for the GA includes all regions detected by MSER on individual (2D) slices.

\begin{figure}
\begin{subfigure}[b]{.3\linewidth}
\includegraphics[scale=0.3]{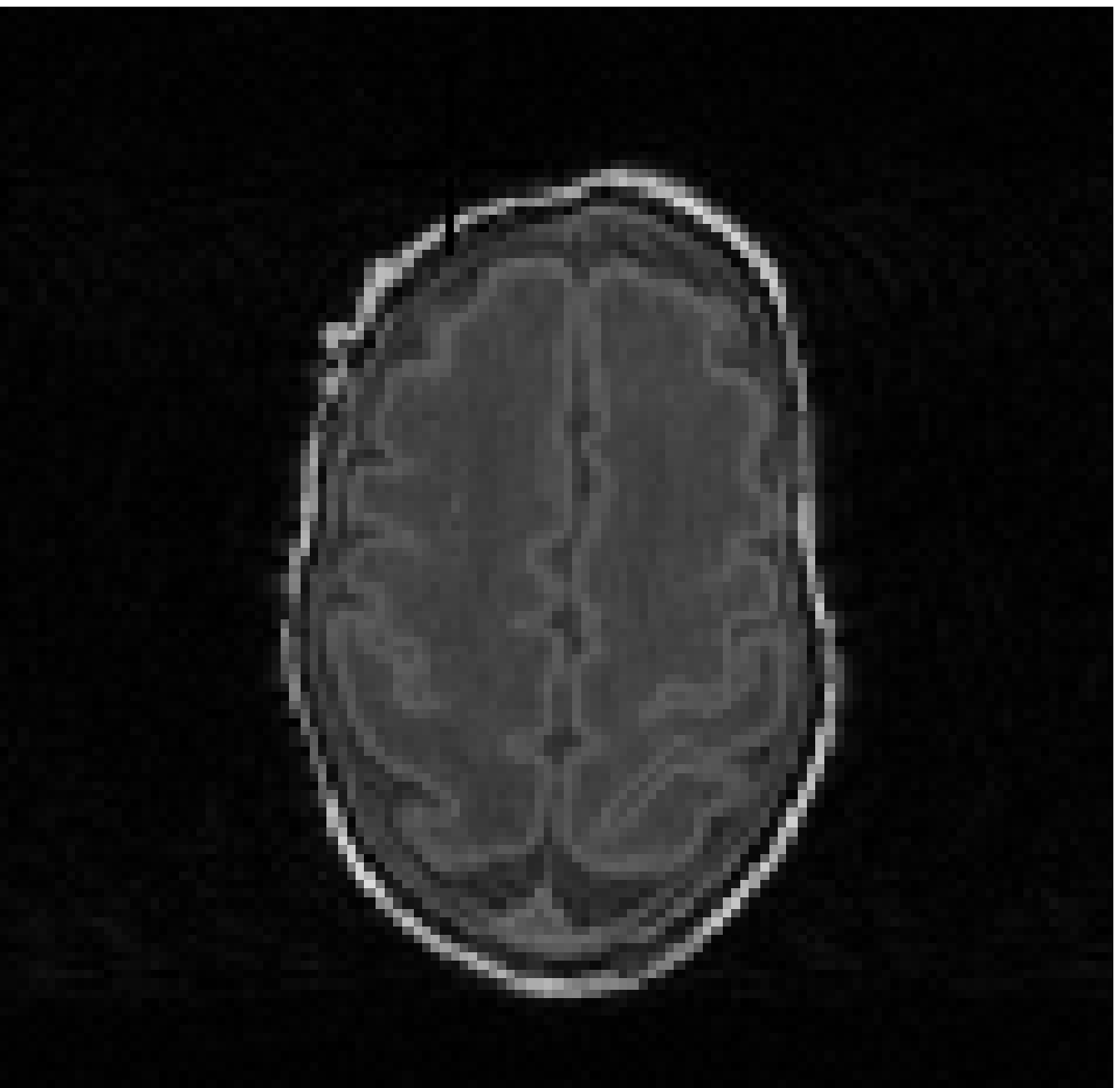}
\caption{An example of a low resolution 2D slice from our preterm neonate dataset.}\label{fig:low-slice}
\end{subfigure}
\begin{subfigure}[b]{.3\linewidth}
\includegraphics[scale=0.5]{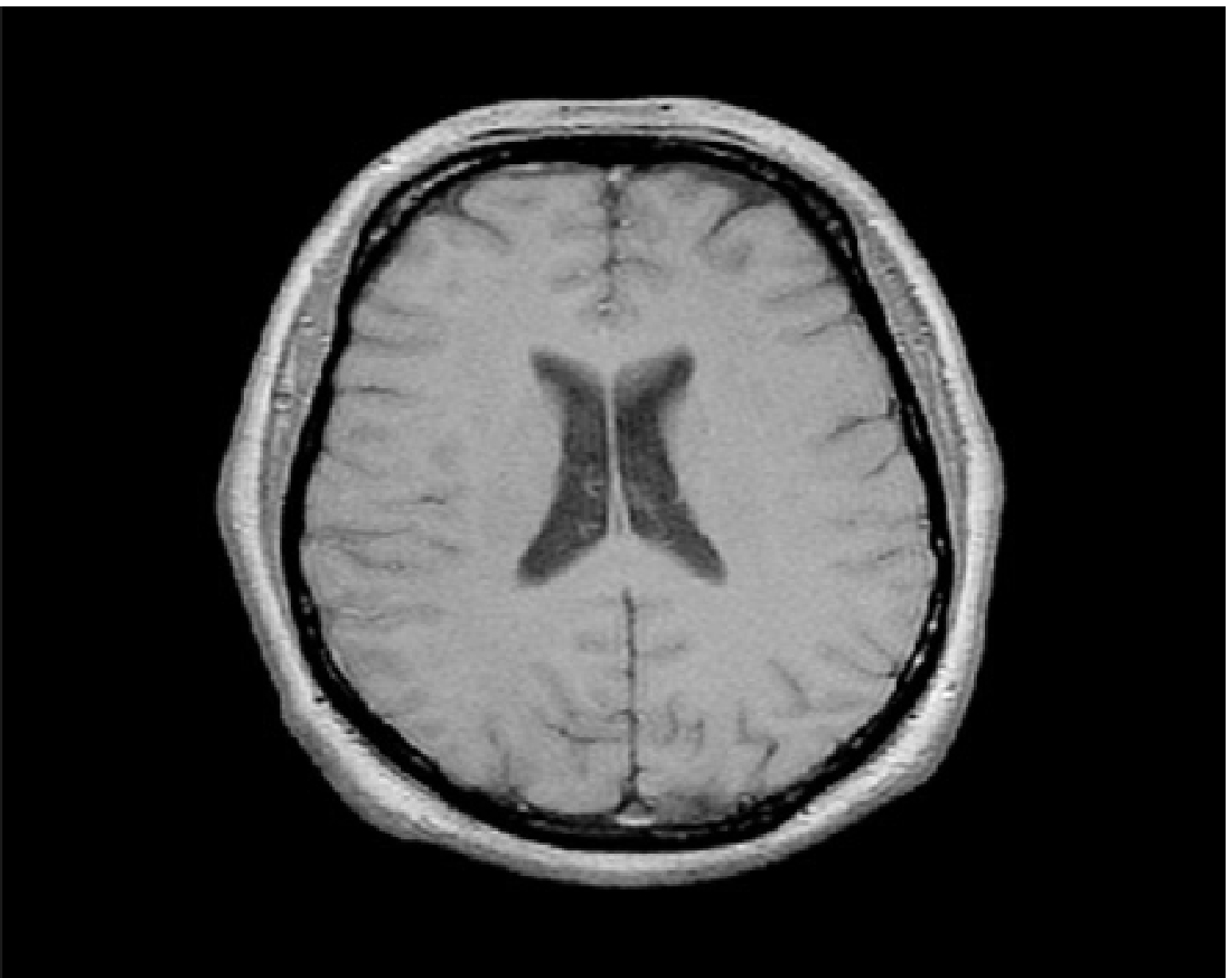}
\caption{An example of high resolution slice used by Ortiz et al. \cite{Ortiz2014117}}\label{fig:high-slice}
\end{subfigure}
\caption{Comparison of image qualities to illustrate the challenge in segmenting low resolution preterm neonate brain MR images.} % of slices used by us and Ortiz et al \cite{Ortiz2014117}}
\label{fig:slice-res}
\end{figure}

Ortiz et al. \cite{Ortiz2014117} applied an atlas-free fully automated method to segment brain MRIs into different types of tissues. During pre-processing, they removed the background noise in the image using Ostu’s method by minimizing the intra-class variance of the signal and noise voxels. Then, $24$ important 1\textsuperscript{st} order statistical features (intensity, mean and variance) and 2\textsuperscript{nd} order statistical features (energy, entropy, contrast etc.) were extracted. Most discriminatory features were selected using a Genetic Algorithm. Next, a Self-Organizing Map (SOM) was trained using the selected features in an unsupervised way. A label representing a type of brain tissue was then assigned to each SOM unit. Their method performed better than Constrained Gaussian Mixture Model \cite{1677729} in classifying WM and CSF, and gave promising results on high-resolution MRIs. Although this method is atlas-free, it does not work well on low-resolution preterm neonate MRIs. An example of a noisy low resolution ($96 \times 112$) preterm neonate MRI used to test our method is shown in Fig. \ref{fig:low-slice}, compared to a relatively noise-free image of much higher resolution ($512 \times 512$) used by Ortiz et al. \cite{Ortiz2014117} shown in Fig. \ref{fig:high-slice}.

Farzan \cite{Farzan2014} first segmented the brain tissues into GM, WM and CSF using Bayesian segmentation and then improved the results using domain knowledge obtained from experts in the form of heuristics. For applying the heuristics, the eight adjacent pixels of each pixel (its neighbours) were considered. A sample heuristic is: If ``neighbours are WM,'' then ``new centre is WM.'' The method assumes Normal Distribution of grey values in all tissues and uses Expectation Maximization (EM) to maximize the likelihood probability of tissues. They compared their outputs against the expert-annotated versions in terms of sensitivity and specificity of each tissue type. Although the use of heuristics to classify tissues is the strength of this method, the target feature on a single 2D slice may not be obvious due to noise or artifacts. To address this issue, we analyze adjacent slices to validate and recover candidate features.

In \cite{Snake2013} the authors applied skull-stripping as the initial step in the brain MR image segmentation process. The proposed hybrid skull-stripping algorithm, based on the adaptive balloon snake (ABS) model has two steps: (1) Pixel clustering using probabilistic fuzzy C-means (PFCM), which outputs a labelled image to identify the brain boundary; followed by (2) a contour initialized outside the surface of the brain. This contour is evolved guided by an ABS model. However, the ABS method employed here has the limitation that it ends up segmenting the contours into multiple objects. In comparison, our method effectively avoids the skull-stripping step and is able to discriminate irrelevant regions and false positives using a distance computation from the brain boundary.

In \cite{Jain2015} the authors propose ``MSmetrix,'' which is an automatic MRI-based lesion segmentation method. This method is acquisition device independent, meaning that no parameter tuning is needed for the type of scanner deployed. Differing from previous work \cite{938237}, where multi-channel images were used simultaneously for lesion segmentation, their approach incorporates human expert input. By using  T1-weighted and FLAIR images independently, they aim to fully exploit the characteristics of each modality. Their unsupervised approach segments 3D T1-weighted and FLAIR MRIs into WM, GM and CS following a probabilistic model, and treating WM lesions as outliers. The method assumes Gaussian distribution of the image intensities for each tissue class. However, this method uses MNI-atlas for skull stripping and GM, WM, CSF classification. Thus, it is not suitable for preterm neonates, where no reliable atlas is available as explained earlier.

There exist other very recent works on detection of brain lesions from MRIs \cite{Roy2017,Kaya2017}, but they are not specifically for preterm neonates. Some popular brain lesion detection methods are embedded in publicly available medical diagnostic software packages \cite{Schmidt2012,Shiee20101524}, but they require multiple scans of the same subject using multiple modalities \cite{Schmidt2012,Shiee20101524}. Our method only needs a single modality and takes T1 images as input. An example of using T1 images is the work of Griffis et al. \cite{Griffis2016}. It is a supervised method for detecting ischemic stroke lesions in T1-weighted MR scans using a naive Bayes classifier, which is trained on expert annotated scans. In contrast, our method is unsupervised and requires no training. Also, Griffis's work \cite{Griffis2016} performs probabilistic segmentation of the scanned MR slices into four tissue classes (GM, GM, CSF and non-brain tissues). The output is subsequently normalized to the Montreal Neurological Institute (MNI) template space using the New Segmentation tool implemented in SPM12 \cite{spm12web}. Although their MNI atlas cannot be used to register preterm brains because of the rapid structural changes in preterm brains as explained earlier, we still tried out their segmentation step on our dataset to analyze its time performance. Their method took 1 minute and 17.5 seconds to complete, whereas our proposed method took 41.5 seconds. This comparison further validates the fast time performance of our proposed method. Our time gain is attributed by skipping the segmentation of WM, and instead, approximating the normal range of WM intensities by collecting samples from the WM region.

Relatively less research has been done on WMI detection in preterm neonates compared to WMI in adults and other tissue abnormalities. In our earlier algorithm \cite{Cheng2015} we use a stochastic process that estimates the likelihood of intensity variations in target pixels belonging to a WMI. The first step is to detect the boundaries between normal and injured regions of the white matter. The next step is to measure pixel similarity to identify WMI regions. While the results showed effective WMI detection, the experiments were performed on relatively high resolution and noise-free slices, which may not often be the case for preterm neonate MR scans. In fact, as we will show later, when this method is tested on low-resolution noisy datasets, its accuracy is considerably lower. Also, the WMI prediction was done on individual slices, without considering adjacent slices. As we will show later, aggregating detection results of adjacent slices to predict WMI is crucial in low-resolution noisy scenarios, where the likelihood of detecting false positives and missing targets on a single slice is high.

\section{Methods}
\begin{figure}
\centering
\includegraphics[width=0.9\textwidth]{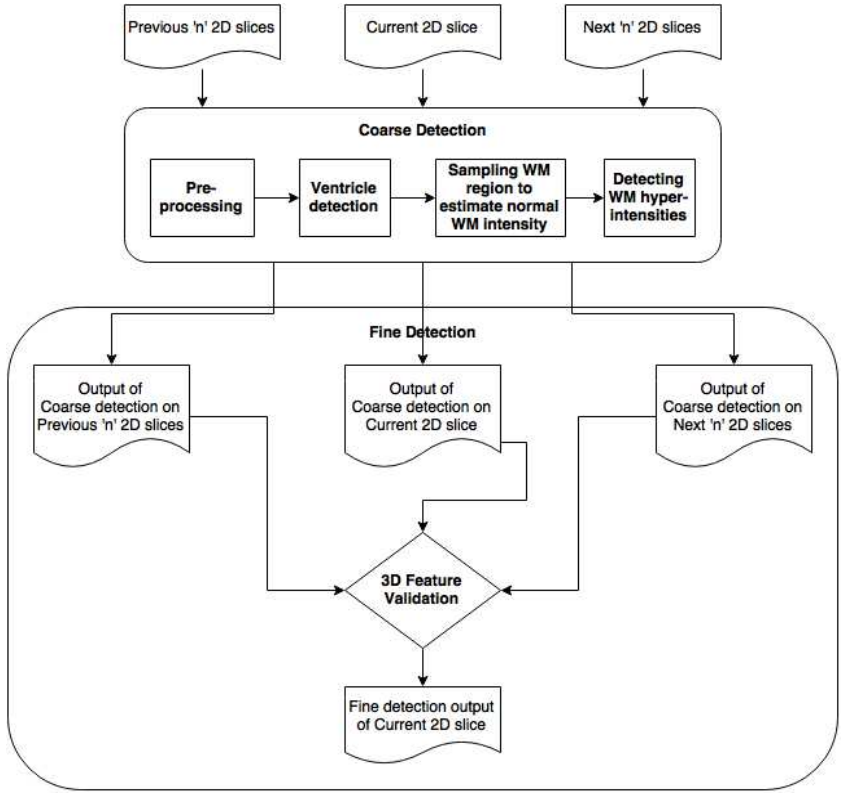}
\caption{An overview of our proposed method.} \label{fig:schematic}
\end{figure}

Our method considers both 2D and 3D spatial correlations. To give a high-level overview of our proposed method: we first detect potential WMIs in each 2D slice of the DICOM volume (coarse detection). We then analyze the WMI pixel correlation between each 2D slice and its adjacent slices in the DICOM volume. In the Fine Detection process, the Coarse Detection result obtained from a 2D slice is then validated with its adjacent slice information, in order to reduce false positives and recover true positives. An overview of the proposed method is shown in Fig. \ref{fig:schematic}. Note that $n$ adjacent slices are defined in the computational model, and $n = 1$ is used in the current implementation.

\subsection{Coarse Detection}

Our central idea behind White Matter Injury detection is to search for abrupt intensity peaks (hyper-intensities) in the white matter region of the brain. Since the intensities in the white matter are normally distributed \cite{Farzan2014,Jain2015}, the WMI represents an ``outlier'' with respect to the range of white matter intensities. As described later, we detect the ventricles as ``blobs'' in a 2D slice image. We then use a contour roughly between the ventricles and the brain boundary for estimating the range of white matter intensities. This is done by considering the intensities inside that contour which do not belong to the ventricles. The outlier intensities which are greater than the ``normal'' range would be potential candidates for WMI. In order to eliminate false positives, we further filter these candidates based on the size and distance criteria before a WM hyper-intensity is classified as WMI. All of these steps are performed on a single 2D slice (coarse detection).

Note that unlike traditional brain lesion detection methods, our method does not need to segment the WM into distinct patches. Instead, we sample the WM region to estimate the normal range of WM intensities. As shown later, this results in a significant reduction in execution time (without compromising accuracy) when compared to earlier segmentation-based WMI detection work \cite{Cheng2015}. However, if the ventricles are included in the estimation of WM normal intensities, the result will be unreliable. Thus, we identify the ventricles and eliminate them from our samples, using a confidence-based patch classification technique described later.

We next describe each individual step of the coarse detection phase in detail, and subsequently we describe the fine detection phase. The reader can refer to Fig. \ref{fig:op-mains} for outputs of individual steps of this entire processing pipeline.

\subsubsection{Background segregation}

Anisotropic Diffusion was used as a pre-processing step to de-noise those brain MRIs which were extremely noisy. The parameters used were $\frac{1}{7}$ for the integration constant, $3$ for the gradient modulus threshold and the 2\textsuperscript{nd} conduction coefficient function as proposed by Perona \& Malik \cite{Perona:1990:SED:78302.78304}, because these parameter values, with $15$ iterations, produced the best results.

Ostu's method \cite{ostu} was applied for separating the brain from the (noisy) background. Ostu's method separates an input grey-scale image into foreground and background by determining a global threshold to minimize the intra-class variance of foreground and background pixels. It iterates through all the possible thresholds in the image to find the threshold that gives the smallest within class variance.

Thus, by running Ostu's method on the input image $I$, we obtained a binary threshold $Th_O$, such that intensities in $I$ falling above $Th_O$ constitute the foreground as determined by Ostu's method. The \textit{holes} in the foreground were filled using the ``fill'' operation based on morphological reconstruction \cite{Soille2004}, giving foreground mask $M_f$.

Similarly, the background mask $M_b$ is obtained by taking the complement of $M_f$ (Eq. \ref{eq:mbmf}). Using the background mask, we ``clean'' the background by altering the intensities of all background pixels to 255 (white). Thus, we remove the unwanted intensity variations (noise) in the background, which may interfere with subsequent processing. The rationale behind setting the background to white (and not black) will be explained later.

\begin{equation}\label{eq:mbmf}
M_b = \overline{M_f}
\end{equation}

\subsubsection{Modelling ventricles as a collection of Maximally Stable Extremal Regions}

We detect the ventricle as blobs inside the brain using the Maximally Stable Extremal Regions (MSER) algorithm \cite{MSERoriginal}. MSER in an image is a connected region, which can be detected by an extremal property of the intensity function in the region and on its outer boundary. MSERs have properties that assist in their superior performance as a stable local detector. First, the set of MSERs is closed under continuous geometric transformations. Second, MSERs are invariant to affine intensity changes and finally, MSERs are detected at different scales.

\subsubsection{Confidence based image patch classification for Ventricle detection}

In order to prevent the MSER algorithm from including other parts of the brain as connected regions of the ventricles, we filter out these invalid parts by assigning a confidence value to each region based on its T1 intensity and its distance from the brain boundary. The applied criteria were motivated by the observation that ventricles are comprised of low-intensity (T1) image patches, and are near the center of the brain. To detect the ventricles, we first calculate the following matrices:

\begin{description}
	\item[$D_p$] \hfill \\
	The distance transform \cite{bwdist} of $M_b$ (using the L1 norm or city-block distance measure) is normalized to the $[0,1]$ range. We experimented using the L1 and L2 norms and got better results with L1. Distance transform of each point inside the brain gives its distance to the nearest point lying on the brain boundary (background mask). Thus, the points lying more towards the centre of the brain (where we expect to find the ventricles) tend to get higher values.
	\item[$I_c$] \hfill \\
	The normalized complement of $I$: Since $I$ is a grey-scale image, this means $I_c = 255 - I$. $I_c$ is then normalized to the $[0,1]$ range. As the points inside the ventricles have low grey scale values (T1 images), the inverted image has high grey scale values. Recalling that in an earlier step, the background was marked white, allowing the ventricles to be easily extracted in the inverted image. After the ventricle detection step, the background is inverted to black with a ``zero'' value.
	\item[$L_p$] \hfill \\
	The $[0,1]$ normalized \textit{Hadamard} product of the matrices $D_p$ and $I_c$, i.e., $L_p = D_p \bullet I_c$: This means that the distance transform of each point is multipled with its (inverted) intensity. Following the definition of $D_p$ and $I_c$, we can infer that the points lying inside the ventricle will have very high values for $L_p$, which can be used as a ``confidence'' measure for determining if a given
point inside the brain belongs to the ventricles.
\end{description}

We apply the MSER algorithm on $I$, which detects blobs. MSER is one of the fastest region detectors because of its linear implementation \cite{MSERlinear}. It is also affine invariant, has good repeatability and performs well to classify patches with similar grey scale values on images containing homogeneous regions with distinctive boundaries. Let us assume that the MSER extracts $R$ regions from $I$, denoted as $r_1$, $r_2$, $r_3$, ..., $r_R$. For each region $r_i$, we calculate the ``average confidence'' $C_i$, which is the average of $L_p$ values for the pixels belonging to $r_i$. Thus, our problem is reduced to finding the ``maximal set of optimal regions'' $S_v = \{r_1, r_2, ... , r_v\}$ forming the ventricles. As demonstrated later, sub-optimal ventricle detection results do not affect the accuracy of WMI detection.

\subsubsection{Patch fitness evaluation for Ventricle detection}

We model ventricle detection as an optimization problem and solve it using Genetic Algorithm (GA) \cite{Goldberg:1989:GAS:534133}. When applying GA to our preterm brain images, for each region $r_i$ (mentioned earlier), we have to make a \textit{binary} choice of either to include it in a \textit{candidate solution}, or leave it out. Thus, we represent a candidate solution as a \textit{bit string}. We define a \textit{fitness function} to obtain the candidate selection solution:

\begin{equation}
F_s = N_s * C_1 * C_2 * ... * C_{N_s}
\end{equation}

\noindent where $N_s$ is the number of regions selected and $C_j$ refers to the confidence of the selected region $j$. Thus, a GA returns the \textit{optimal} choice of regions $S_v$, which are \textit{most likely} to constitute the ventricles. As we will see later, this sub-optimal result of ventricle detection would not affect the overall effectiveness of the proposed method.

\noindent We also define a mask $M_v$ for all pixels $p_i$ belonging to the ventricles determined by the GA.

\begin{equation}
M_v = \{p_i \in r_j \thinspace \forall \thinspace r_j \in S_v\}
\end{equation}

\subsubsection{Detecting WM hyper-intensities}

Next, we consider the white matter (WM) region around the ventricles. Our goal is to exclude the region where no WMI is present. Let $D_v$ be the distance transform of $M_v$ and we choose a contour that follows the relation:

\begin{equation}\label{eq:contour}
\abs{D_p - D_v} \leq 1
\end{equation}

\noindent
This generates a contour, whose points are roughly halfway between the ventricles and the brain boundary. Thus, the mask $M_w$ \textit{enclosed} by the contour gives the candidate region, with the ventricles and the patches falsely detected as ventricles represented as ``holes''. However, as we will show later, false ventricle detection does not affect the final WMI extraction. Also, ventricle patches included in $M_w$ undetected by MSER (or GA) will not interfere in WMI detection, as their intensities are below the WM mean intensity.

Using the pixels described by $M_w$, we calculate their Median $M_d$ and Median Absolute Deviation $M_a$. Any grey level $g$ in the image $I$, satisfying Eq. \ref{eq:Mc}, represents a potential WMI based on the ``Modified Z-score'' metric \cite{iglewicz1993detect}.

\begin{equation}	\label{eq:Mc}
0.6745 \times \frac{g - M_d}{M_a} > 0
\end{equation}

\noindent The Modified Z-score uses the \textit{median} instead of the \textit{mean}, as the former is more robust to outliers. Also, as mentioned earlier, the parts of the ventricles undetected by MSER (or GA) lie in the range of values for $g$ satisfying Eq. \ref{eq:nMc}, and thus are not included as WMI.

\begin{equation}	\label{eq:nMc}
0.6745 \times \frac{g - M_d}{M_a} < 0
\end{equation}

\subsection{Filtering WM hyper-intensities}

The next step is to define a mask $M_c$ for the potential WMI candidates, such that pixels of image $I$ whose grey level $g$ satisfies Eq. \ref{eq:Mc} belong to $M_c$. We enumerate the 8-connected \textit{objects} found in $M_c$ using the method described in \cite{bwlabel} (page 40-48). The general procedure is described below.

\begin{enumerate}
\item Run-length encode the input image.
\item Scan the runs, assign preliminary labels, and store label equivalences in a local equivalence table.
\item Resolve the equivalence classes.
\item Relabel the runs based on the resolved equivalence classes.
\end{enumerate}

The above procedure returns a set of objects $S_w = \{O_1, O_2, ... , O_w\}$. We then compute the sizes of the corresponding objects as $\{N_{O_1}, N_{O_2}, ... , N_{O_w}\}$. We discard the $5\%$ \textit{largest} objects, e.g., segments of the skull boundary, as they are outliers. Our experimental observations verified that in T1 images, the skull shares similar intensity as the WM injuries; both lie above the normal range of WM intensities. Thus, the skull naturally forms the biggest objects in the set $S_w$ (within the top $5\%$). In the next step, we perform a \textit{binary} classification (``big'' and ``small'') of the remaining objects based on their sizes, using the K-means clustering algorithm. We initialize the starting means (or ``centroids'') of the K-means algorithm with the sizes of the smallest and the largest objects (among the remaining $95\%$). Note that the WM injuries often fall in the ``small'' category, while the ``big'' category contains brain tissue boundaries. We impose these ``size constraint'' in our algorithm. Using the process described above, we further enforce a ``distance constraint'', based on our expert-annotated dataset, that WM injuries cannot lie close to the skull. More false positive WMI detections are eliminated subsequently in the fine detection process described next.

\subsection{Fine Detection}

The second phase of our algorithm combines the coarse detection results from adjacent DICOM slices. This is motivated by the understanding that white matter injury (WMI) position cannot change abruptly across slices. We consider slice number $i$ in which a WMI is detected at a particular location $(x, y)$; then, the same WMI spanning across slice $(i \pm n)$  will be roughly at the same location ($n = 1$ in the current implementation). We allow a \textit{distance tolerance threshold} to account for a slight variation in position. By interpolating between slices, the algorithm can also identify noise and recover occluded WMI caused by weak intensity contrast. False positives can be reduced by considering adjacent slices. Since the coarse detection step is computationally efficient, the overall time performance is improved because fine detection is applied only on a small set of slices, which contain potential WMI candidates.

Let $l_i \in L_{-1}$, $l_j \in L_{0}$ and $l_k \in L_{1}$ be the centroids of the WMI regions detected in slice numbers $(n - 1)$, $n$ and $(n + 1)$ respectively. Our fine detection constraint is defined in Eq. \ref{eq:Fd1} and Eq. \ref{eq:Fd2}, where $T$ is the set of true positives predicted by the fine detection step and it consists of elements $l_j$ which satisfies either Eq. \ref{eq:Fd1} or Eq. \ref{eq:Fd2}. The notation $dist$ is the distance between two cluster centres and $D_{th}$ denotes the distance tolerance threshold. We use a normalized Euclidean distance $(0,1)$ with $D_{th} = 0.1$. Experiments showed that these parameter values produced the best results.

\begin{equation}\label{eq:Fd1}
l_j \in T \Leftrightarrow \exists \ l_i \in L_{-1} \mid dist(l_j, l_i) \leq D_{th}
\end{equation}
\begin{equation}\label{eq:Fd2}
l_j \in T \Leftrightarrow \exists \ l_k \in L_{1} \mid dist(l_j, l_k) \leq D_{th}
\end{equation}

\section{Results}	% Subject 1 ==> 0038, Subject 2 ==> 0024, Subject 3 ==> 0042

\begin{figure}
  \newcommand*\FigVSkip{0.5em}
  \newcommand*\FigHSkip{0.1em}
  \newsavebox\FigBox
  \centering
  \sbox{\FigBox}{\includegraphics[scale=0.1]{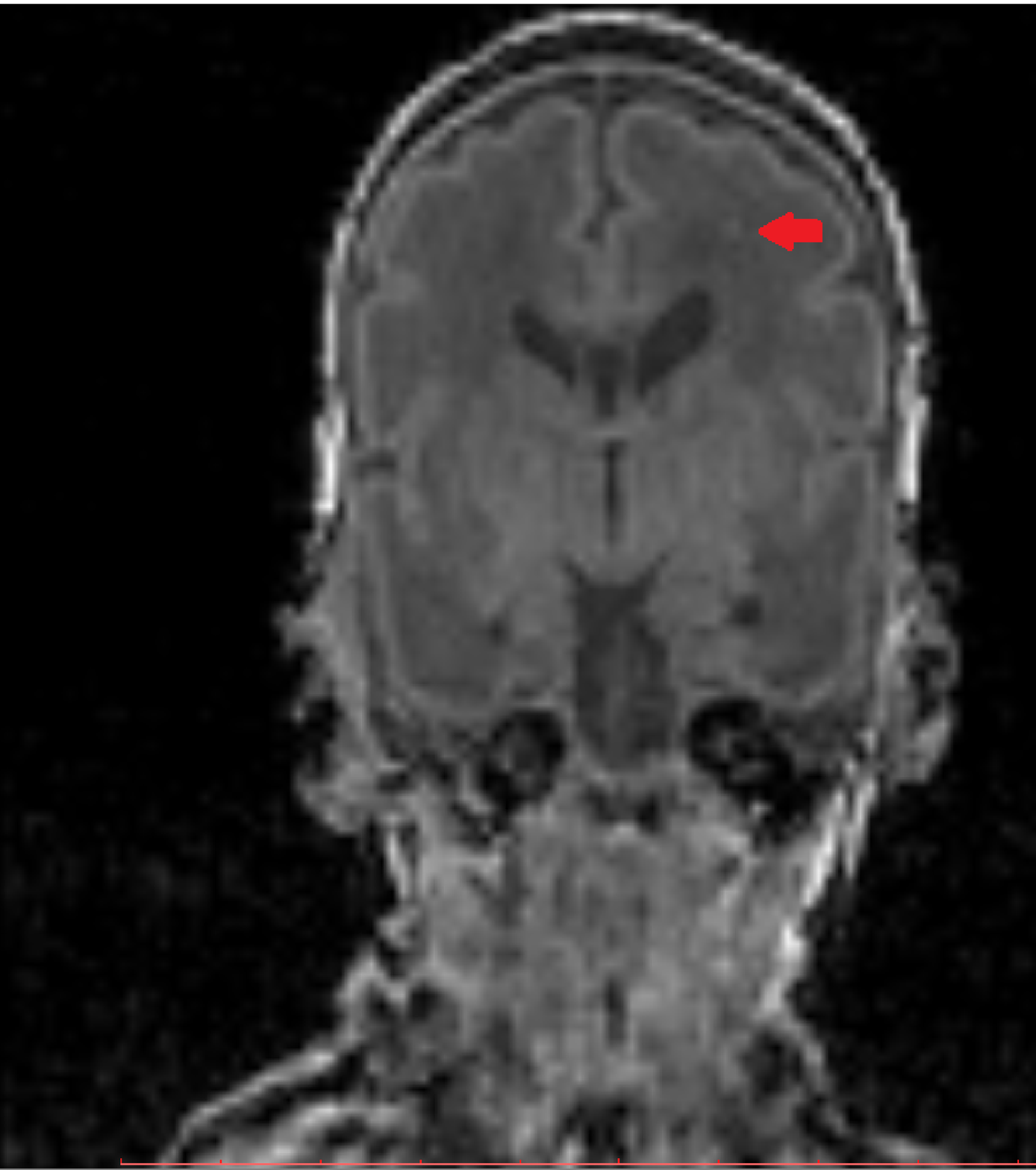}}
  \begin{minipage}{\wd\FigBox}
    \centering\usebox{\FigBox}
    \subcaption{WMI marked on a 2D slice from DICOM stack of 2\textsuperscript{nd} subject}
  \end{minipage}\hspace*{\FigHSkip}
  \sbox{\FigBox}{\includegraphics[scale=0.12]{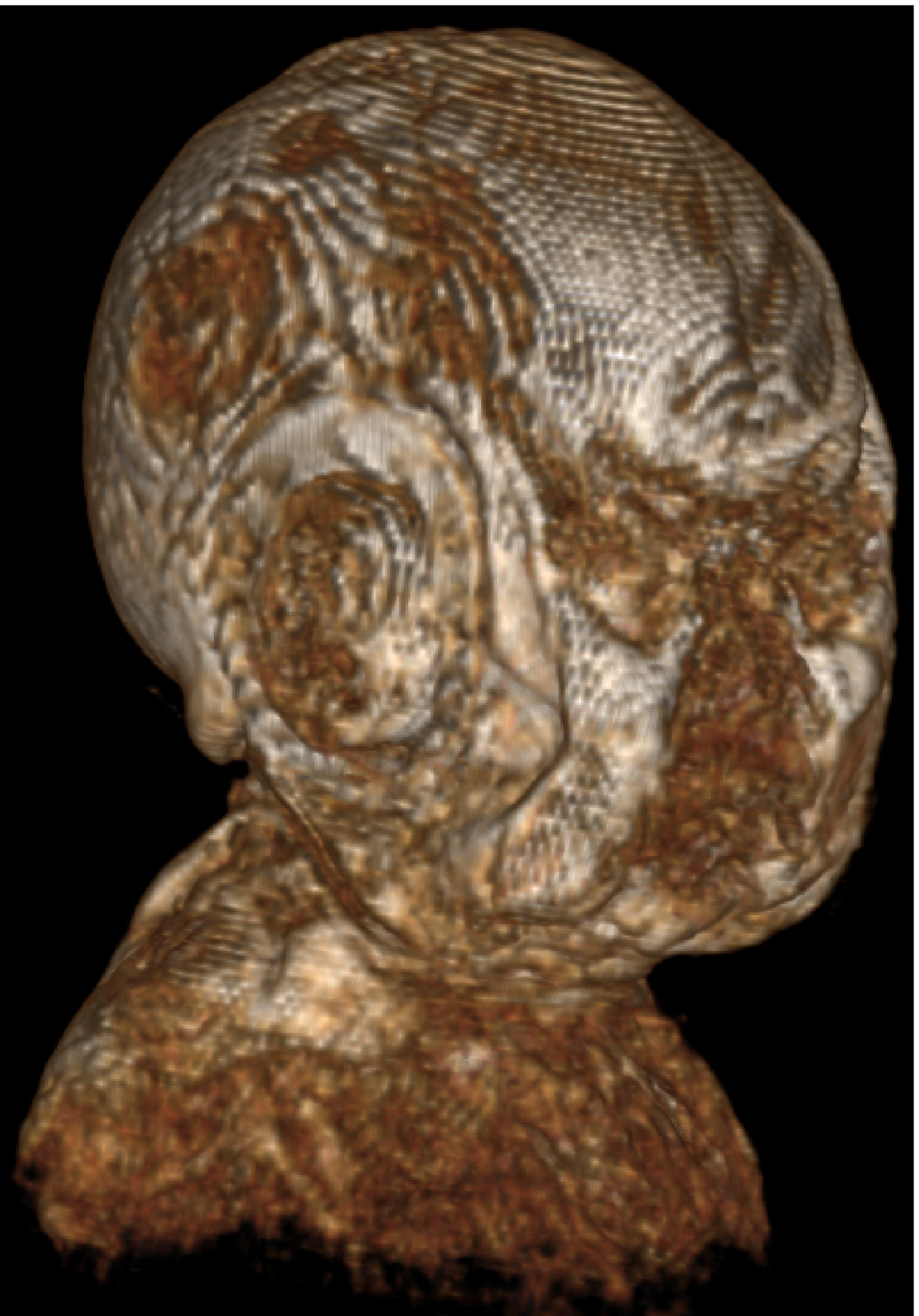}}
  \begin{minipage}{\wd\FigBox}
    \centering\usebox{\FigBox}
    \subcaption{3D reconstruction from DICOM stack of 2\textsuperscript{nd} subject}
  \end{minipage}\hspace*{\FigHSkip}
  \sbox{\FigBox}{\includegraphics[scale=0.12]{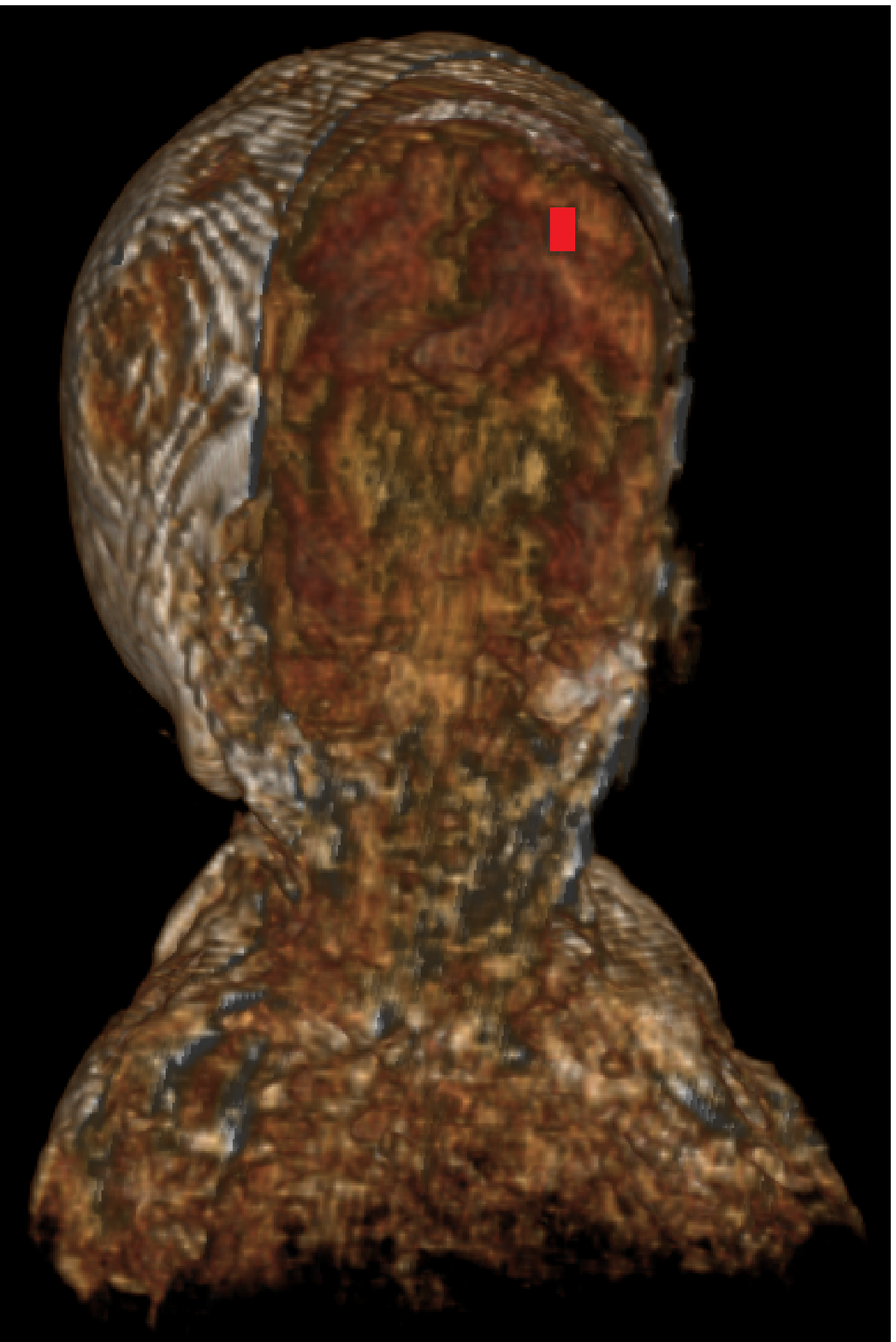}}
  \begin{minipage}{\wd\FigBox}
    \centering\usebox{\FigBox}
    \subcaption{Visualization of WMI on coronal cross-section of 2\textsuperscript{nd} subject}
  \end{minipage}
  \sbox{\FigBox}{\includegraphics[scale=0.12]{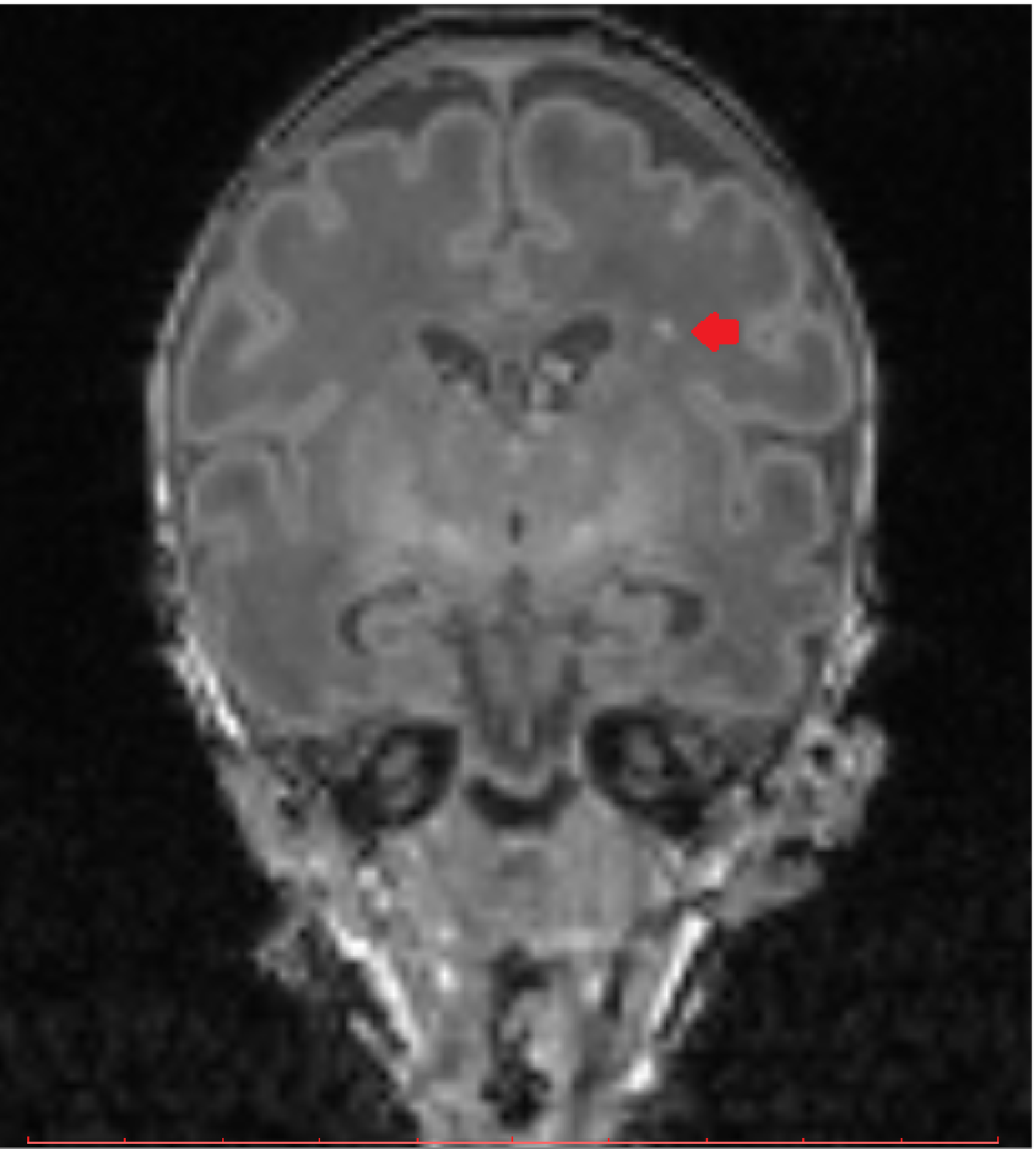}}
  \begin{minipage}{\wd\FigBox}
    \centering\usebox{\FigBox}
    \subcaption{WMI marked on a 2D slice from DICOM stack of 3\textsuperscript{rd} subject}
  \end{minipage}\hspace*{\FigHSkip}
  \sbox{\FigBox}{\includegraphics[scale=0.12]{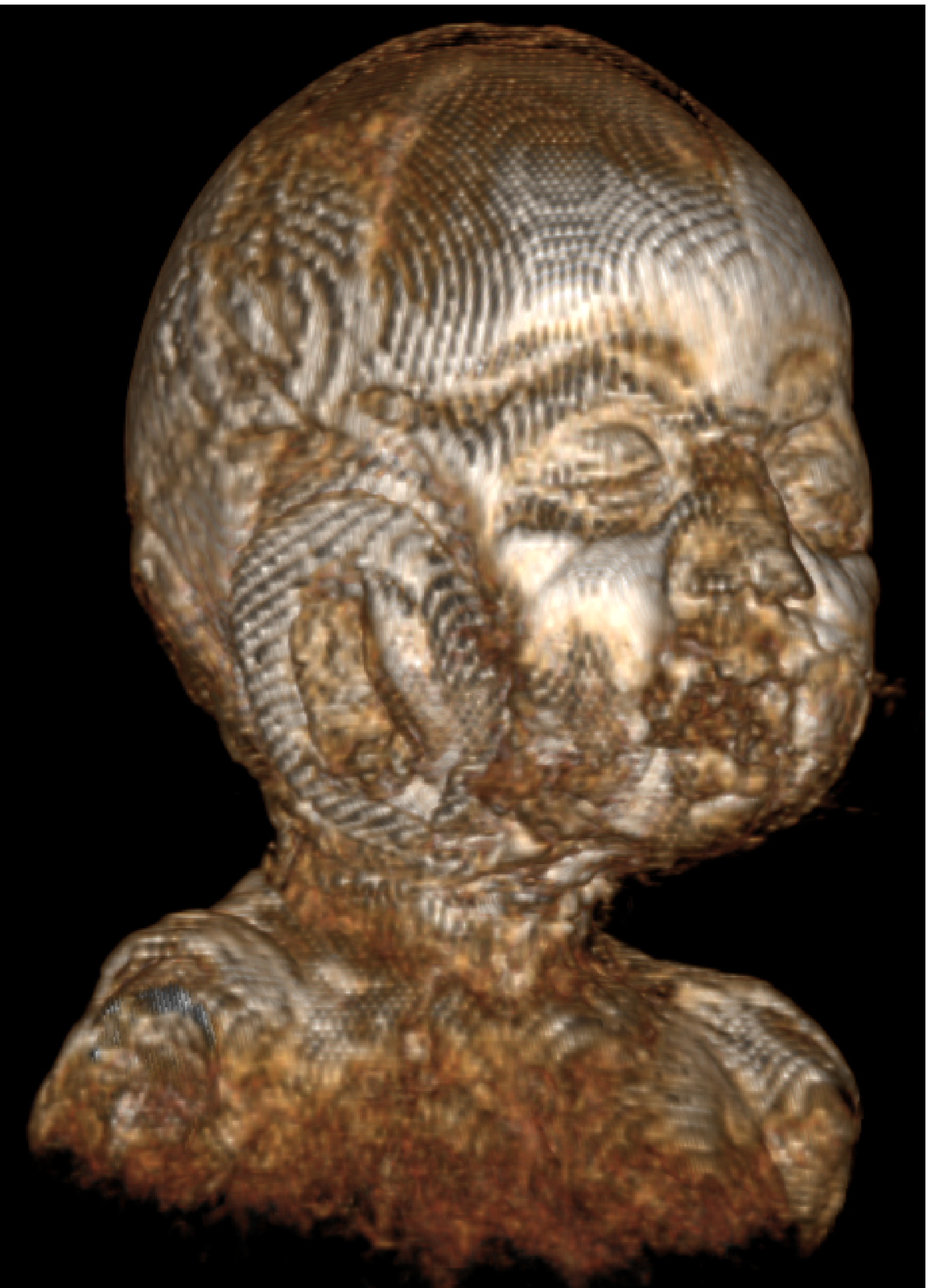}}
  \begin{minipage}{\wd\FigBox}
    \centering\usebox{\FigBox}
    \subcaption{3D reconstruction from DICOM stack of 3\textsuperscript{rd} subject}
  \end{minipage}\hspace*{\FigHSkip}
  % Save second image 
  \sbox{\FigBox}{\includegraphics[scale=0.12]{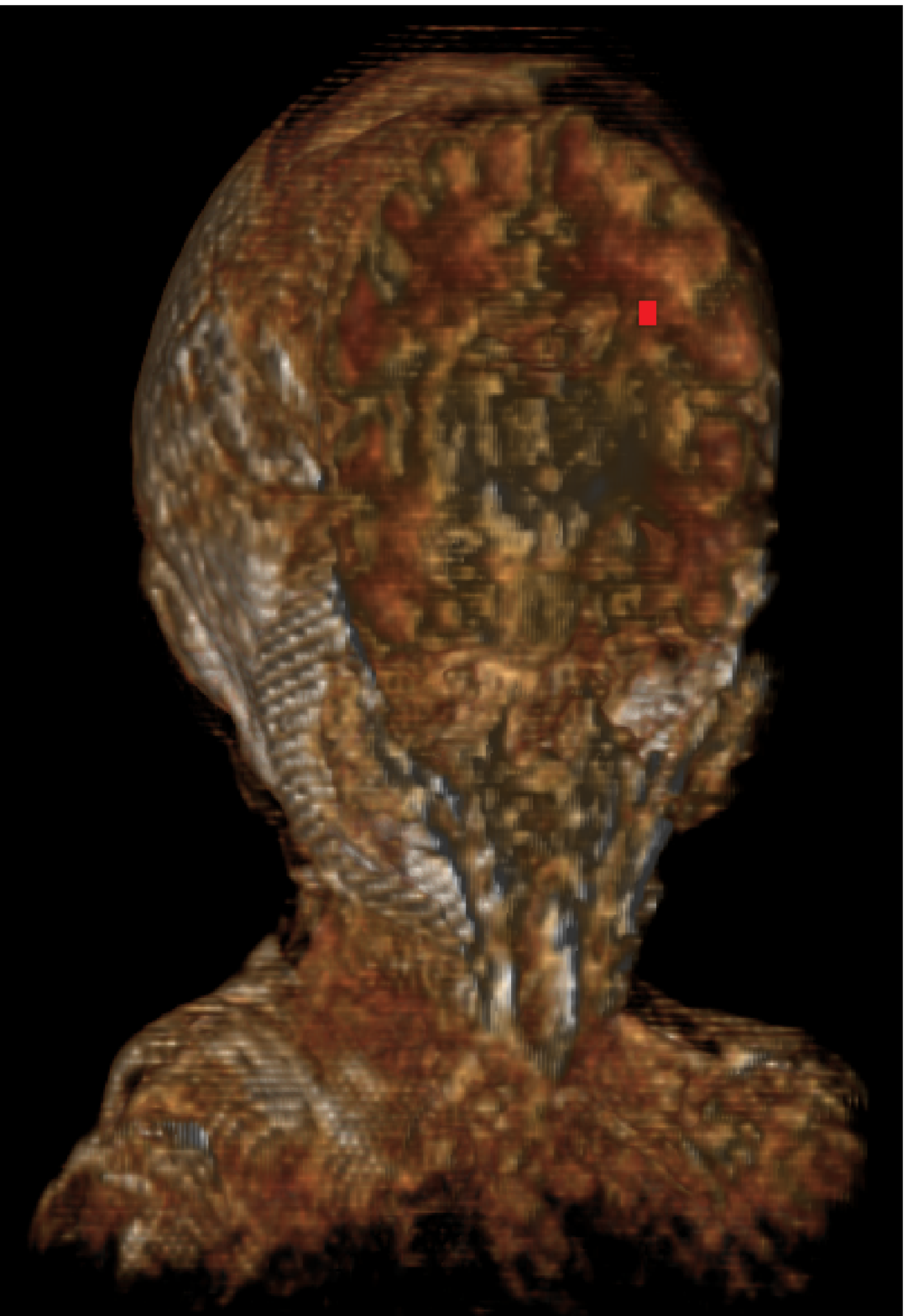}}
  \begin{minipage}{\wd\FigBox}
    \centering\usebox{\FigBox}
    \subcaption{Visualization of WMI on coronal cross-section of 3\textsuperscript{rd} subject}
  \end{minipage}
  \caption{Visualization of representative Ground Truth WMIs on coronal cross-sections of subjects.}
  \label{fig:Vis3D}
\end{figure}

We evaluate our method qualitatively and quantitatively using noisy and low-resolution ($96 \times 112$) preterm neonate brain DICOM slices from three subjects provided by the SickKids Hospital in Toronto. Wherever applicable, WMIs were marked by expert radiologists on the slices as ground truth. Fig. \ref{fig:Vis3D} presents two representative Ground Truth WMIs marked on coronal cross-sections of 3D models of the two of those subjects.

\begin{figure}
\begin{subfigure}[b]{.24\linewidth}
\includegraphics[width=\linewidth]{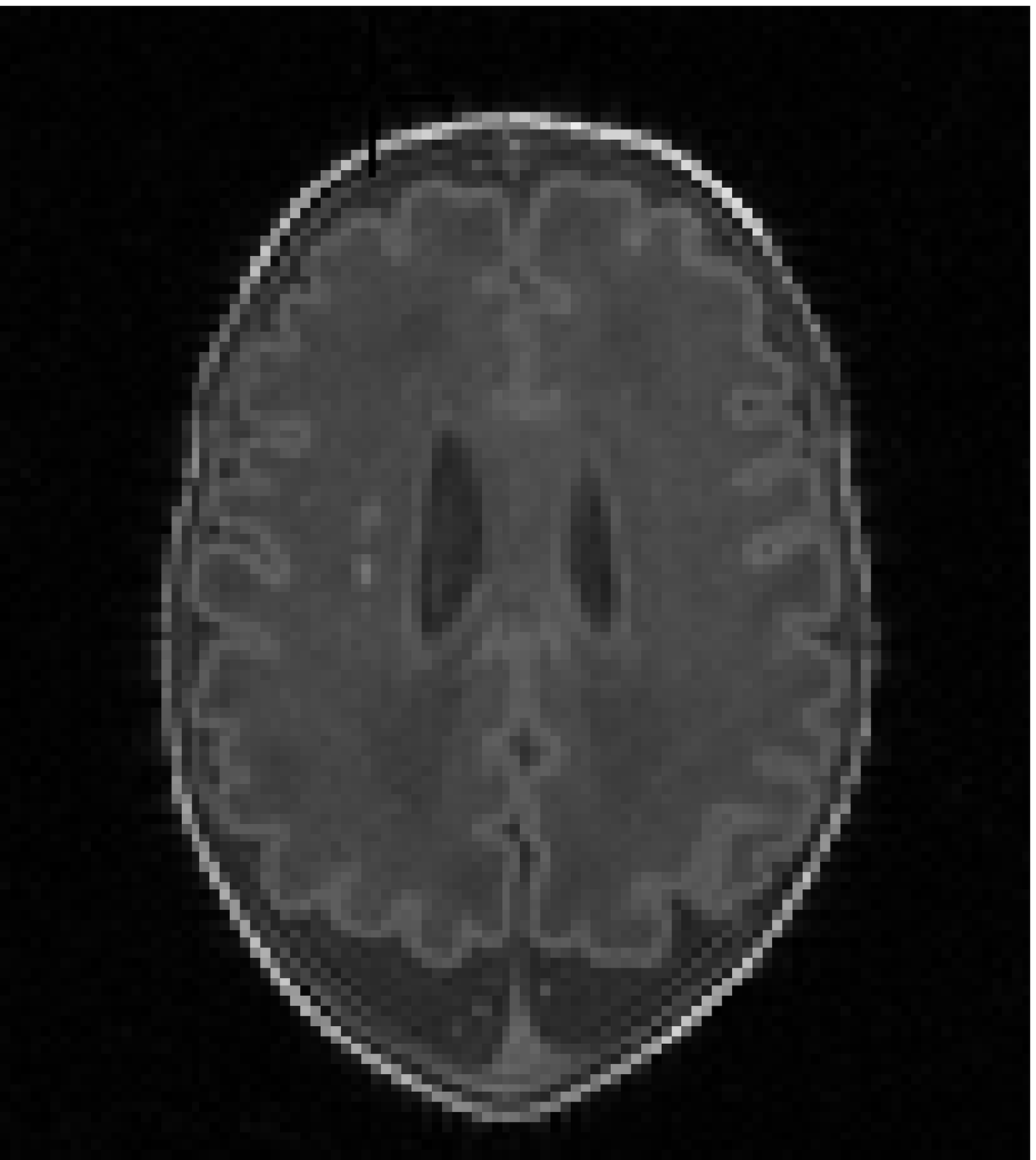}
\caption{Input Slice}\label{fig:input-slice}
\end{subfigure}
\begin{subfigure}[b]{.24\linewidth}
\includegraphics[width=\linewidth]{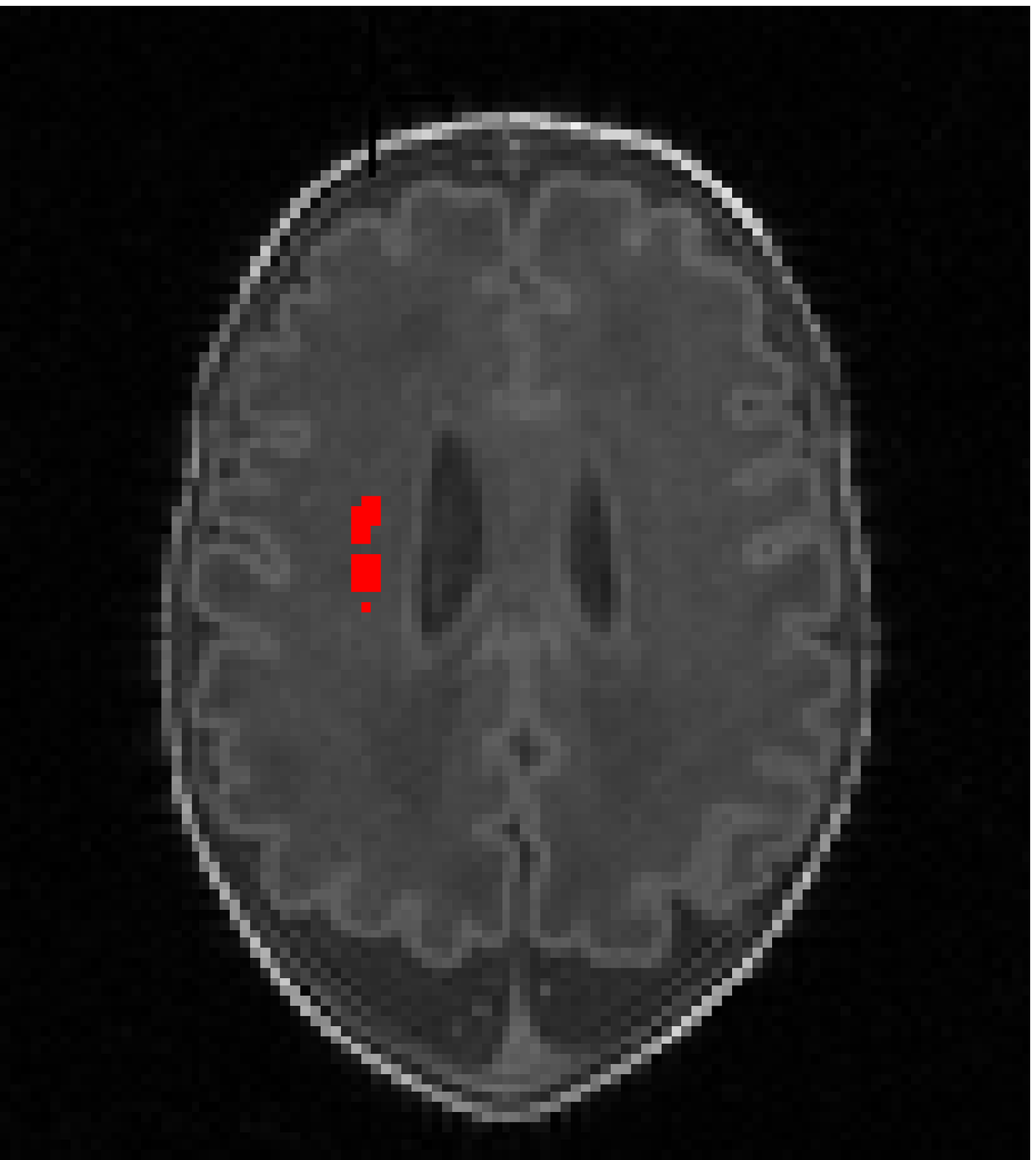}
\caption{Ground Truth with annotated WMI (Red)}\label{fig:ground-truth}
\end{subfigure}
\begin{subfigure}[b]{.24\linewidth}
\includegraphics[width=\linewidth]{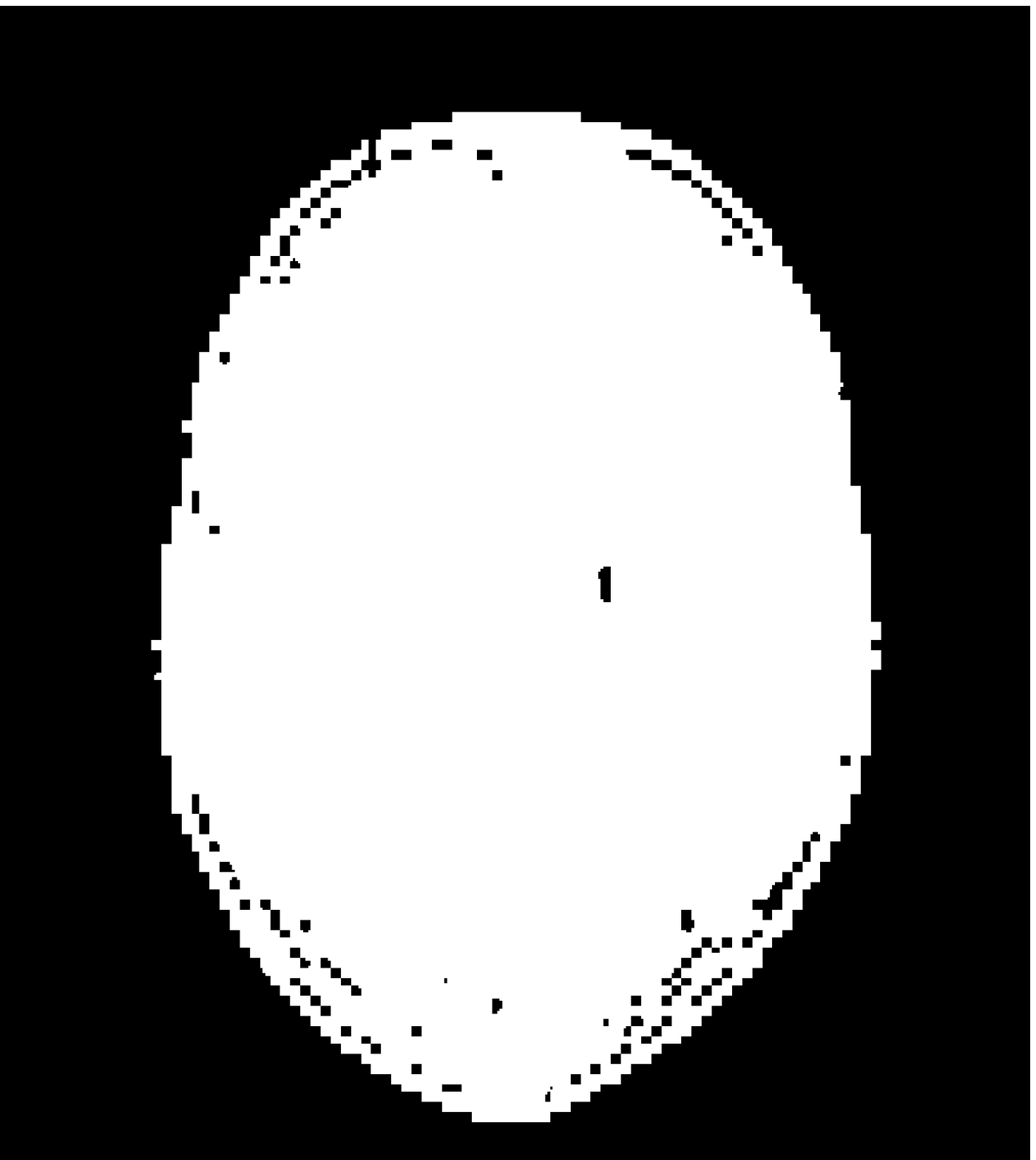}
\caption{Ostu's method}\label{fig:op-ostu}
\end{subfigure}
\begin{subfigure}[b]{.24\linewidth}
\includegraphics[width=\linewidth]{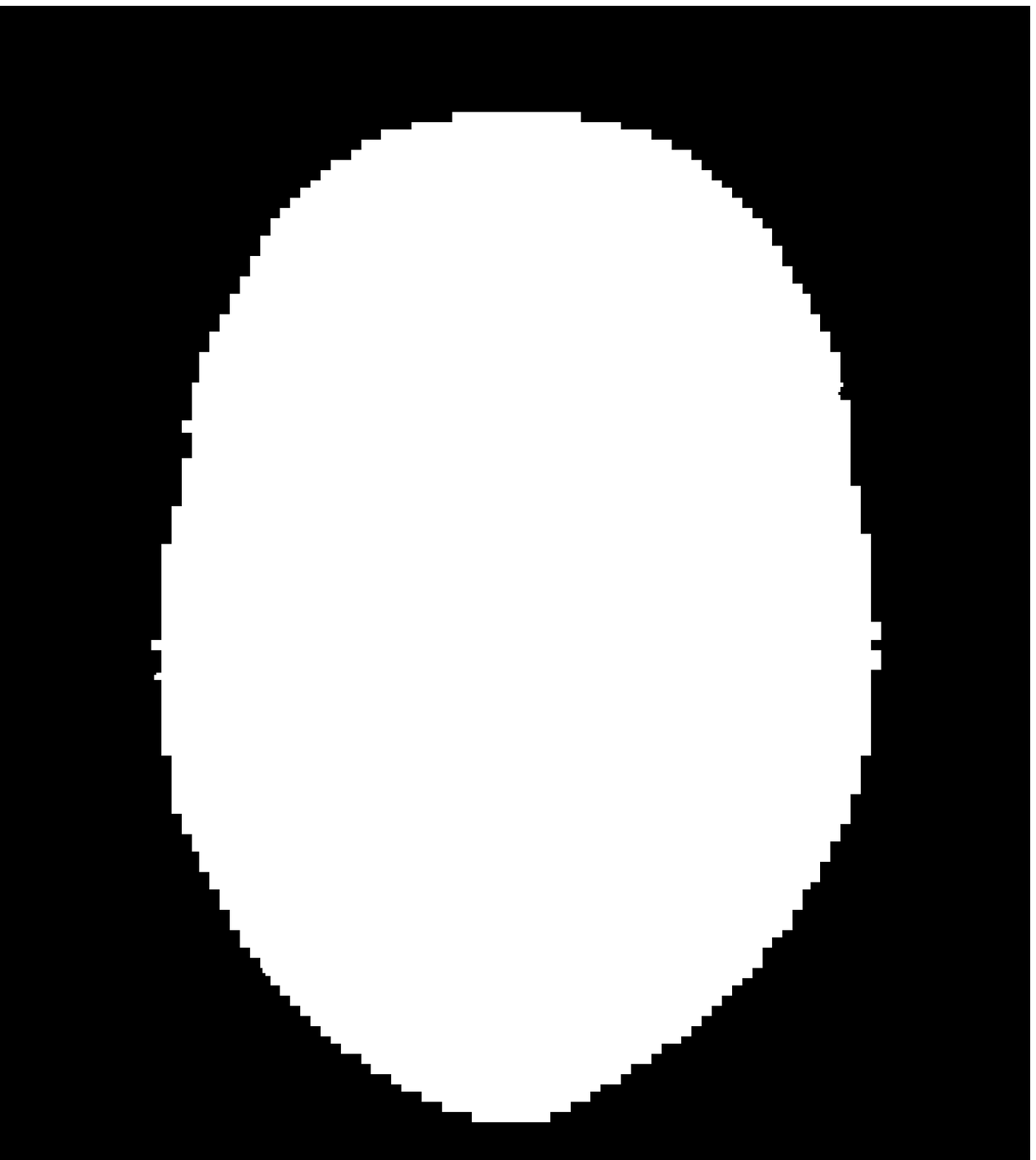}
\caption{Hole-filling}\label{fig:op-hole}
\end{subfigure}

\begin{subfigure}[b]{.24\linewidth}
\includegraphics[width=\linewidth]{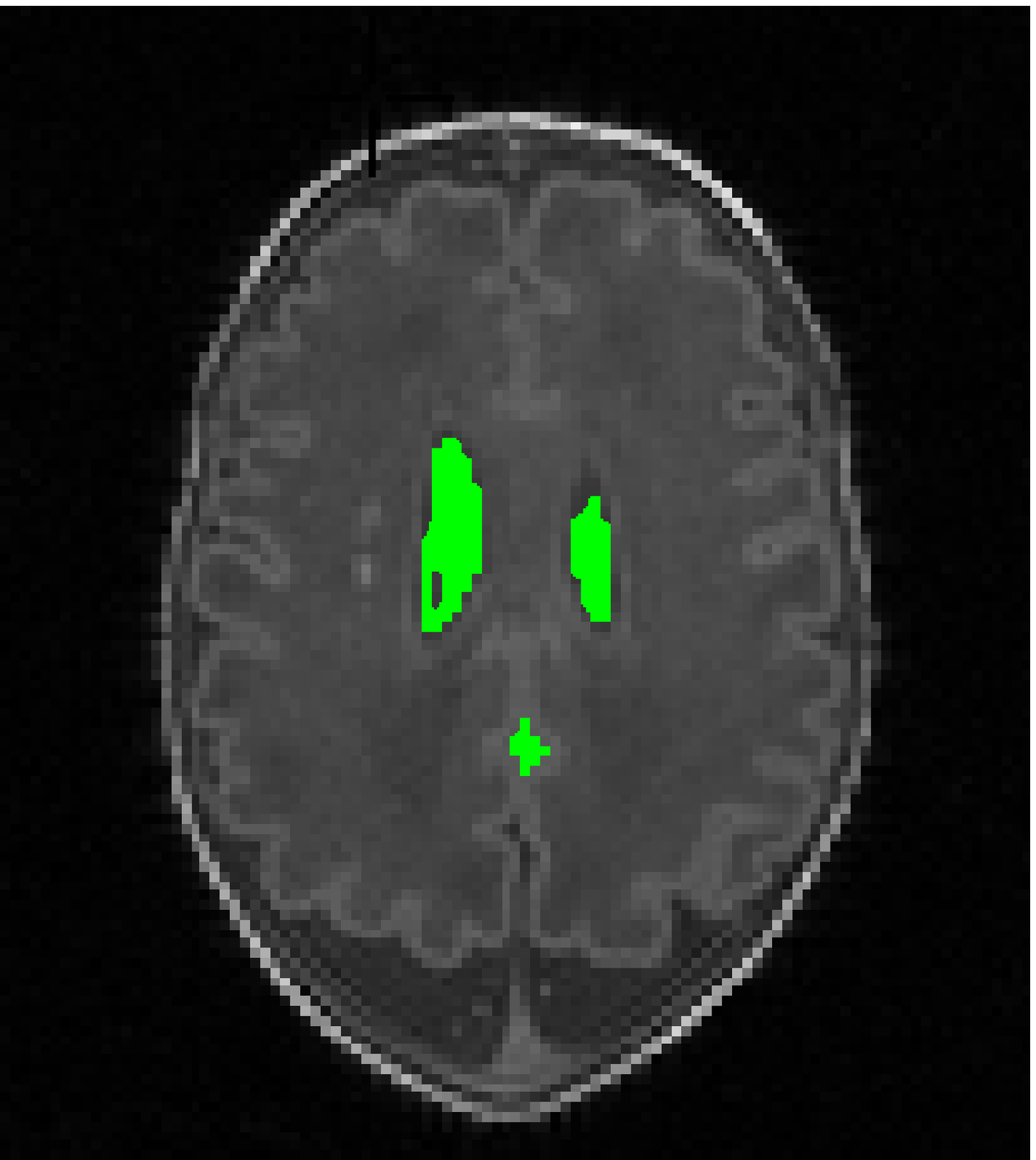}
\caption{Ventricle detection}\label{fig:ventricle}
\end{subfigure}
\begin{subfigure}[b]{.24\linewidth}
\includegraphics[width=\linewidth]{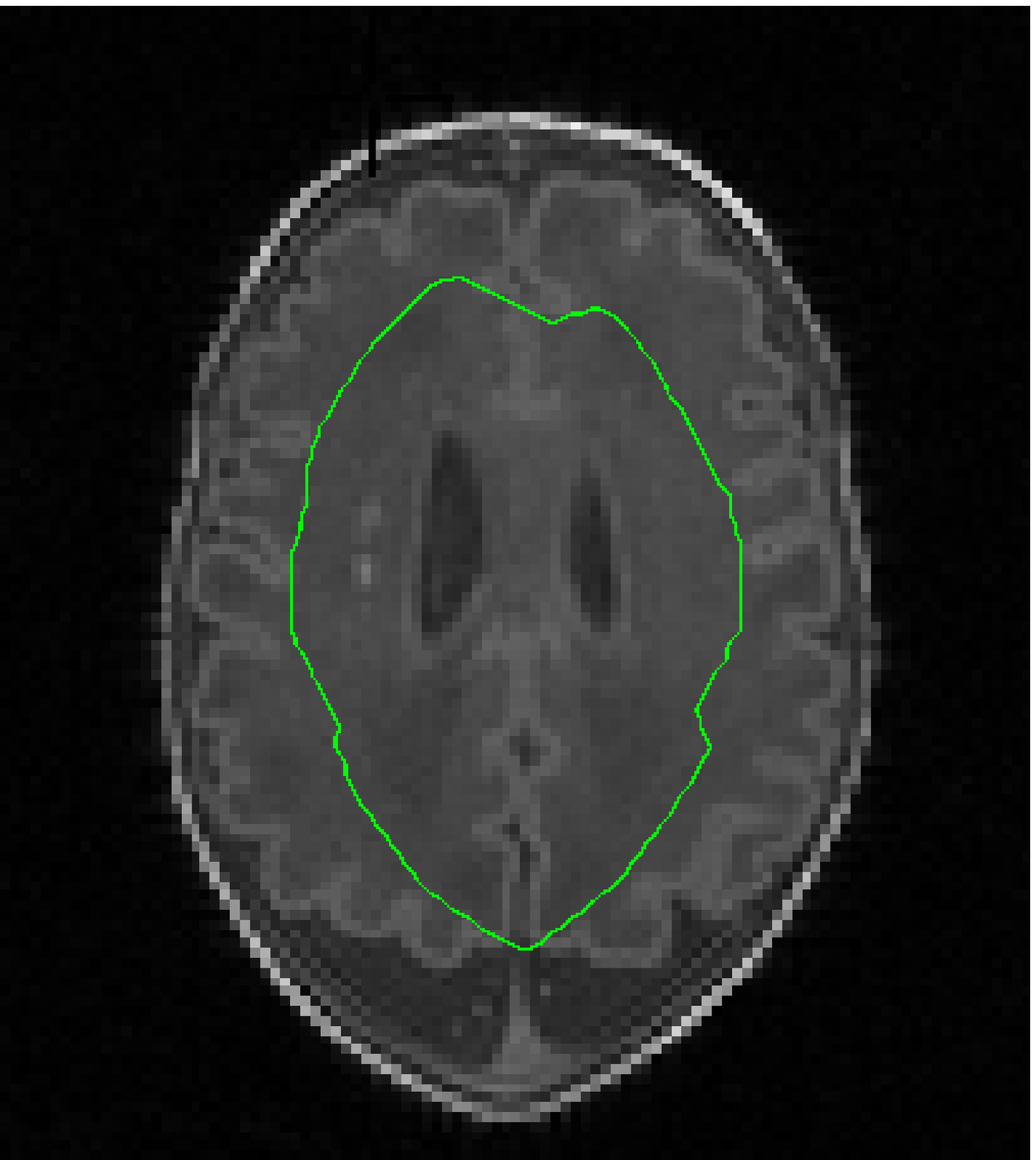}
\caption{Contour of Eq. \ref{eq:contour}}\label{fig:brain-crop}
\end{subfigure}
\begin{subfigure}[b]{.24\linewidth}
\includegraphics[width=\linewidth]{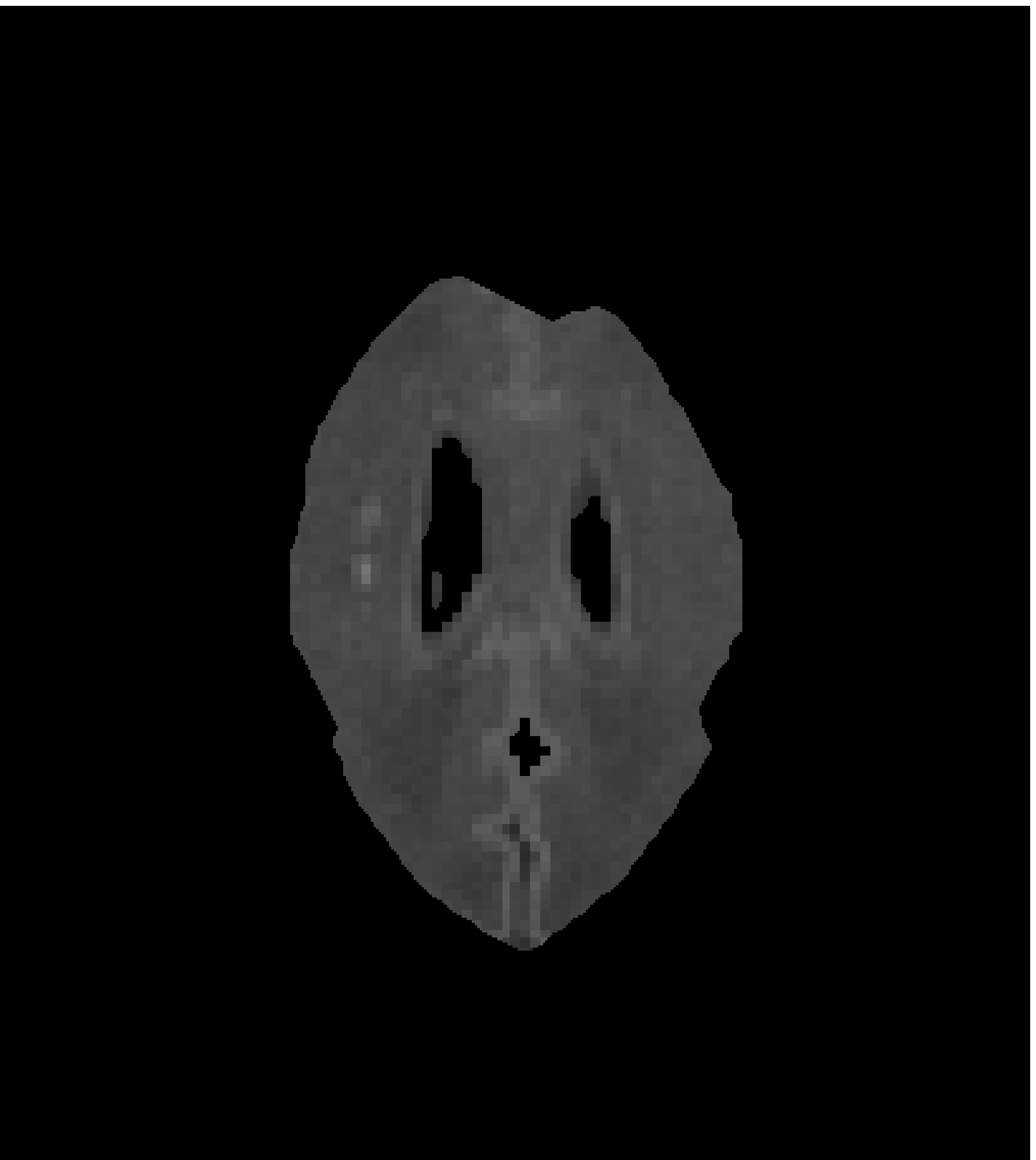}
\caption{Region mask $M_w$}\label{fig:contour-crop}
\end{subfigure}
\begin{subfigure}[b]{.24\linewidth}
\includegraphics[width=\linewidth]{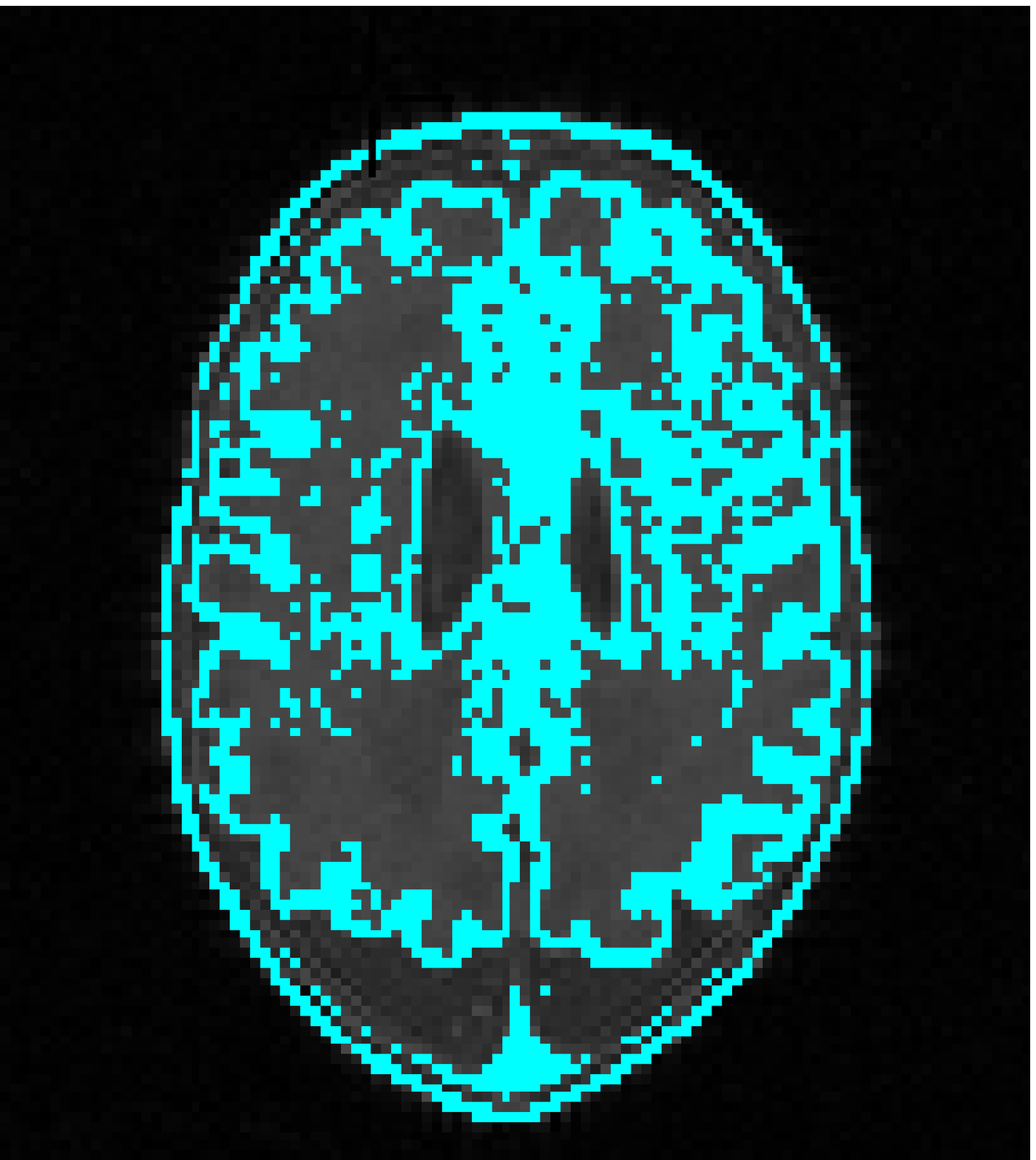}
\caption{Hyperintensities detection}\label{fig:hyper}
\end{subfigure}

\begin{subfigure}[b]{.24\linewidth}
\includegraphics[width=\linewidth]{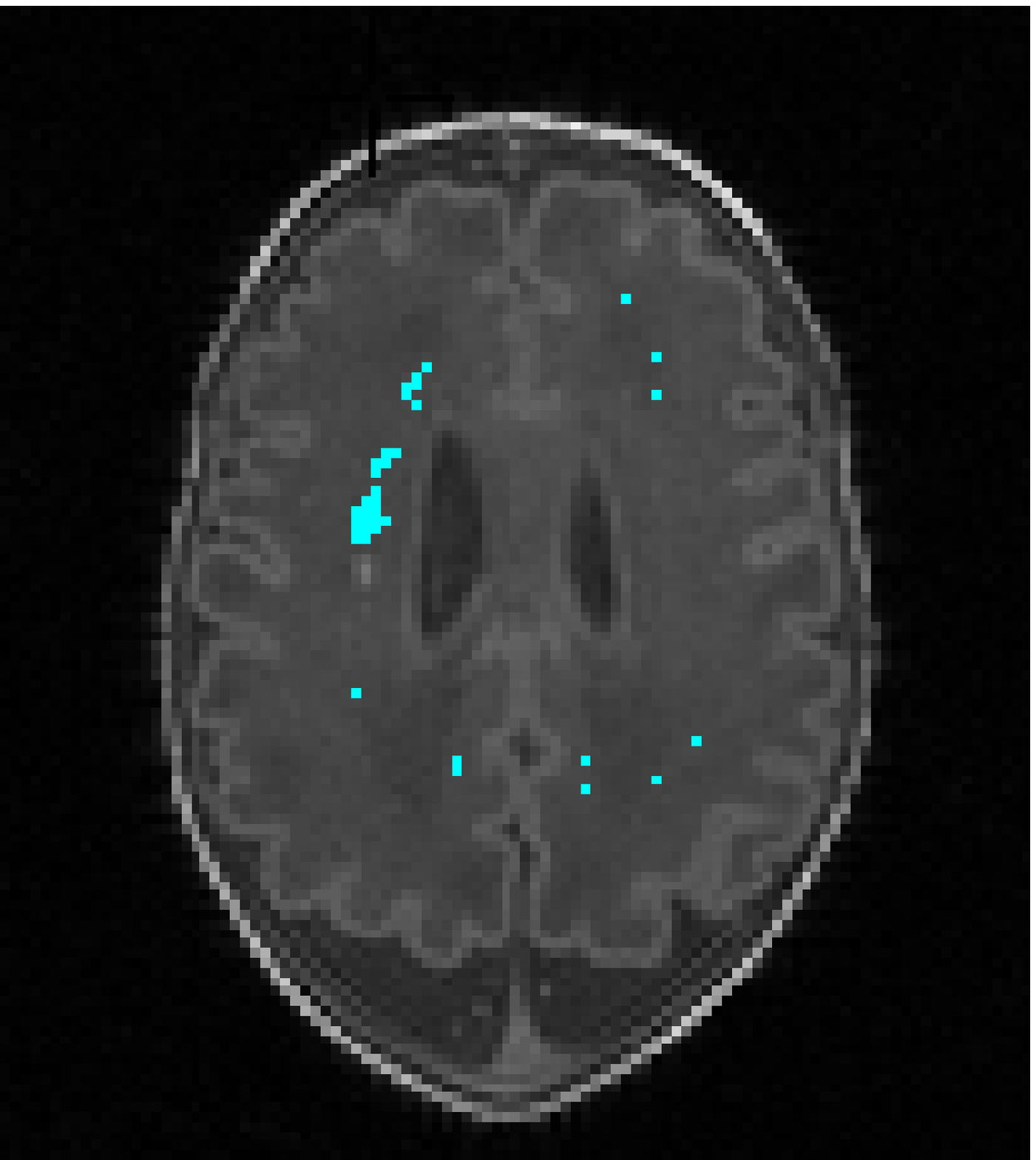}
\caption{Coarse detection includes false positives}\label{fig:coarse}
\end{subfigure}
\begin{subfigure}[b]{.24\linewidth}
\includegraphics[width=\linewidth]{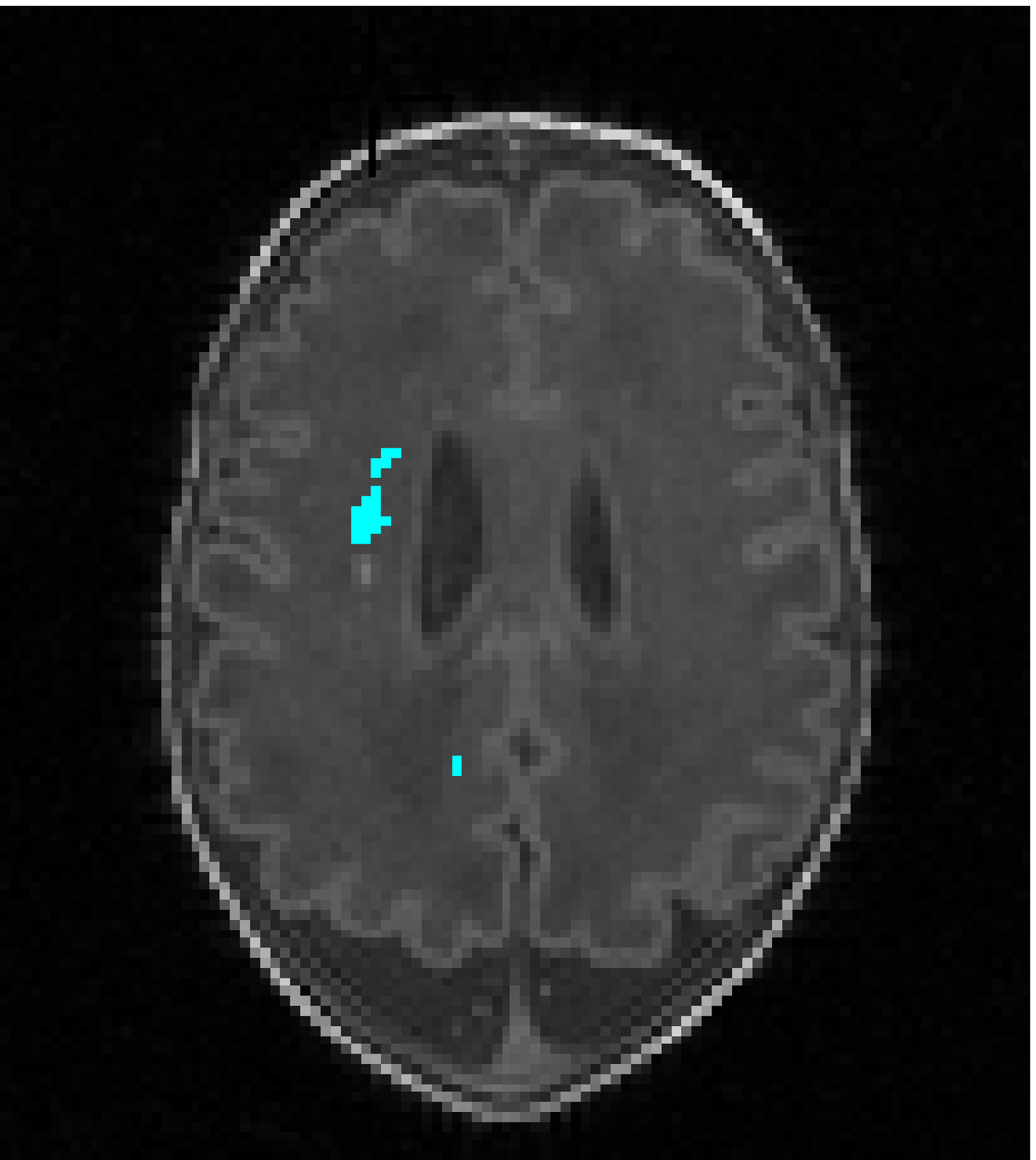}
\caption{Fine detection removes many false positives}\label{fig:fine}
\end{subfigure}
\caption{An illustration of the various steps in our proposed method.}
\label{fig:op-mains}
\end{figure}

\begin{figure}
    \centering
    \begin{subfigure}[b]{0.15\textwidth}
        \centering
        \includegraphics[width=\textwidth]{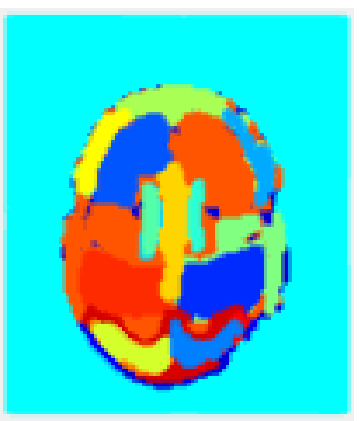}
    \end{subfigure}
    \hfill
    \begin{subfigure}[b]{0.15\textwidth}
        \centering
        \includegraphics[width=\textwidth]{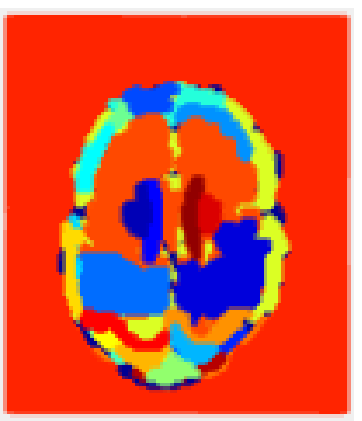}
    \end{subfigure}
    \hfill
    \begin{subfigure}[b]{0.15\textwidth}
        \centering
        \includegraphics[width=\textwidth]{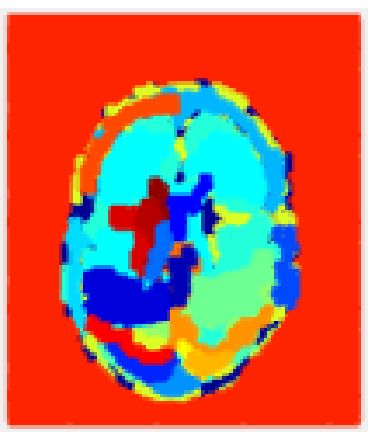}
    \end{subfigure}
    \hfill
    \begin{subfigure}[b]{0.15\textwidth}
        \centering
        \includegraphics[width=\textwidth]{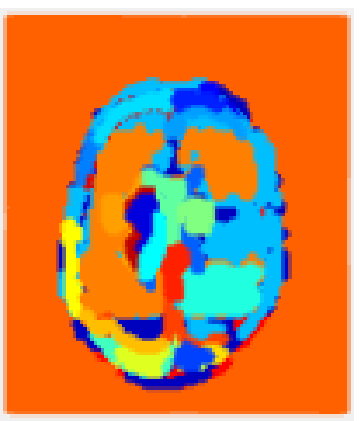}
    \end{subfigure}
    \hfill
    \begin{subfigure}[b]{0.15\textwidth}
        \centering
        \includegraphics[width=\textwidth]{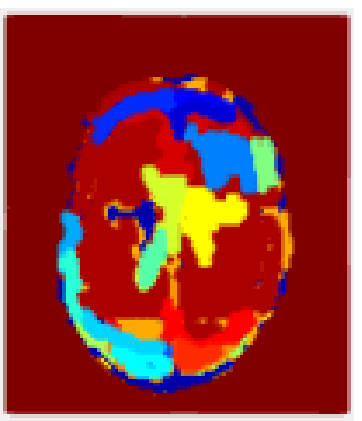}
    \end{subfigure}
    \vfill
    \begin{subfigure}[b]{0.15\textwidth}
        \centering
        \includegraphics[width=\textwidth]{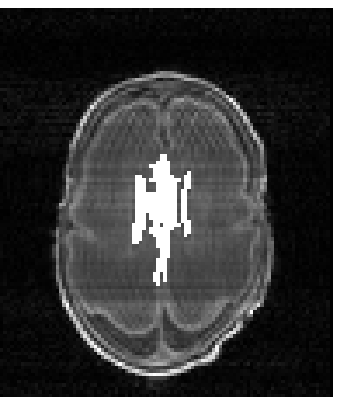}
    \end{subfigure}
    \hfill
    \begin{subfigure}[b]{0.15\textwidth}
        \centering
        \includegraphics[width=\textwidth]{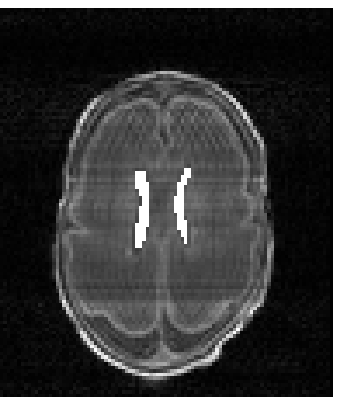}
    \end{subfigure}
    \hfill
    \begin{subfigure}[b]{0.15\textwidth}
        \centering
        \includegraphics[width=\textwidth]{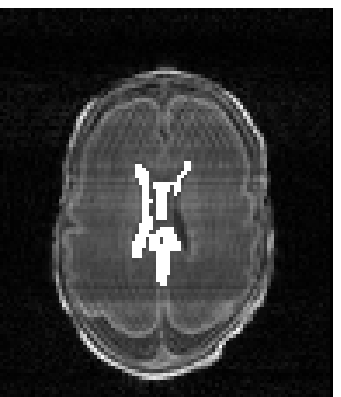}
    \end{subfigure}
    \hfill
    \begin{subfigure}[b]{0.15\textwidth}
        \centering
        \includegraphics[width=\textwidth]{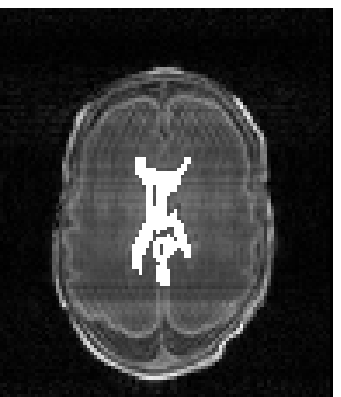}
    \end{subfigure}
    \hfill
    \begin{subfigure}[b]{0.15\textwidth}
        \centering
        \includegraphics[width=\textwidth]{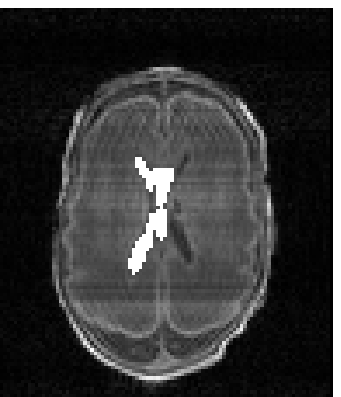}
    \end{subfigure}
    \caption{Result of blob detection (top row) and refinement (bottom row) to locate ventricles using the MSER algorithm on Slices 1 through 7. Each detected blob has been shown using a different colour for ease of visualization.}
    \label{fig:MSER}
\end{figure}

It should be noted that WMI were not present in the slices from the first subject. However, those slices present more challenging scenarios for the ventricle detection step, due to the particular shape of the ventricles in them, as compared to slices from the other two subjects. Thus, in Fig. \ref{fig:MSER} we show results of ventricle detection on slices from the first subject and demonstrate the full WMI detection process on a representative slice from the third subject in Fig. \ref{fig:op-mains}. Also, throughout this paper, we consider WMI detection on slices from the second and third subject only. For quantitative evaluation, we use the standard metrics: sensitivity and specificity, given in Eq. \ref{eq:sensitivity} and Eq. \ref{eq:specificity}. We compare our results with a recent work on WMI detection \cite{Cheng2015}. It should, however, be noted that, the set of pixels in each slice used for computing sensitivity and specificity are those belonging to the brain and not the background. Also, in case of \cite{Cheng2015}, significant variation in WMI detection performance in terms of sensitivity and specificity was noted by even slightly varying the upper threshold, $\mathcal{T}$ for marking the potential WMI boundaries after calculating the transition matrix for each subject. Thus, for each subject, the value of $\mathcal{T}$ has to be tuned separately to get the best performance for that subject, while using the method proposed in \cite{Cheng2015}. To cover these cases, we varied the value of of $\mathcal{T}$ for Subject-2 and Subject-3 to report the resulting variation in WMI detection performance for the method \cite{Cheng2015}. On the other hand, our proposed method does not have such issues as it does not use any parameters in the main detection phase. The method \cite{Cheng2015} has several other parameters in addition to $\mathcal{T}$. However, the most significant variations in its output resulted particularly from even minute changes in $\mathcal{T}$, and hence we specifically highlight this parameter. Also, note that our work focuses on WMI detection in MRIs of preterm neonate brains, specifically on WM hyper-intensities in T1 images. Since there is insufficient relevant research result in the literature to compare, we chose a few WM lesion detection methods to show that applying those methods to our application does not produce better results.

\begin{equation}\label{eq:sensitivity}
sensitivity = \frac{True\ Positives}{True\ Positives + False\ Negatives}
\end{equation}
\begin{equation}\label{eq:specificity}
specificity = \frac{True\ Negatives}{True\ Negatives + False\ Positives}
\end{equation}

	% Slices 69 to 75 of Subject-1 (0038) are referred to as slices "1 to 7"

In Fig. \ref{fig:op-mains} we illustrate the steps of our proposed method with a DICOM slice, and compare our WMI detection result with the expert-annotated ground truth. An input slice and ground truth are shown in Fig. \ref{fig:input-slice} and Fig. \ref{fig:ground-truth} respectively. The output of Ostu's method before and after the morphological hole filling operation is shown in Fig. \ref{fig:op-ostu} and Fig. \ref{fig:op-hole} respectively. GA returns the set of blobs most likely to constitute the ventricles, as shown in Fig. \ref{fig:ventricle}. However, it may be argued that for this particular input slice, the ventricles appear as relatively simple, regular shapes. Thus, we additionally show that the proposed ventricle detection approach works even for slices where the ventricles appear as more complex, irregular shapes, in Fig. \ref{fig:MSER}. However, these slices do not contain WMI and are thus not used henceforth in showing WMI detection. The detected contour around the ventricles is shown in Fig. \ref{fig:brain-crop}. True ventricles and the patches falsely detected as ventricles are represented as ``holes'' in Fig. \ref{fig:contour-crop}. Brain hyperintensities based on the ``Modified Z-score'' metric are shown in Fig. \ref{fig:hyper}. The result of coarse detection, after imposing the size and distance constraints, is shown in Fig. \ref{fig:coarse}. Comparing with the ground truth shown in Fig. \ref{fig:ground-truth}, we see that there is both true positive and false positive WM injuries. Many of these false positives are eliminated by our fine detection step, as shown in Fig. \ref{fig:fine}.

\subsection{Quantitative Results: WMI Detection Accuracy}

The sensitivity and specificity comparison of proposed method with the method in \cite{Cheng2015} is presented in Table \ref{tab:resultsvarpar}. Since we vary the value of $\mathcal{T}$ \cite{Cheng2015}, the scores reported for the method in \cite{Cheng2015} have been averaged over those obtained for individual values of $\mathcal{T}$ for each slice. It can be seen that the average sensitivity and specificity are higher in case of the proposed method as compared to the one in \cite{Cheng2015}, even though the proposed method does not segment the WM region. However, as we can see from Table \ref{tab:resultsvarpar}, the chief advantage of the proposed method  as compared to \cite{Cheng2015} lies in saving computational time by bypassing the full WM segmentation.

\begin{table}[]
\centering
\caption{Accuracy \& Execution Time comparison of Our Method with Method \cite{Cheng2015} by varying its parameter values}
\label{tab:resultsvarpar}
\begin{tabular}{@{}lllllll@{}}
\toprule
Average & \begin{tabular}[c]{@{}l@{}}Sensitivity\\ (Method \cite{Cheng2015})\end{tabular} & \begin{tabular}[c]{@{}l@{}}Sensitivity\\ (Proposed)\end{tabular} & \begin{tabular}[c]{@{}l@{}}Specificity\\ (Method \cite{Cheng2015})\end{tabular} & \begin{tabular}[c]{@{}l@{}}Specificity\\ (Proposed)\end{tabular} \\ \midrule
        & 47.90 & 49.77 & 99.23 & 99.49
\end{tabular}
\begin{tabular}{@{}lllllll@{}}
\toprule
Time(ms) & \begin{tabular}[c]{@{}l@{}}Per Slice\\ (Method \cite{Cheng2015})\end{tabular} & \begin{tabular}[c]{@{}l@{}}Per Slice\\ (Proposed)\end{tabular} & \begin{tabular}[c]{@{}l@{}}Per Volume\\ (Method \cite{Cheng2015})\end{tabular} & \begin{tabular}[c]{@{}l@{}}Per Volume\\ (Proposed)\end{tabular} \\ \midrule
         & 4150 & 500 & 796800 & 41500
\end{tabular}
\end{table}

In order to assess the effectiveness of our size and distance constraint features, we also compute the average sensitivity and specificity scores across all slices for the proposed method in each of the following scenarios:
\begin{enumerate}
  \item with both constraints;
  \item with no constraint;
  \item with only size constraint; and,
  \item with only distance constraint;
\end{enumerate}
The above results are aggregated and presented in Table \ref{tab:results_effective}.

\begin{table}[]
\centering
\caption{Effectiveness of the Size and Distance Constraints on Proposed Method}
\label{tab:results_effective}
\begin{tabular}{@{}lllllll@{}}
\toprule
Constraint & \begin{tabular}[c]{@{}l@{}}Sensitivity \\(average)\end{tabular} & \begin{tabular}[c]{@{}l@{}}True Positives \\(average)\end{tabular} & \begin{tabular}[c]{@{}l@{}}Specificity \\(average)\end{tabular} & \begin{tabular}[c]{@{}l@{}}False Positives \\(average)\end{tabular} \\ \midrule
Both     & 49.77  & 36.71   & 99.49  & 501.86  \\
None     & 100    & 110.71  & 56.70  & 32271   \\
Size     & 49.77  & 36.71   & 97.82  & 1657.57 \\
Distance & 49.77  & 36.71   & 99.30  & 501.86  \\ \bottomrule
\end{tabular}
\end{table}

We conduct another experiment by restricting the minimum allowed lesion size, in terms of number of voxels, and the effect of this on WMI detection accuracy is presented in Table \ref{tab:results_md}. The motivation behind this experiment is that sometimes random White Matter intensity variations may occur simply as a result of noise, and not the presence of actual White Matter Injury, as argued by authors in \cite{Cabezas2014}.

\begin{table}[]
\centering
\caption{Proposed Method's Reduced performance with Minimum Lesion Size constraint \cite{Cabezas2014}}
\label{tab:results_md}
\begin{tabular}{@{}lllllll@{}}
\toprule
Minimum Size & \begin{tabular}[c]{@{}l@{}}Sensitivity \\(average)\end{tabular} & \begin{tabular}[c]{@{}l@{}}True Positives \\(average)\end{tabular} & \begin{tabular}[c]{@{}l@{}}Specificity \\(average)\end{tabular} & \begin{tabular}[c]{@{}l@{}}False Positives \\(average)\end{tabular} \\ \midrule
Min.Size N/A    & 49.77  & 36.71   & 99.49  & 501.86  \\
Min.Size = 100  & 21.20  & 27.57   & 99.87  & 99.71   \\
Min.Size = 150  & 6.92   & 26.28   & 99.94  & 49.71   \\
Min.Size = 250  & 0      & 0       & 100    & 0       \\ \bottomrule
\end{tabular}
\end{table}

\subsection{Quantitative Results: Execution Time Performance}

For time performance, the average (serial) per-slice execution time of the proposed method is around $500$ milliseconds on a Ubuntu 14.04 PC with 16 GB RAM and an Intel Core \textit{i}7-4790 3.60 GHz CPU. Note that for the per-volume execution time to process a DICOM stack of 192 slices, the proposed method takes less than $500 \times 192$ milliseconds. This is because our method first performs coarse detection on all slices (which takes about 210 milliseconds per slice), benefiting the subsequent fine detection. The fine detection step focuses on a smaller set of WMI candidates and takes less than 6 milliseconds per slice. Thus, the total time taken to process the entire DICOM volume of 192 slices is 41460 milliseconds, or $\approx$ 41.5 seconds.

In comparison, the average per slice execution time of the segmentation-based preterm WMI detection method \cite{Cheng2015} is around 3.75 seconds (segmentation) + 0.4 seconds (detection) = 4.15 seconds (total). Its per volume execution time is 796.8 seconds = 13 minutes and 16.8 seconds. Also, note that the earlier method \cite{Cheng2015} performs only coarse detection without considering adjacent slices. This is because most of the execution time is taken up by the segmentation phase, whereas our proposed method bypasses segmentation, achieving better time performance without compromising accuracy.

Thus, from a comparison of the WMI detection accuracy and execution time of proposed method and our earlier work \cite{Cheng2015} shown in Table \ref{tab:resultsvarpar}, we can conclude that our new method greatly reduces the execution time, but still manages to improve WMI detection accuracy compared to our old method.

\subsection{Qualitative Results}

\begin{figure}
\centering
	\begin{subfigure}[b]{0.15\textwidth}
    	\centering
    	\includegraphics[width=\textwidth]{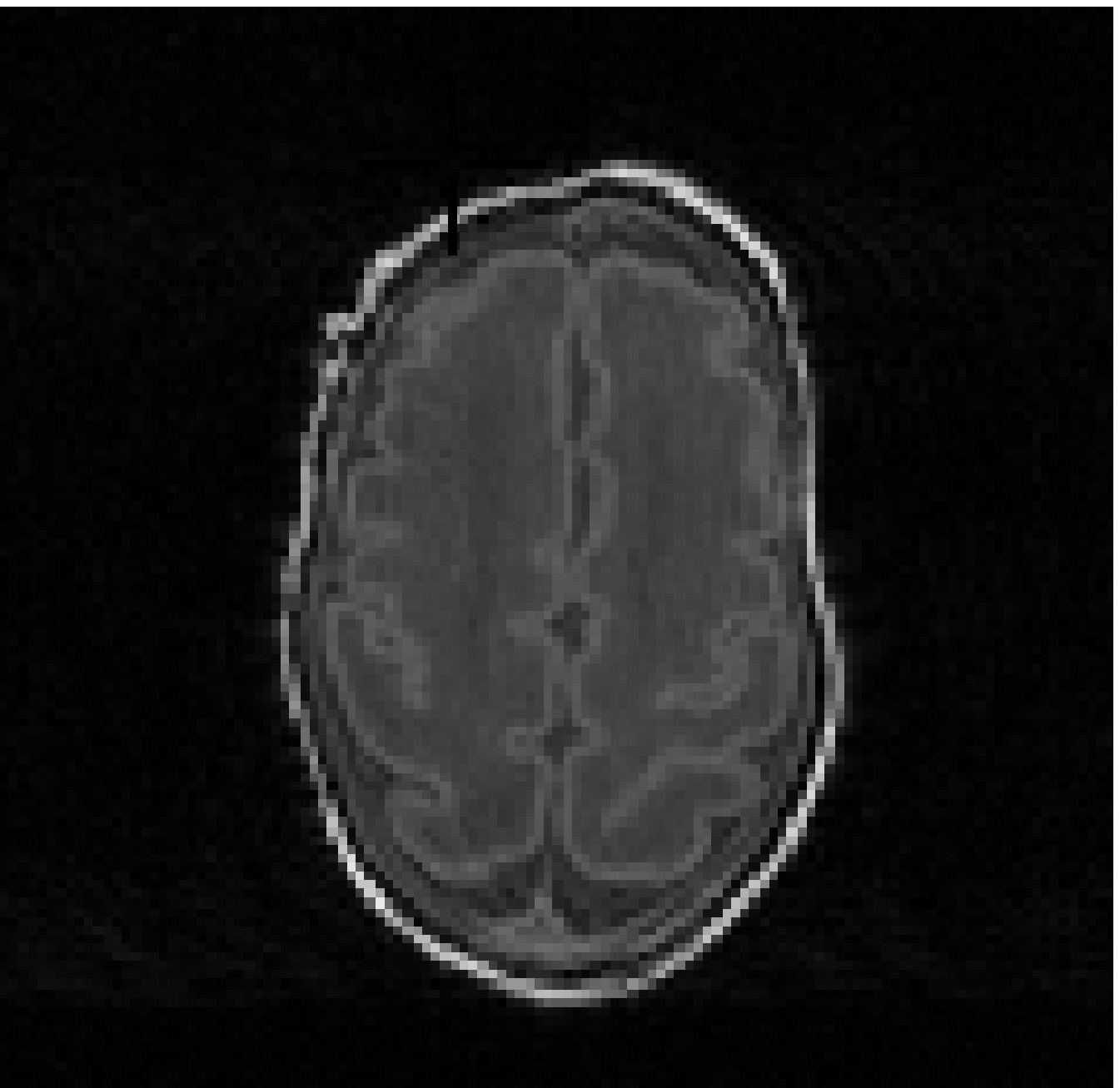}
    	\caption{Input Slice}
    	\label{fig:Slice_24_5}
	\end{subfigure}
	\begin{subfigure}[b]{0.15\textwidth}
    	\centering
    	\includegraphics[width=\textwidth]{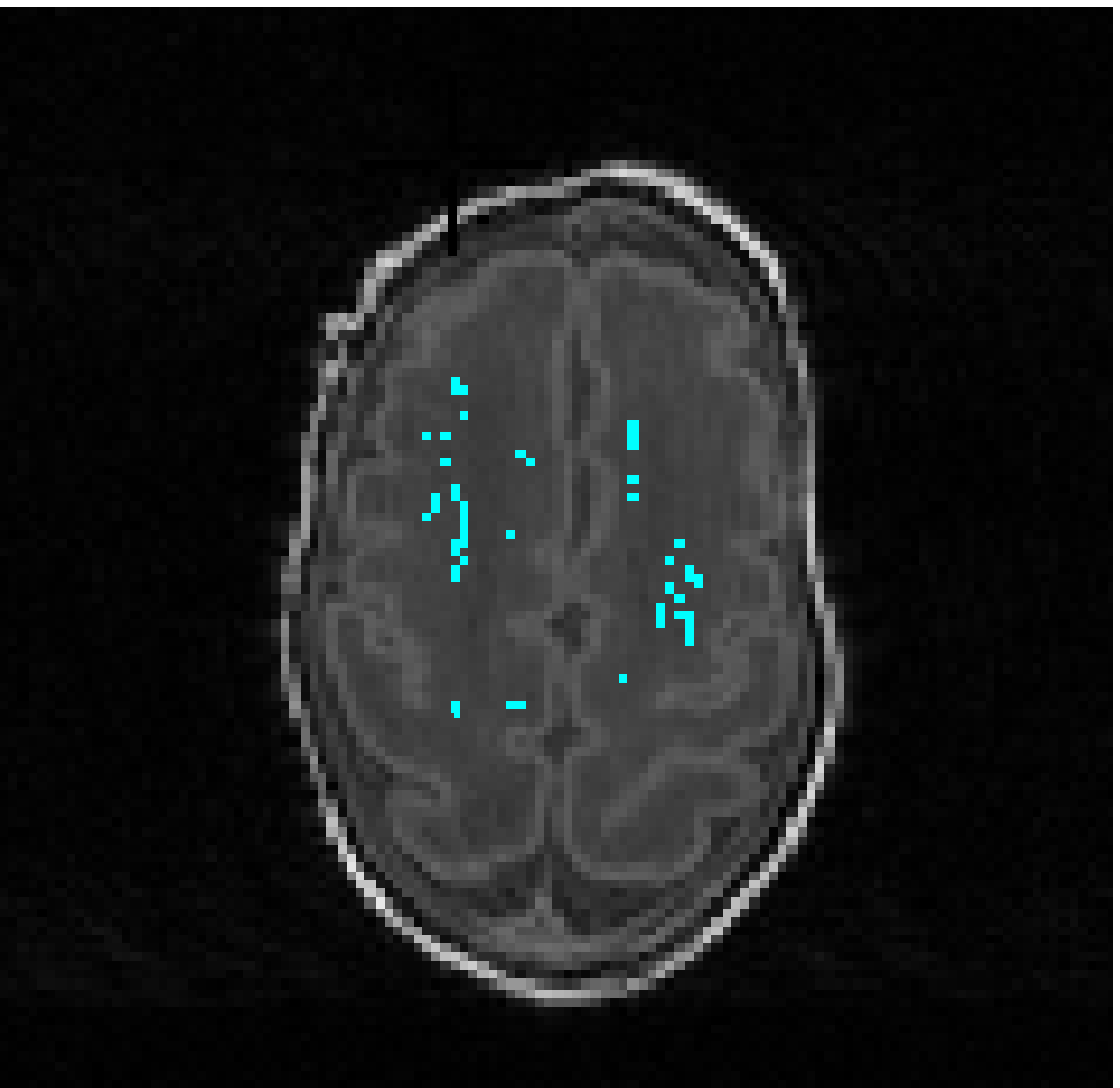}
    	\caption{Our Method}
    	\label{fig:Proposed_24_5}
	\end{subfigure}
	\begin{subfigure}[b]{0.15\textwidth}
    	\centering
    	\includegraphics[width=\textwidth]{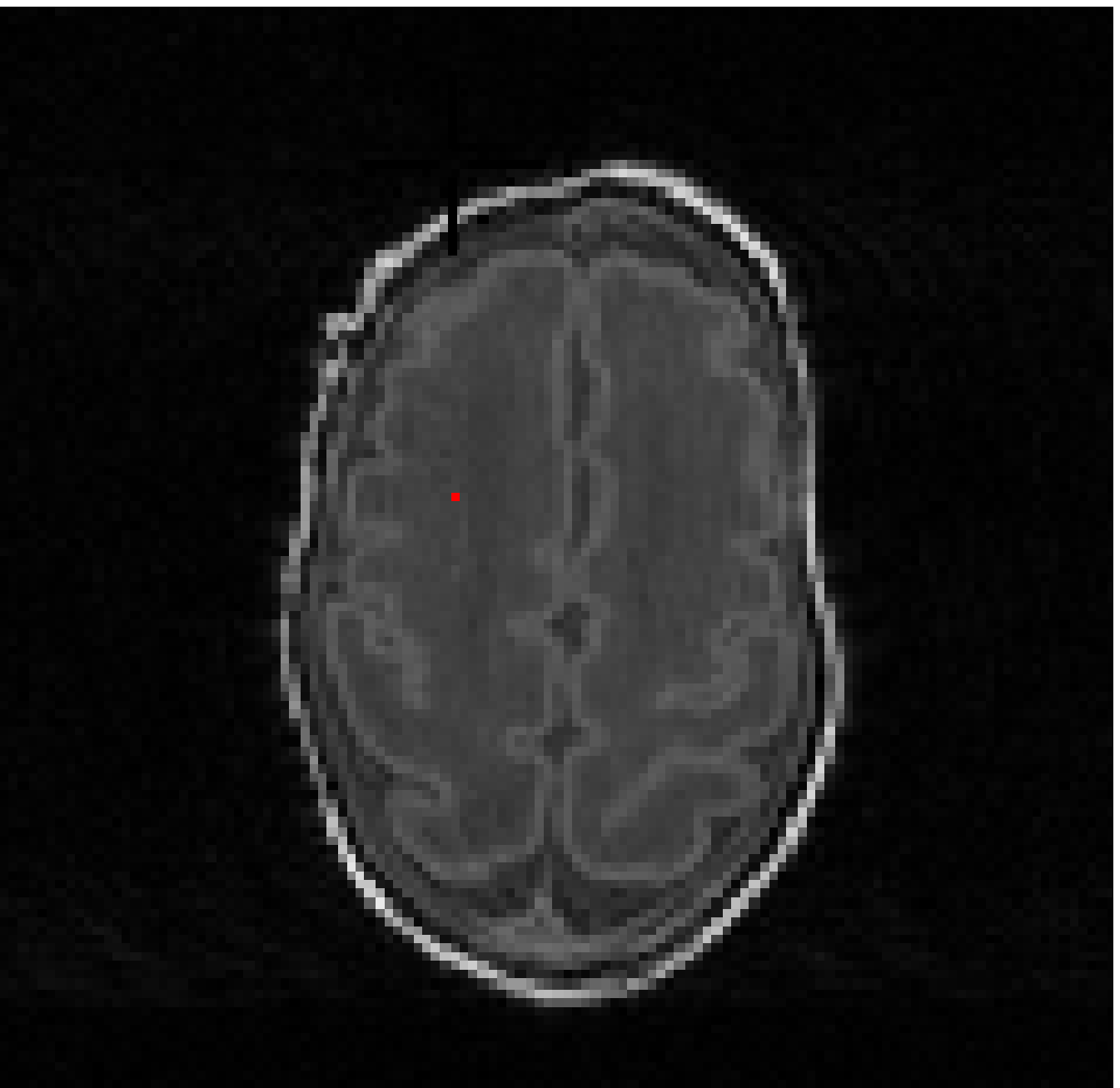}
    	\caption{Ground Truth}
    	\label{fig:Ground_24_5}
	\end{subfigure}
	\hfill
	\begin{subfigure}[b]{0.15\textwidth}
    	\centering
    	\includegraphics[width=\textwidth]{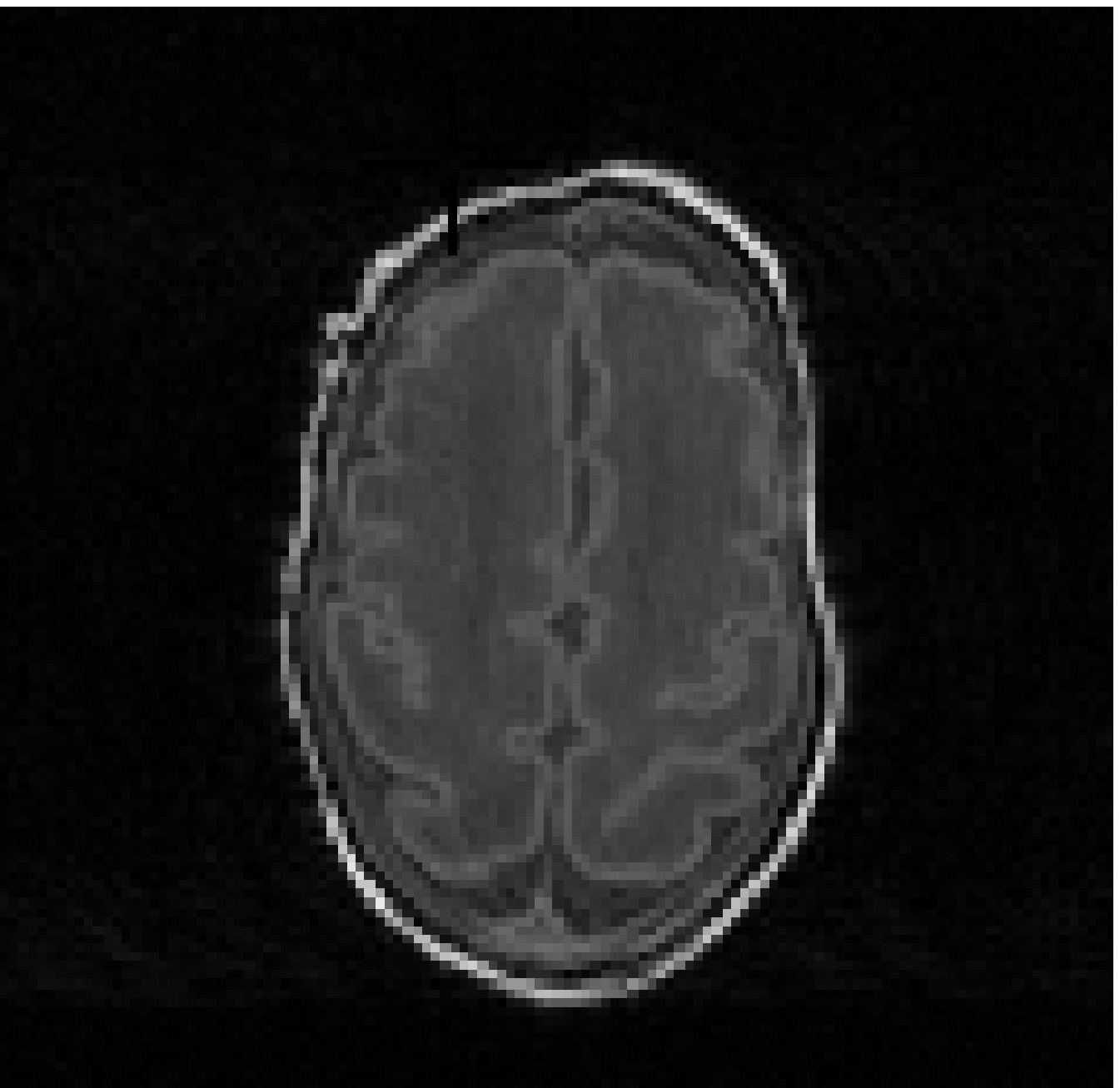}
    	\caption{$\mathcal{T}=0.005$ \cite{Cheng2015}}
    	\label{fig:Prof_24_5_005}
    \end{subfigure}
    \hfill
    \begin{subfigure}[b]{0.15\textwidth}
    	\centering
    	\includegraphics[width=\textwidth]{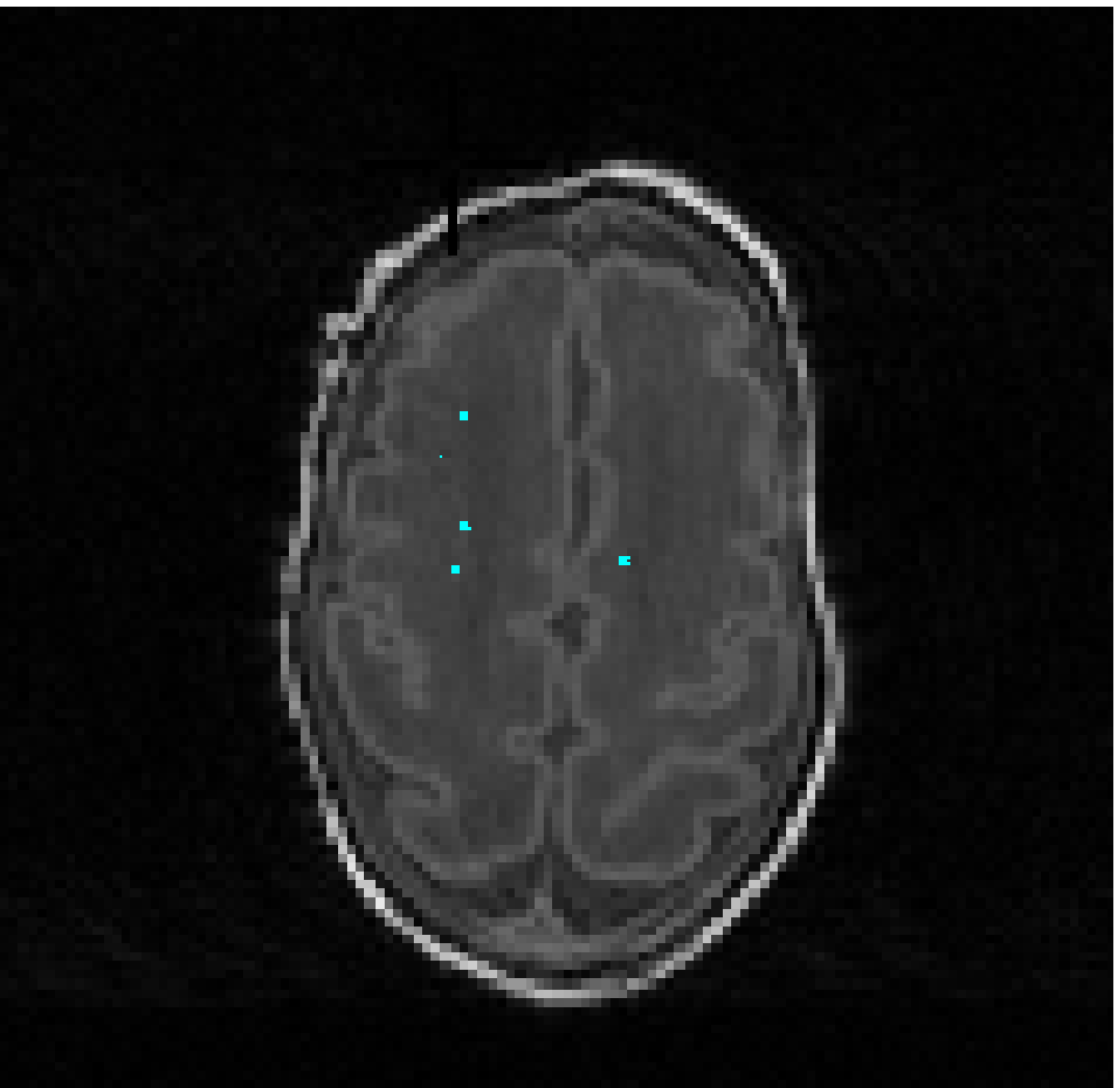}
    	\caption{$\mathcal{T}=0.015$}
    	\label{fig:Prof_24_5_015}
    \end{subfigure}
    \begin{subfigure}[b]{0.15\textwidth}
    	\centering
    	\includegraphics[width=\textwidth]{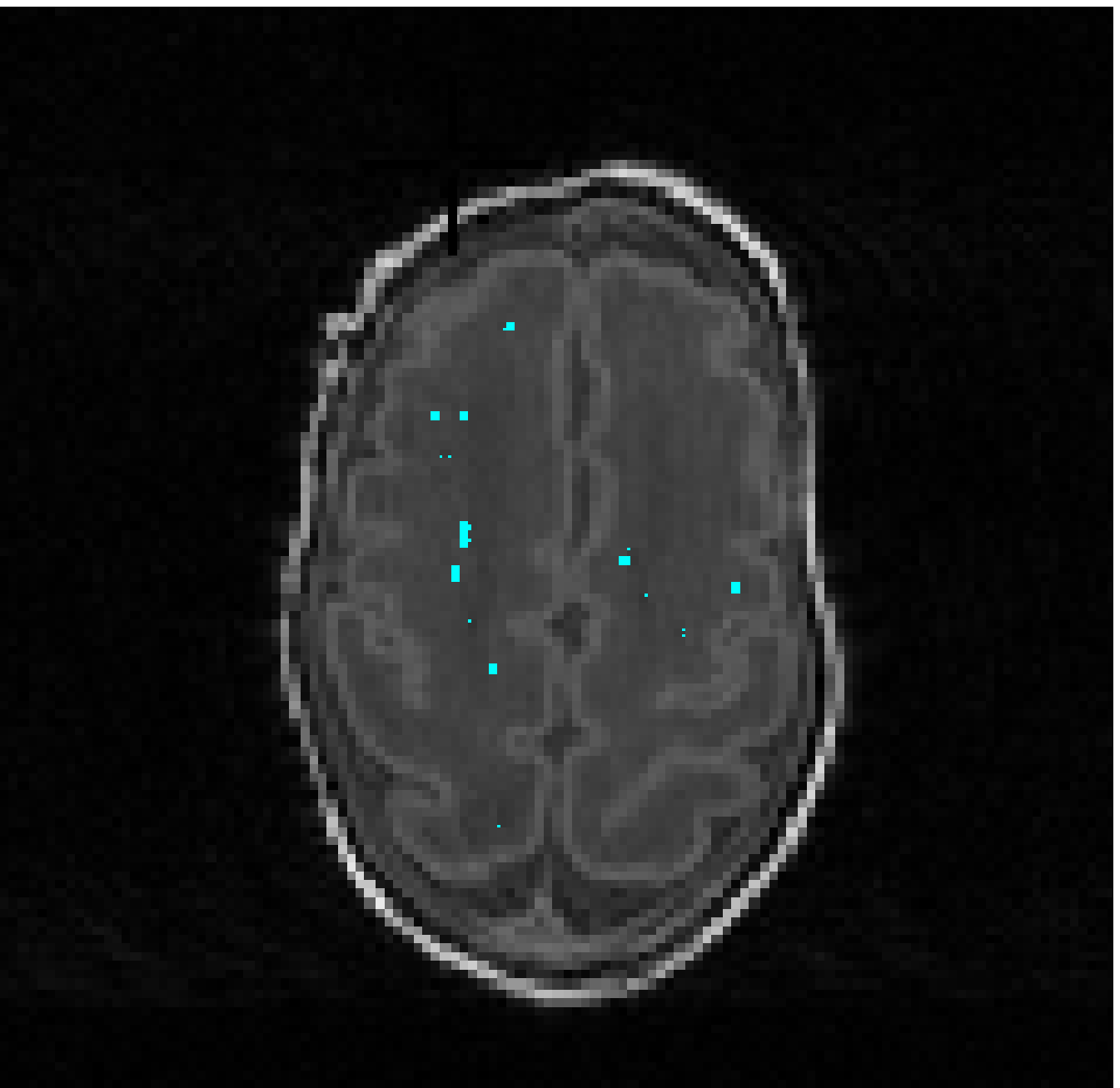}
    	\caption{$\mathcal{T}=0.025$}
    	\label{fig:Prof_24_5_025}
	\end{subfigure}
	\begin{subfigure}[b]{0.15\textwidth}
    	\centering
    	\includegraphics[width=\textwidth]{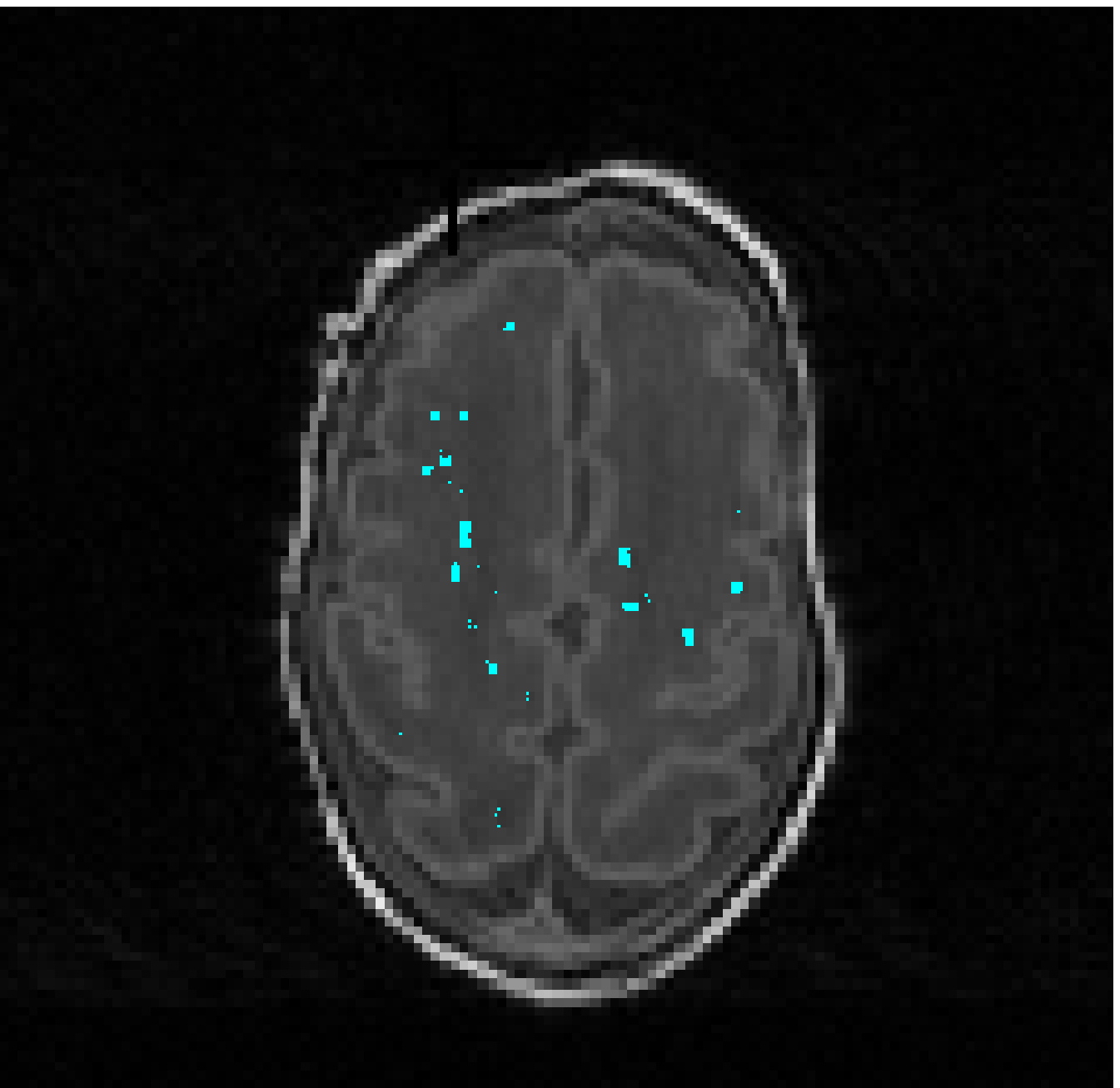}
    	\caption{$\mathcal{T}=0.035$}
    	\label{fig:Prof_24_5_035}
	\end{subfigure}
	\begin{subfigure}[b]{0.15\textwidth}
    	\centering
    	\includegraphics[width=\textwidth]{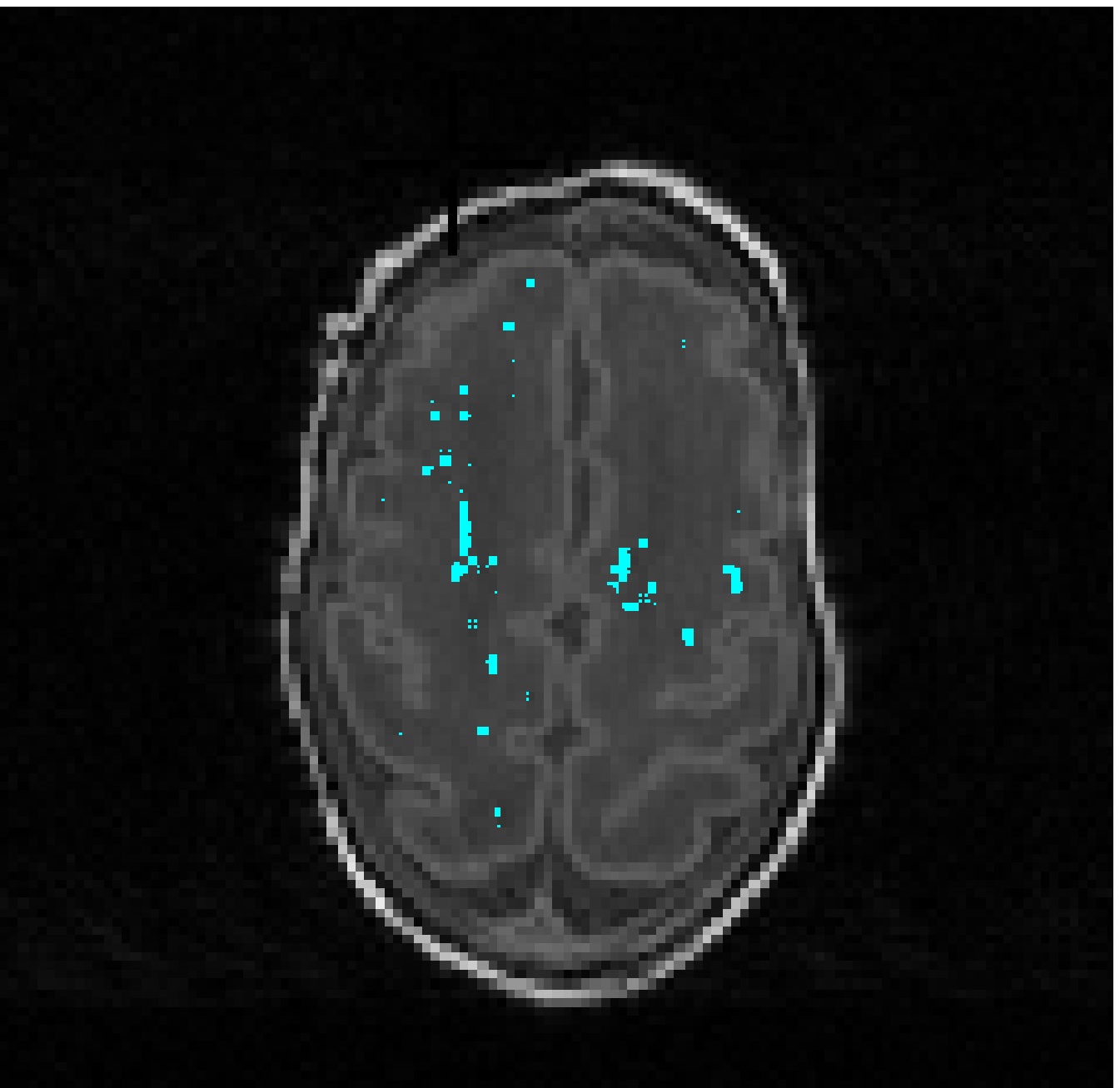}
    	\caption{$\mathcal{T}=0.045$}
    	\label{fig:Prof_24_5_045}
	\end{subfigure}
	\begin{subfigure}[b]{0.15\textwidth}
    	\centering
    	\includegraphics[width=\textwidth]{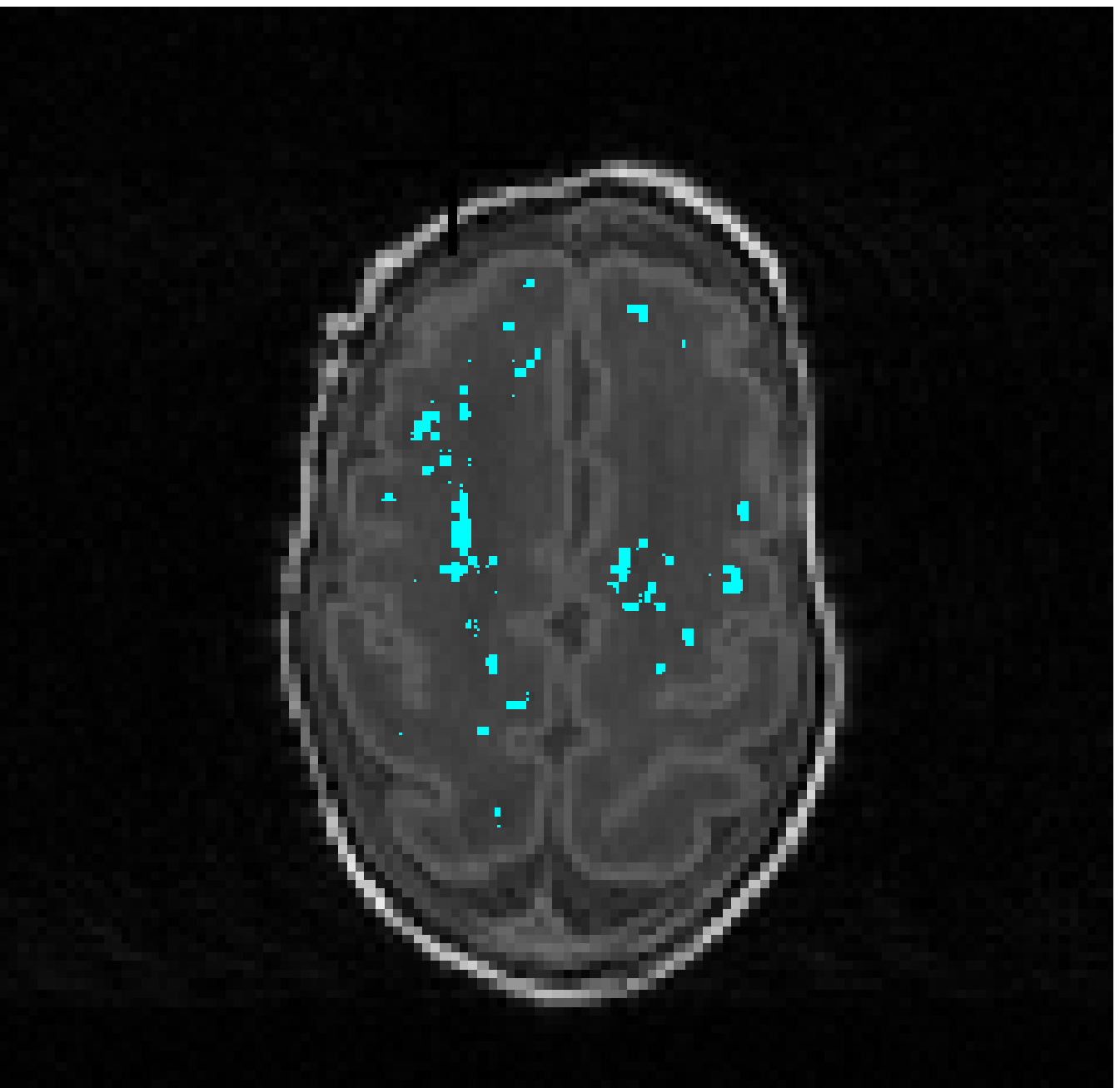}
    	\caption{$\mathcal{T}=0.055$}
    	\label{fig:Prof_24_5_055}
	\end{subfigure}
\caption{Comparison of Proposed Method and \cite{Cheng2015} using Slice 9.}
\label{fig:compare_24_5}
\end{figure}

\begin{figure}
\centering
	\begin{subfigure}[b]{0.15\textwidth}
    	\centering
    	\includegraphics[width=\textwidth]{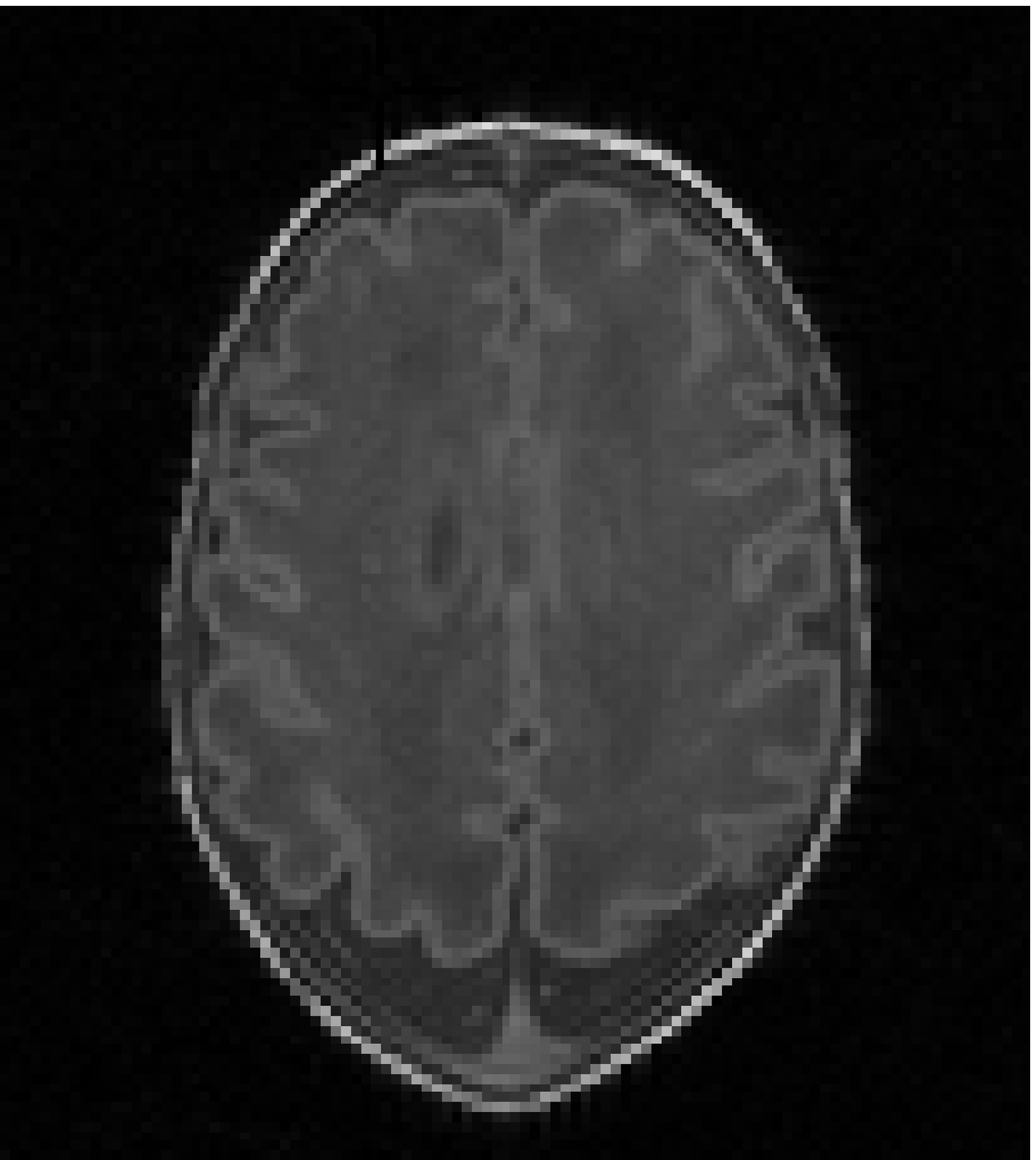}
    	\caption{Input Slice}
    	\label{fig:Slice_42_2}
	\end{subfigure}
	\begin{subfigure}[b]{0.15\textwidth}
    	\centering
    	\includegraphics[width=\textwidth]{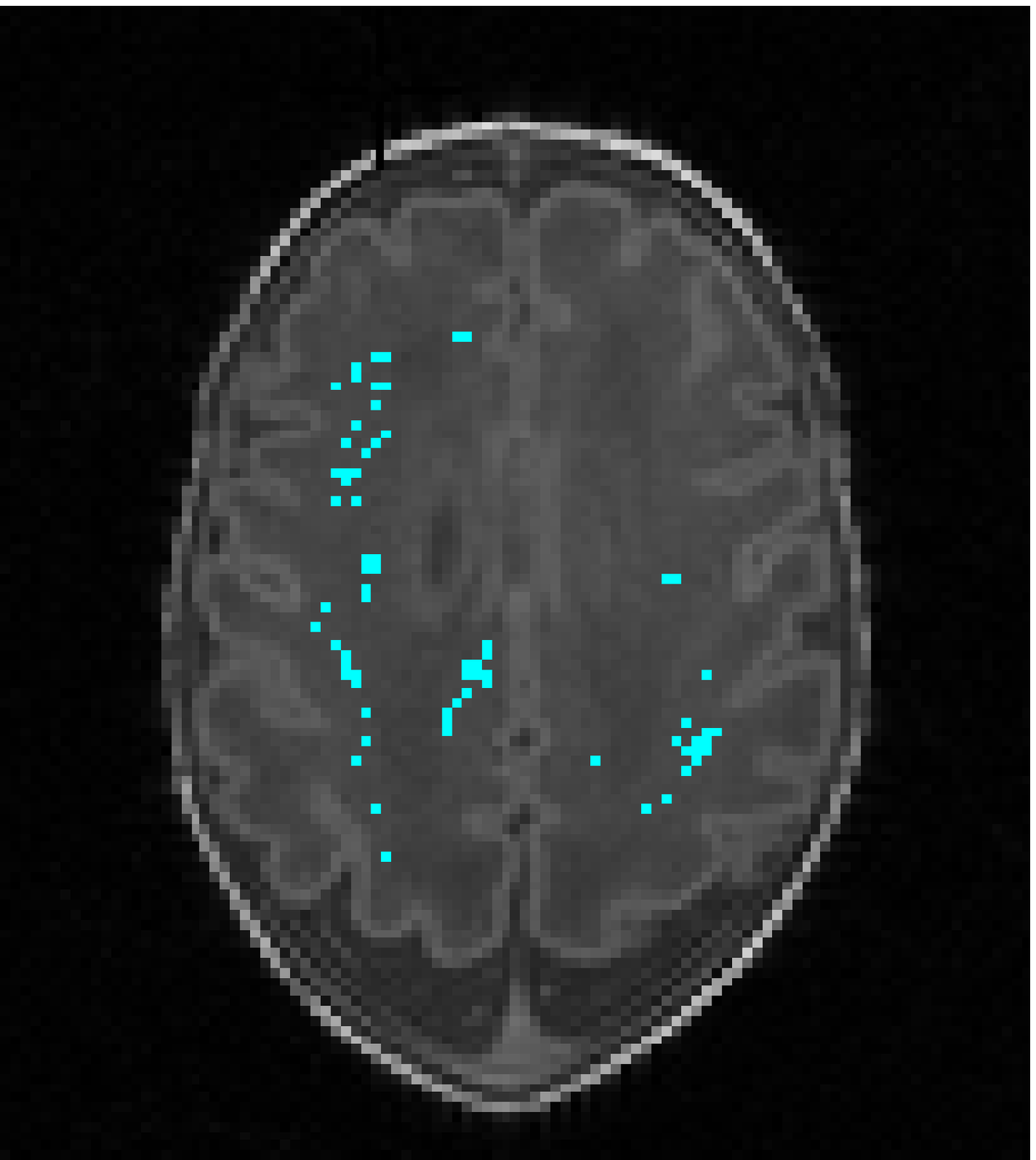}
    	\caption{Our Method}
    	\label{fig:Proposed_42_2}
	\end{subfigure}
	\begin{subfigure}[b]{0.15\textwidth}
    	\centering
    	\includegraphics[width=\textwidth]{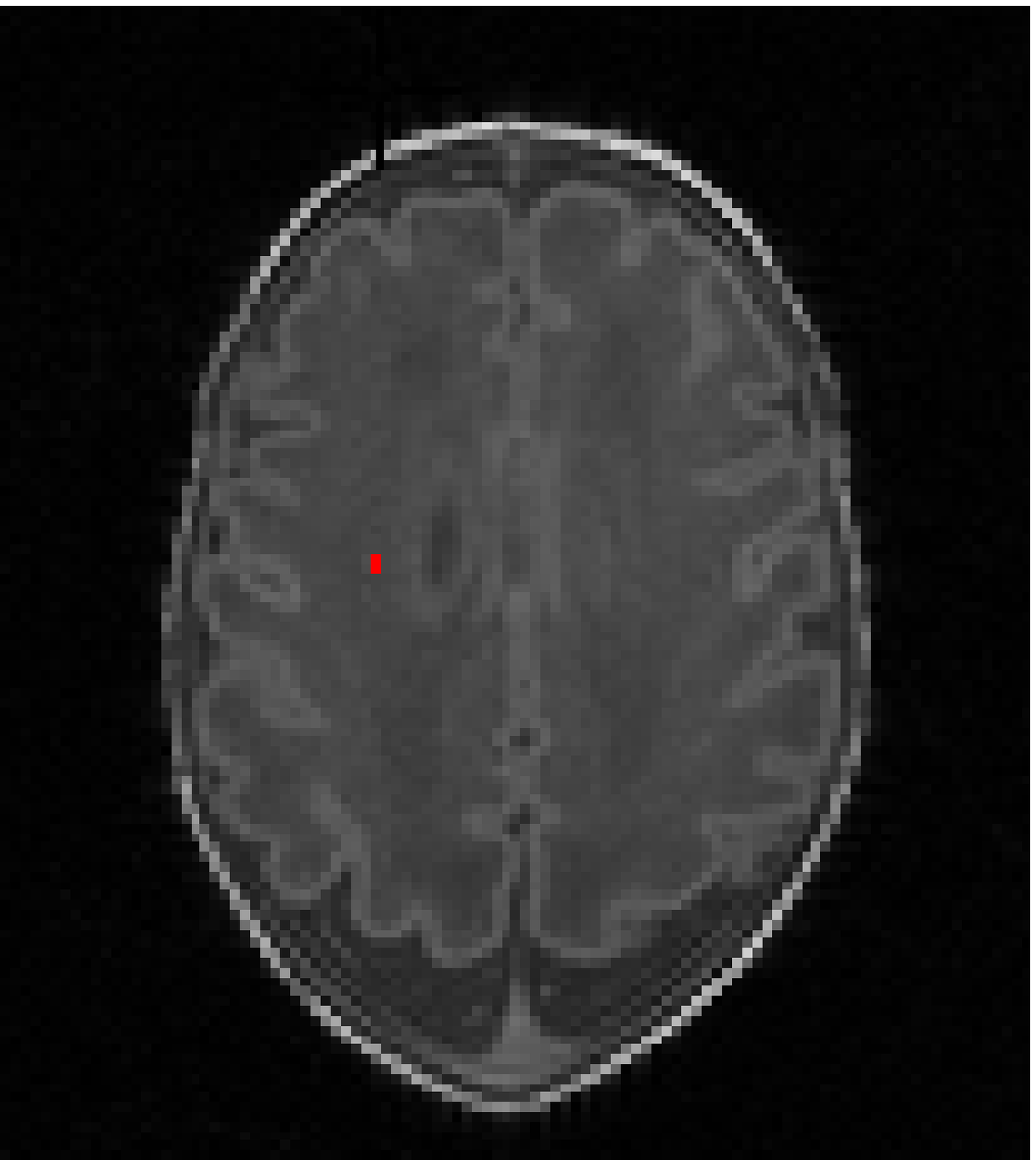}
    	\caption{Ground Truth}
    	\label{fig:Ground_42_2}
	\end{subfigure}
	\hfill
	\begin{subfigure}[b]{0.15\textwidth}
    	\centering
    	\includegraphics[width=\textwidth]{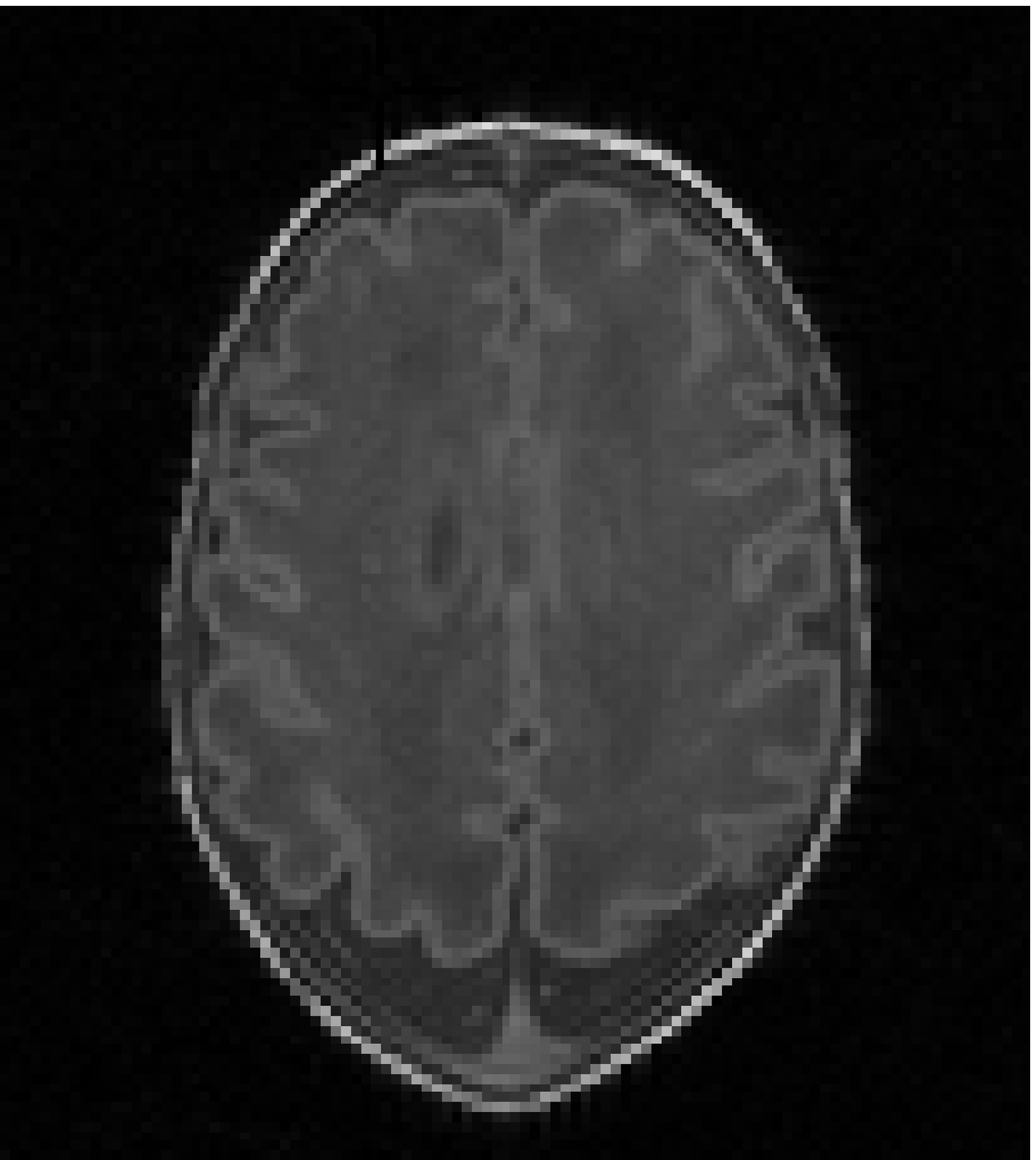}
    	\caption{$\mathcal{T}=0.005$ \cite{Cheng2015}}
    	\label{fig:Prof_42_2_005}
    \end{subfigure}
    \hfill
    \begin{subfigure}[b]{0.15\textwidth}
    	\centering
    	\includegraphics[width=\textwidth]{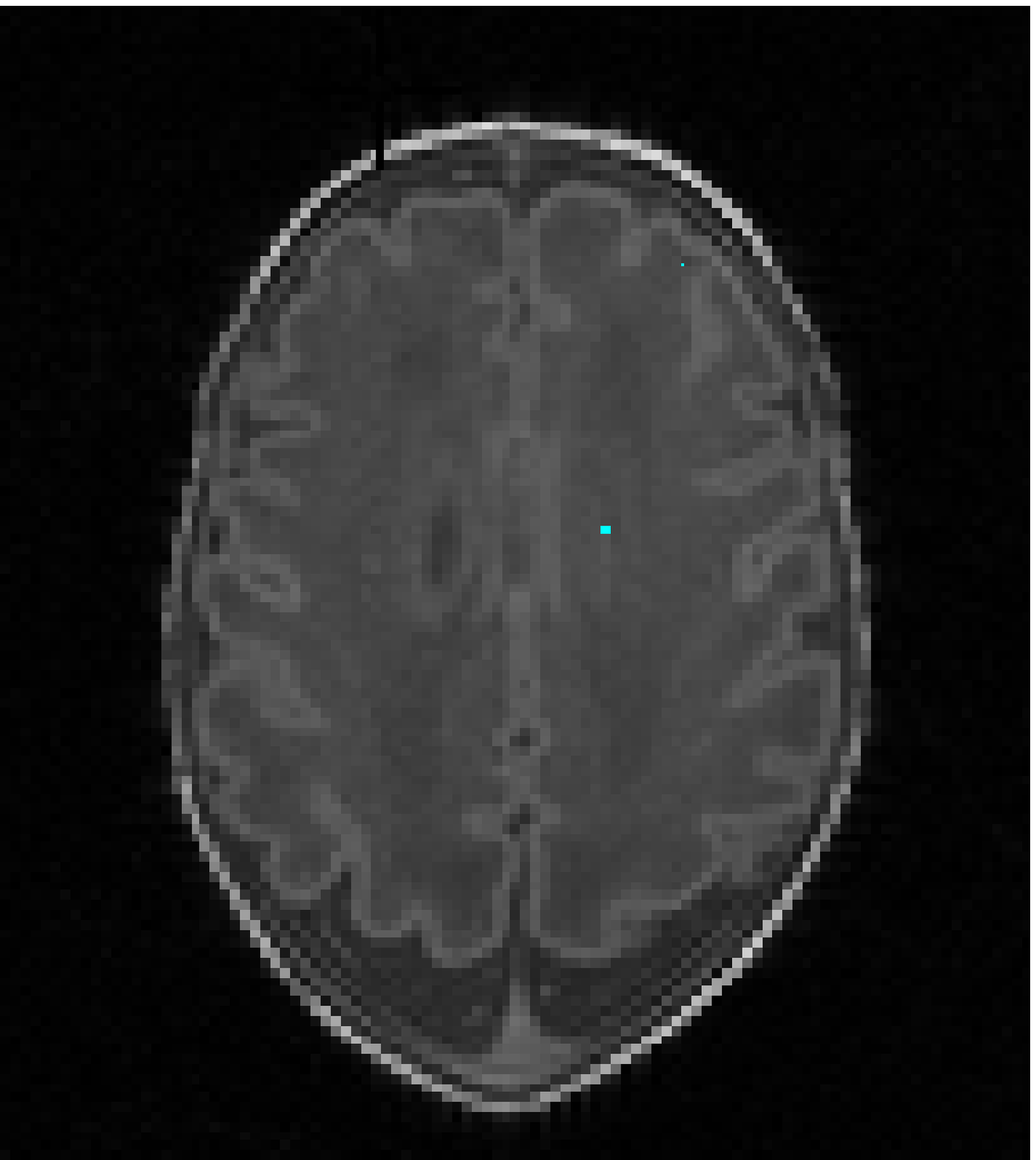}
    	\caption{$\mathcal{T}=0.015$}
    	\label{fig:Prof_42_2_015}
    \end{subfigure}
    \begin{subfigure}[b]{0.15\textwidth}
    	\centering
    	\includegraphics[width=\textwidth]{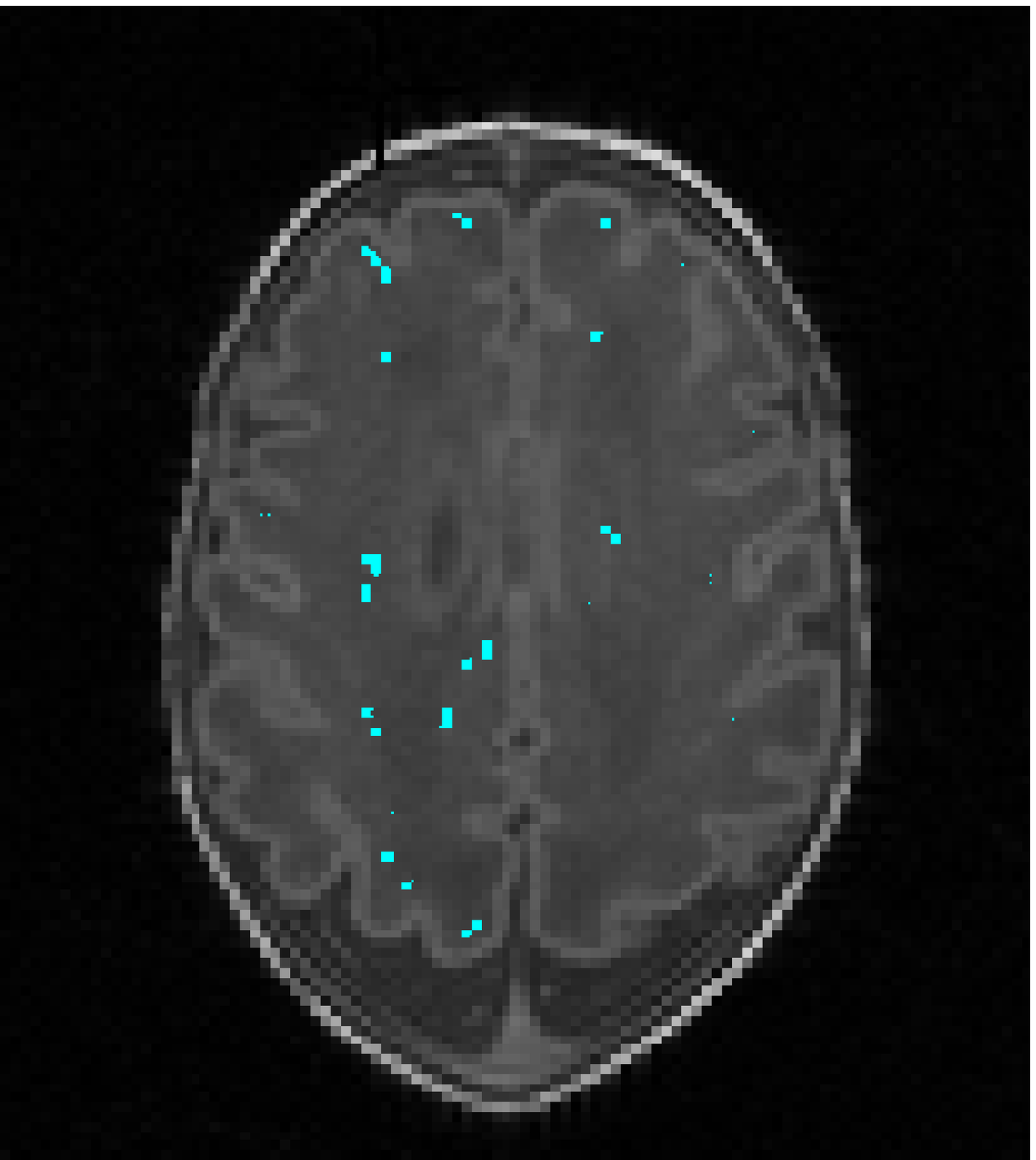}
    	\caption{$\mathcal{T}=0.025$}
    	\label{fig:Prof_42_2_025}
	\end{subfigure}
	\begin{subfigure}[b]{0.15\textwidth}
    	\centering
    	\includegraphics[width=\textwidth]{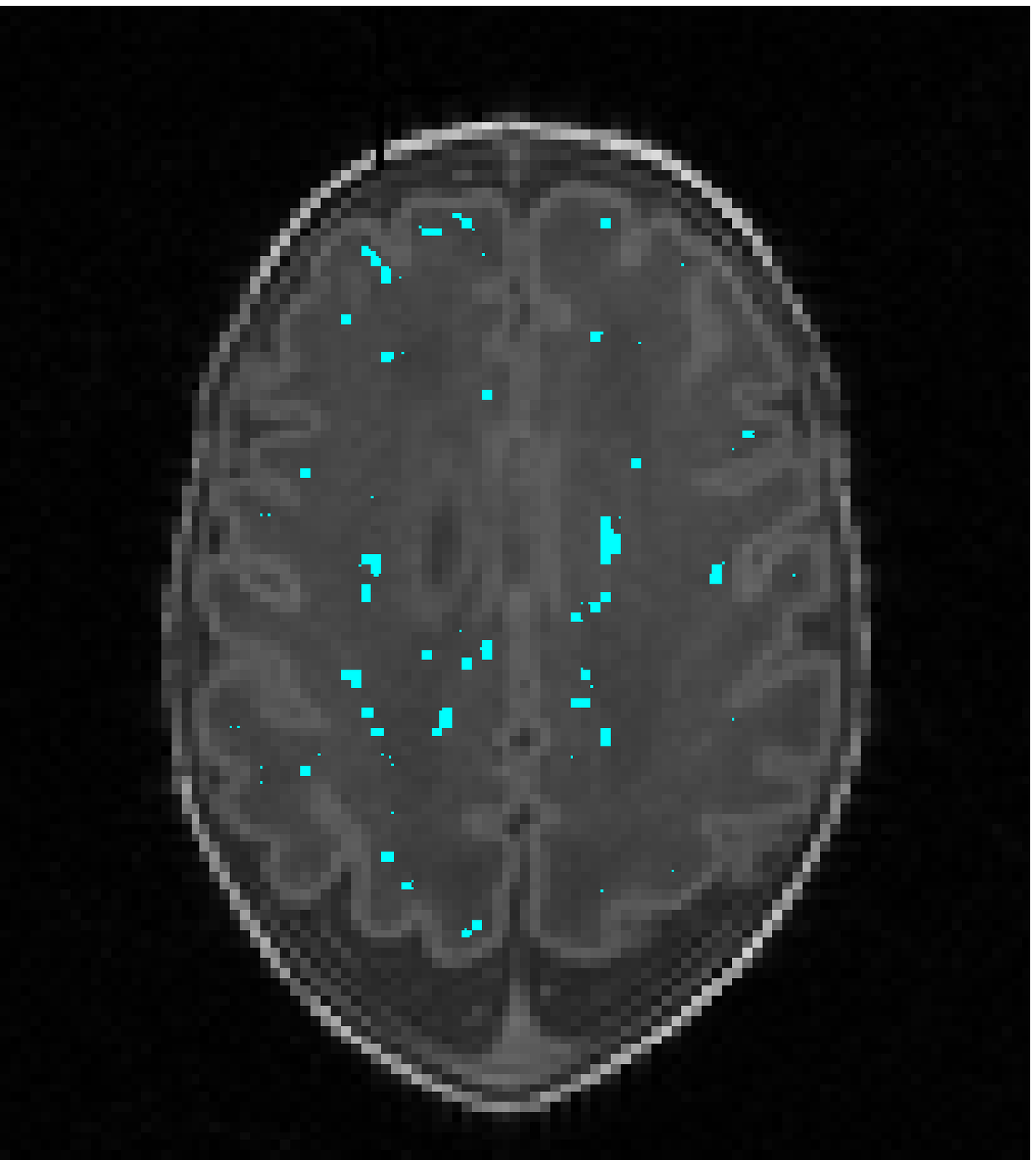}
    	\caption{$\mathcal{T}=0.035$}
    	\label{fig:Prof_42_2_035}
	\end{subfigure}
	\begin{subfigure}[b]{0.15\textwidth}
    	\centering
    	\includegraphics[width=\textwidth]{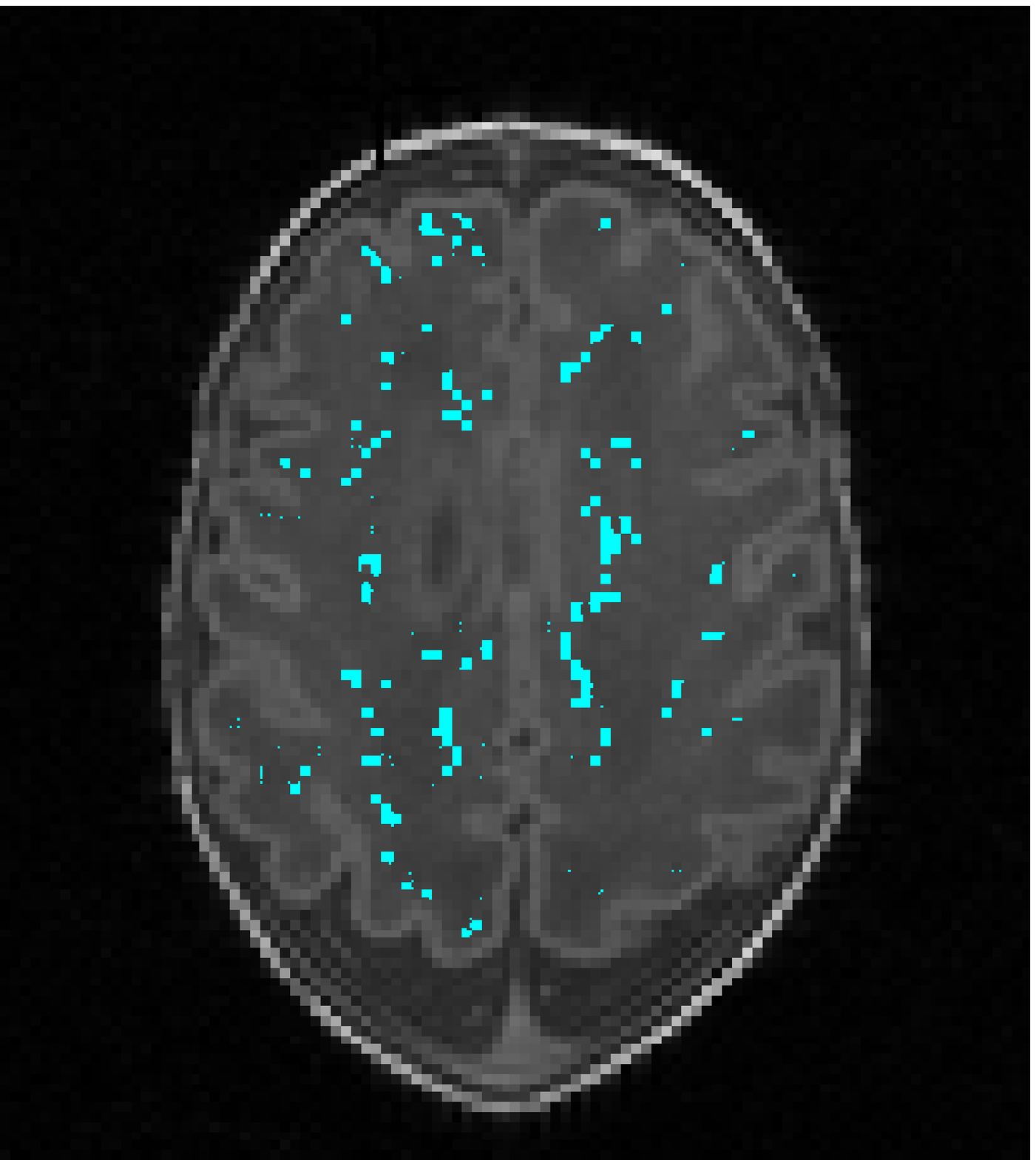}
    	\caption{$\mathcal{T}=0.045$}
    	\label{fig:Prof_42_2_045}
	\end{subfigure}
	\begin{subfigure}[b]{0.15\textwidth}
    	\centering
    	\includegraphics[width=\textwidth]{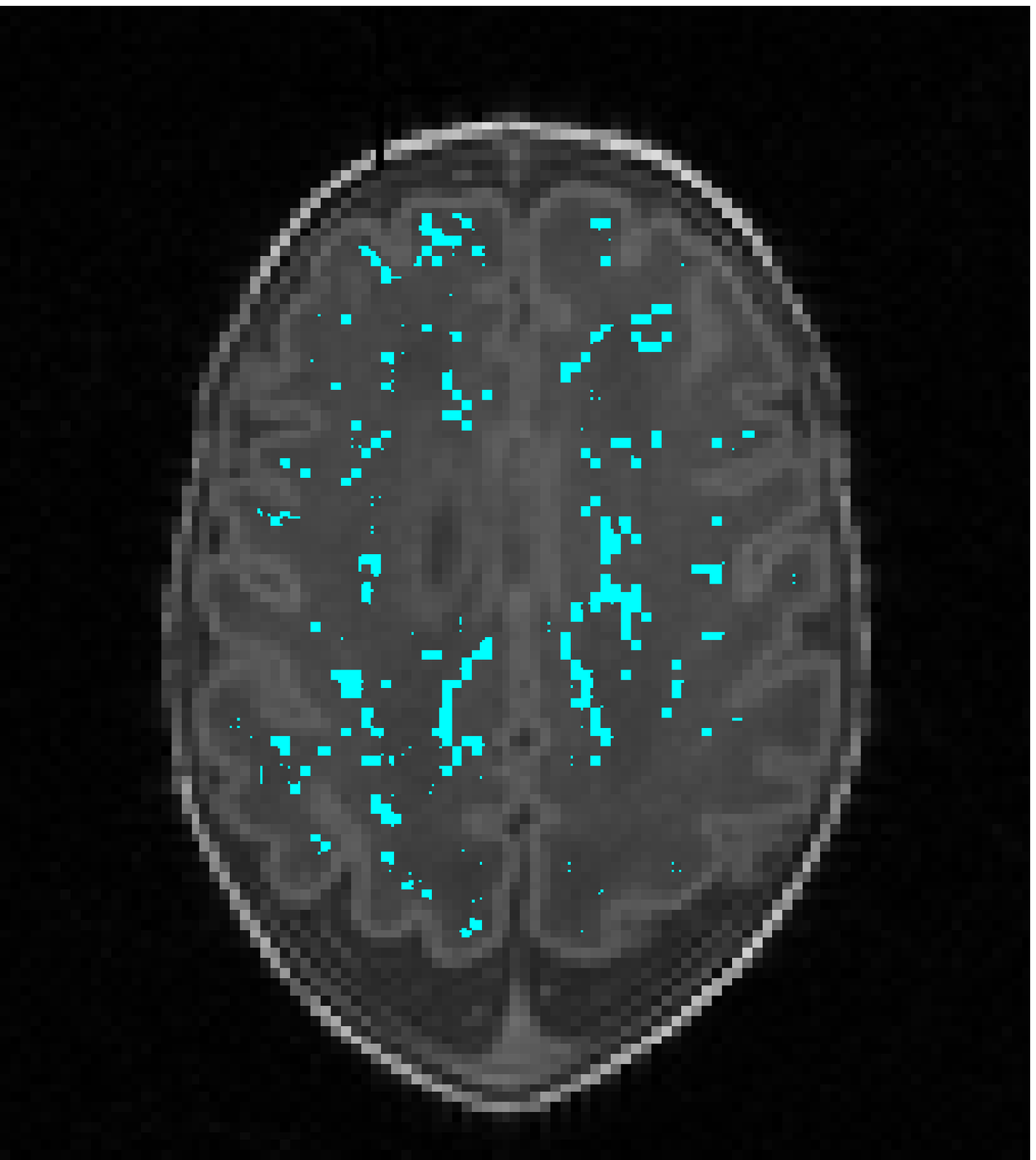}
    	\caption{$\mathcal{T}=0.055$}
    	\label{fig:Prof_42_2_055}
	\end{subfigure}
\caption{Comparison of Proposed Method and \cite{Cheng2015} using Slice 10.}
\label{fig:compare_42_2}
\end{figure}

\begin{figure}
\centering
	\begin{subfigure}[b]{0.15\textwidth}
    	\centering
    	\includegraphics[width=\textwidth]{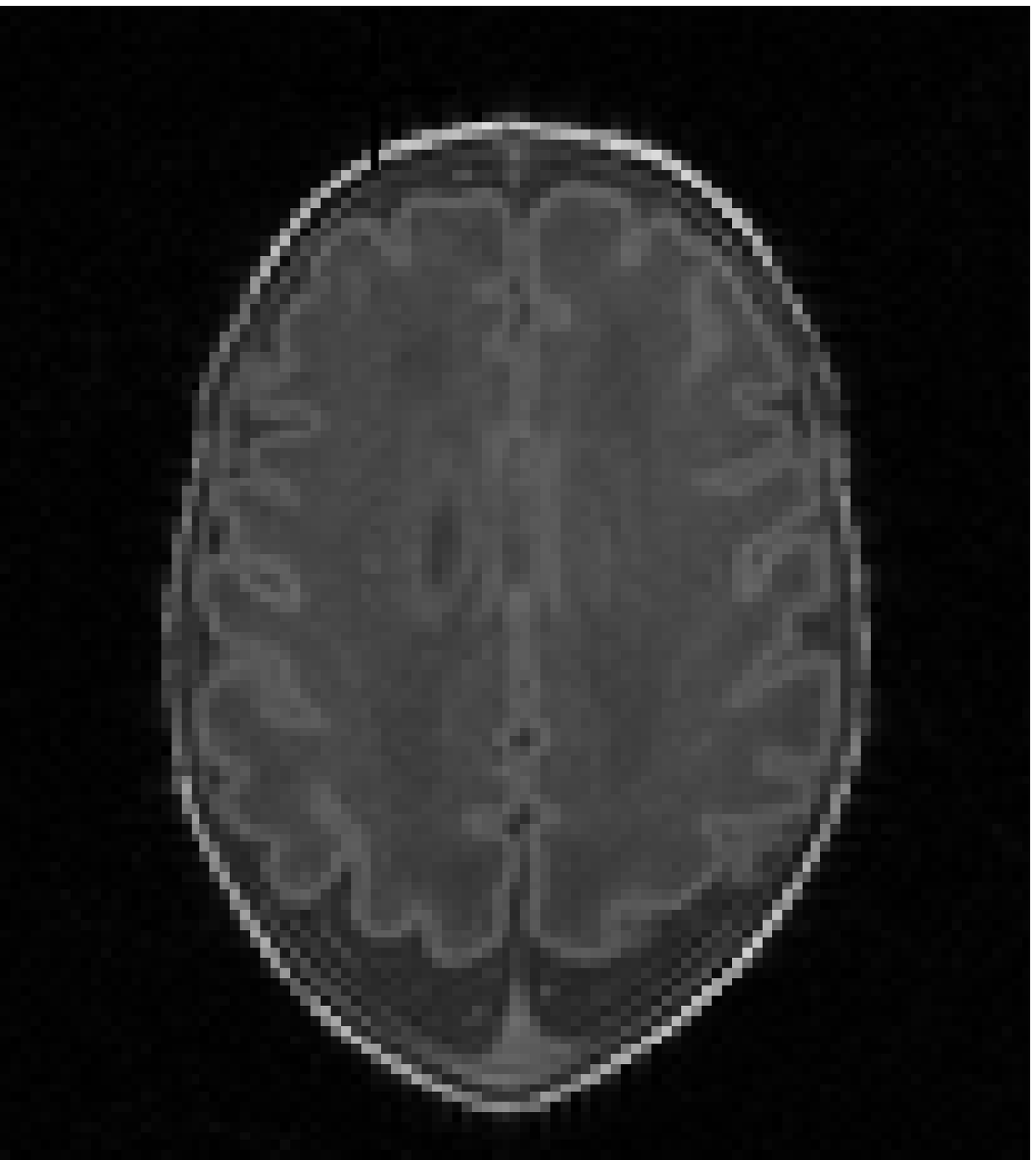}
    	\caption{Input Slice}
    	\label{fig:Slice_42_3}
	\end{subfigure}
	\begin{subfigure}[b]{0.15\textwidth}
    	\centering
    	\includegraphics[width=\textwidth]{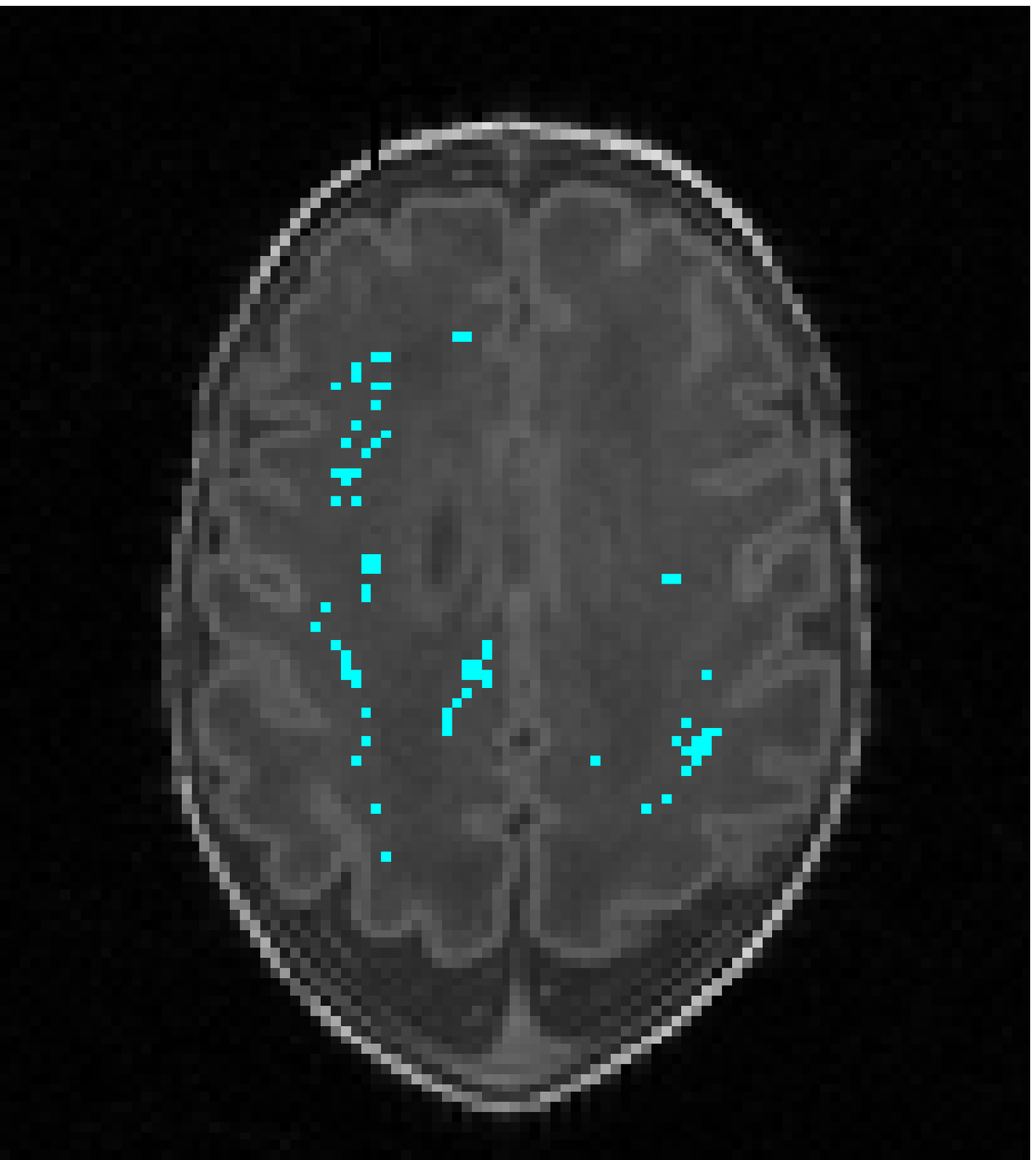}
    	\caption{Our Method}
    	\label{fig:Proposed_42_3}
	\end{subfigure}
	\begin{subfigure}[b]{0.15\textwidth}
    	\centering
    	\includegraphics[width=\textwidth]{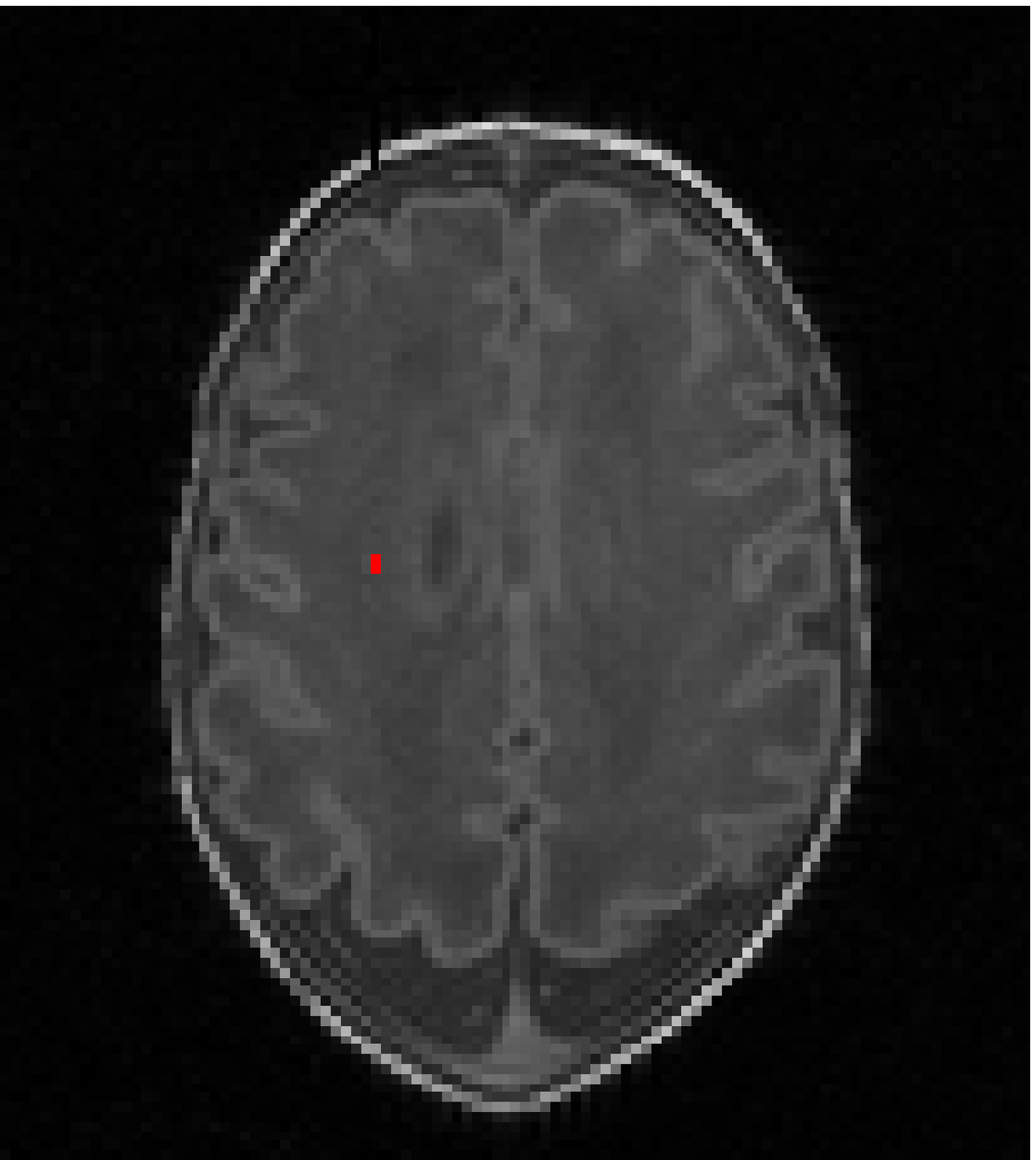}
    	\caption{Ground Truth}
    	\label{fig:Ground_42_3}
	\end{subfigure}
	\hfill
	\begin{subfigure}[b]{0.15\textwidth}
    	\centering
    	\includegraphics[width=\textwidth]{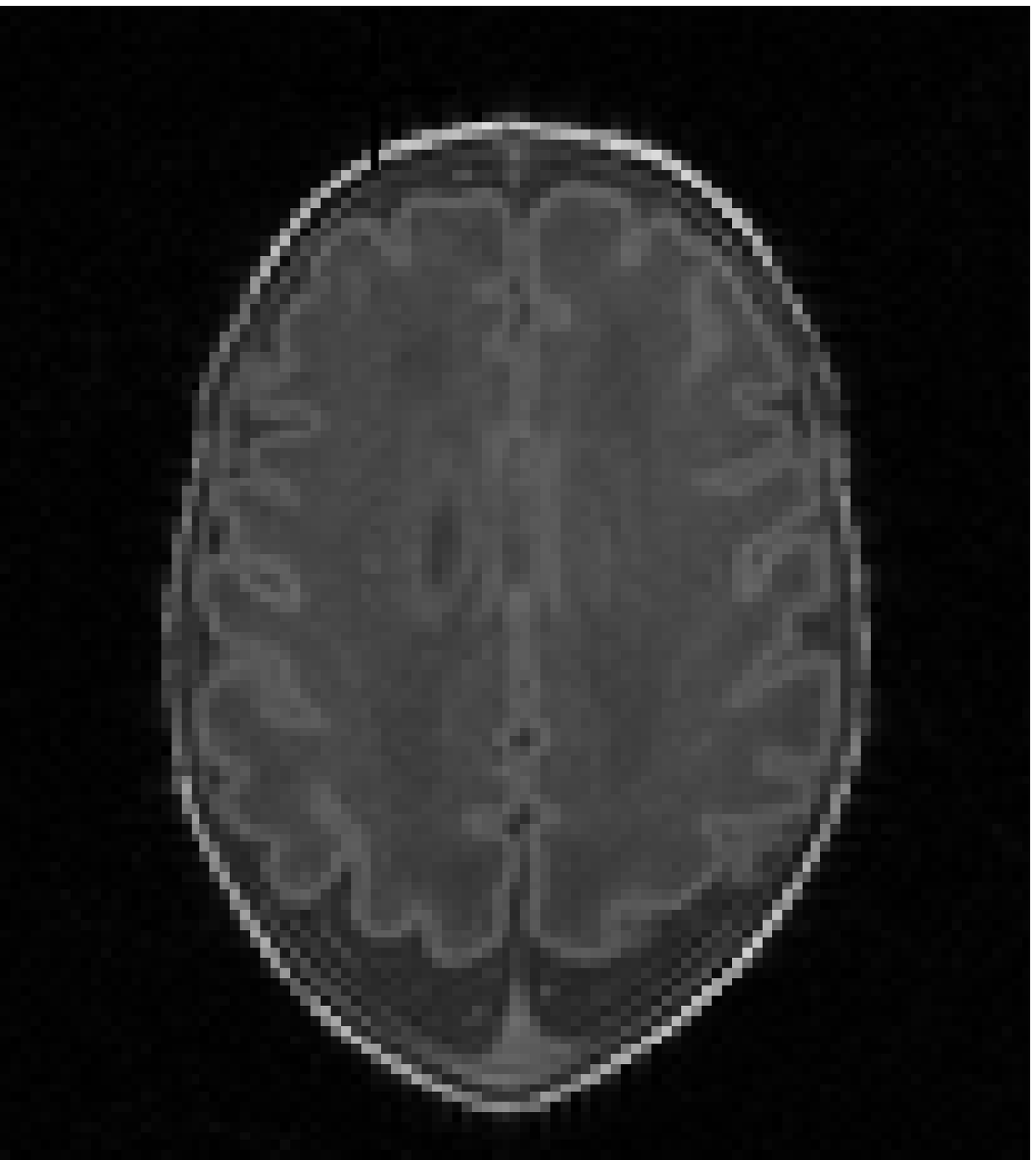}
    	\caption{$\mathcal{T}=0.005$ \cite{Cheng2015}}
    	\label{fig:Prof_42_3_005}
    \end{subfigure}
    \hfill
    \begin{subfigure}[b]{0.15\textwidth}
    	\centering
    	\includegraphics[width=\textwidth]{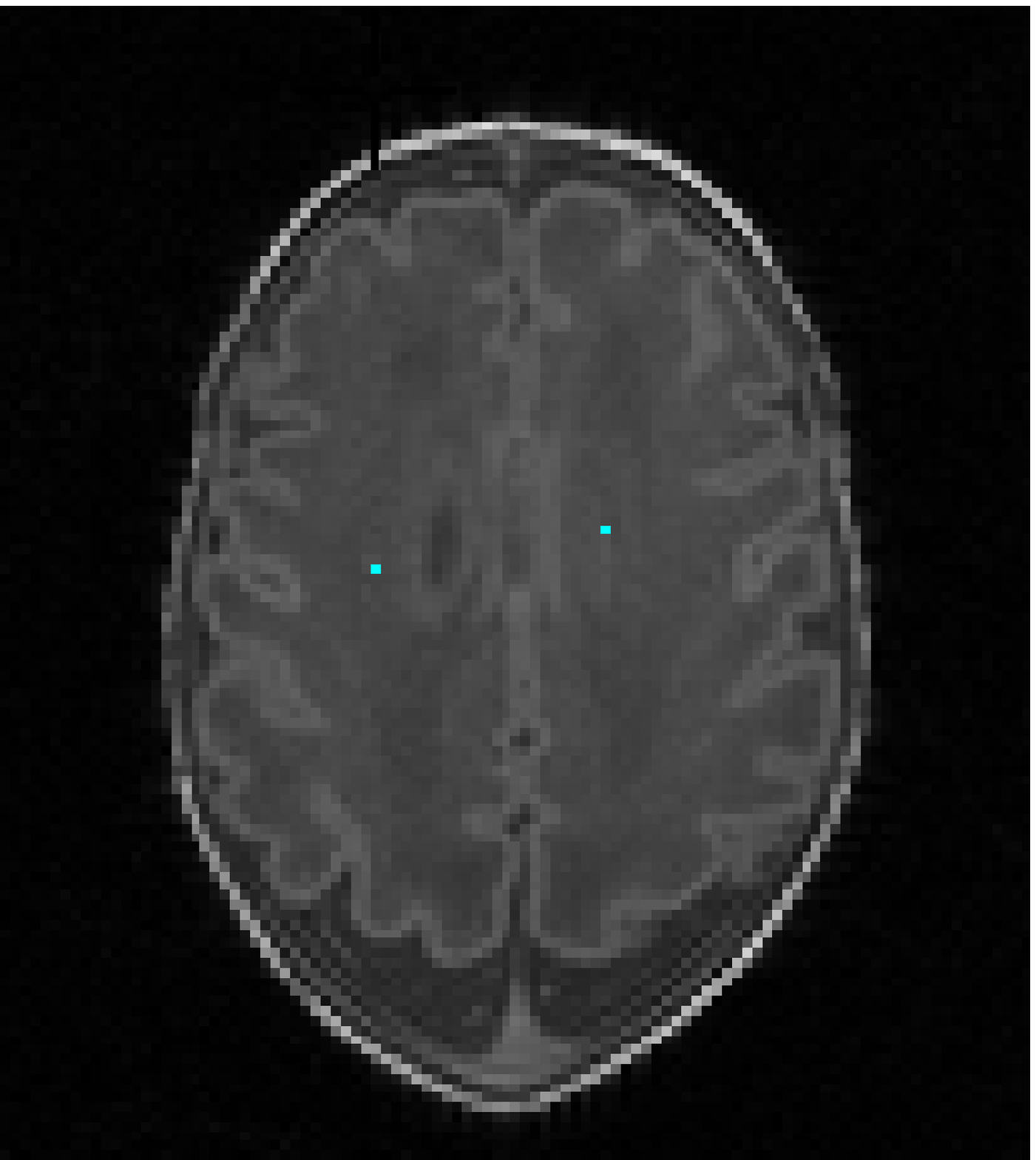}
    	\caption{$\mathcal{T}=0.015$}
    	\label{fig:Prof_42_3_015}
    \end{subfigure}
    \begin{subfigure}[b]{0.15\textwidth}
    	\centering
    	\includegraphics[width=\textwidth]{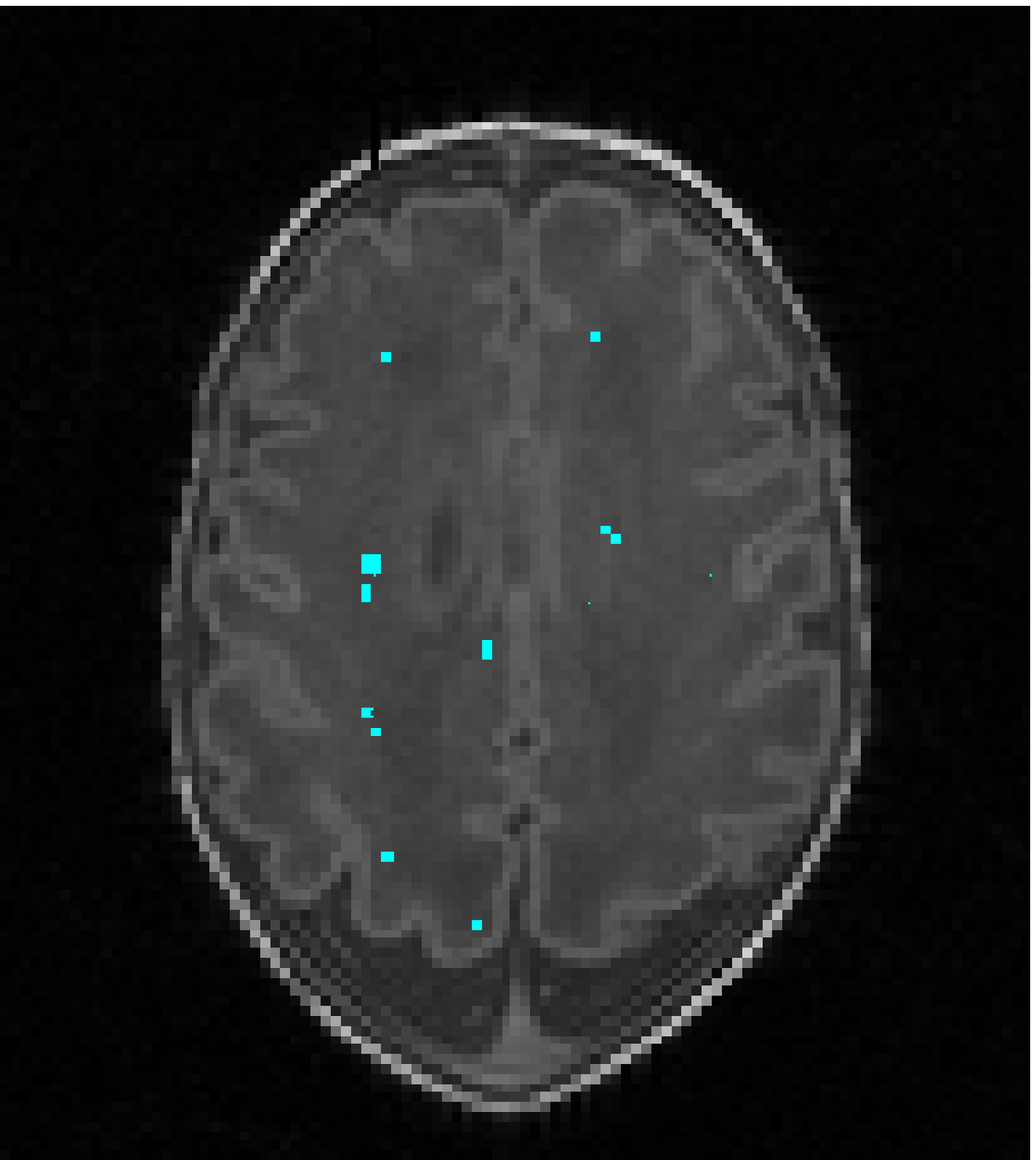}
    	\caption{$\mathcal{T}=0.025$}
    	\label{fig:Prof_42_3_025}
	\end{subfigure}
	\begin{subfigure}[b]{0.15\textwidth}
    	\centering
    	\includegraphics[width=\textwidth]{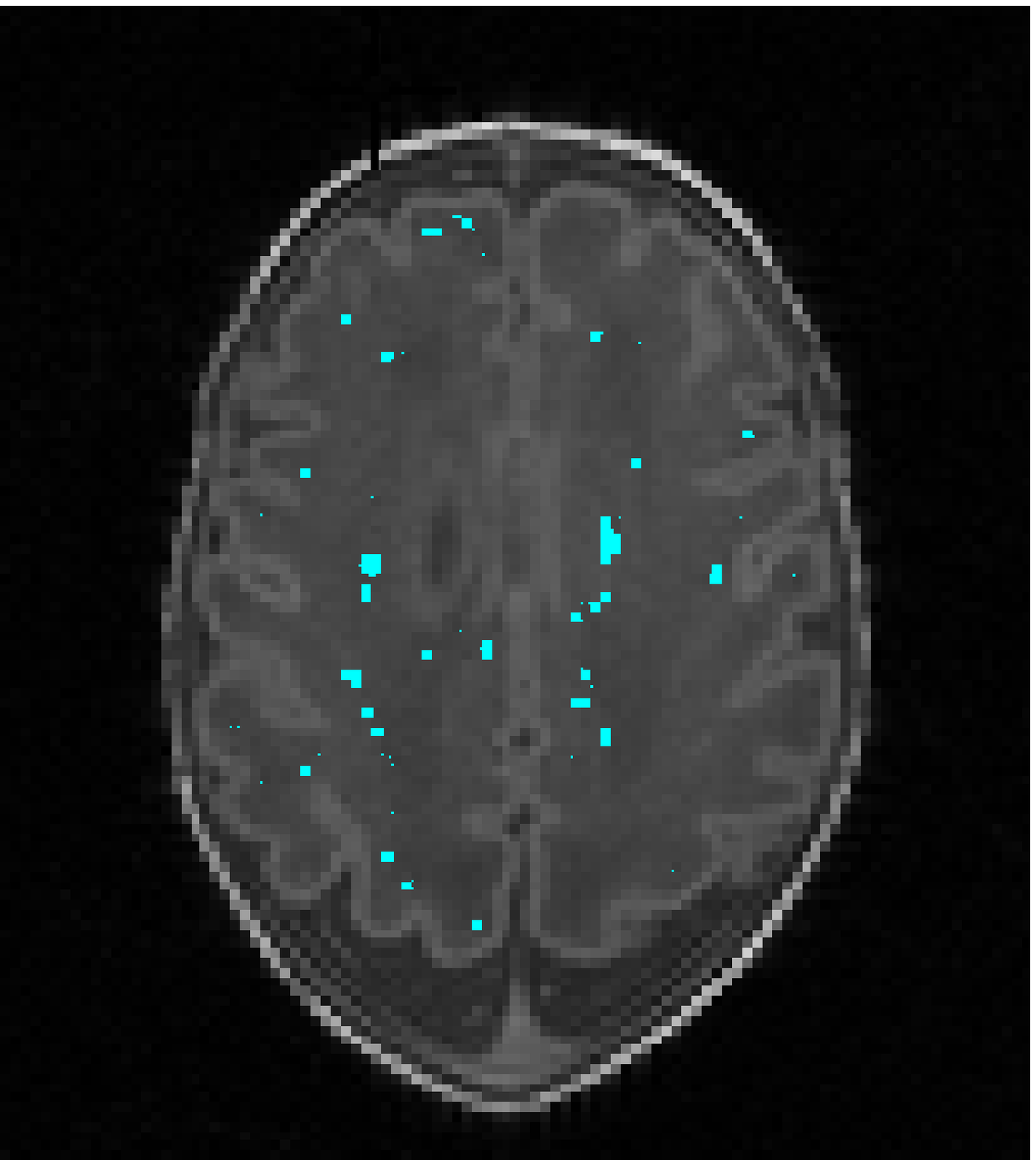}
    	\caption{$\mathcal{T}=0.035$}
    	\label{fig:Prof_42_3_035}
	\end{subfigure}
	\begin{subfigure}[b]{0.15\textwidth}
    	\centering
    	\includegraphics[width=\textwidth]{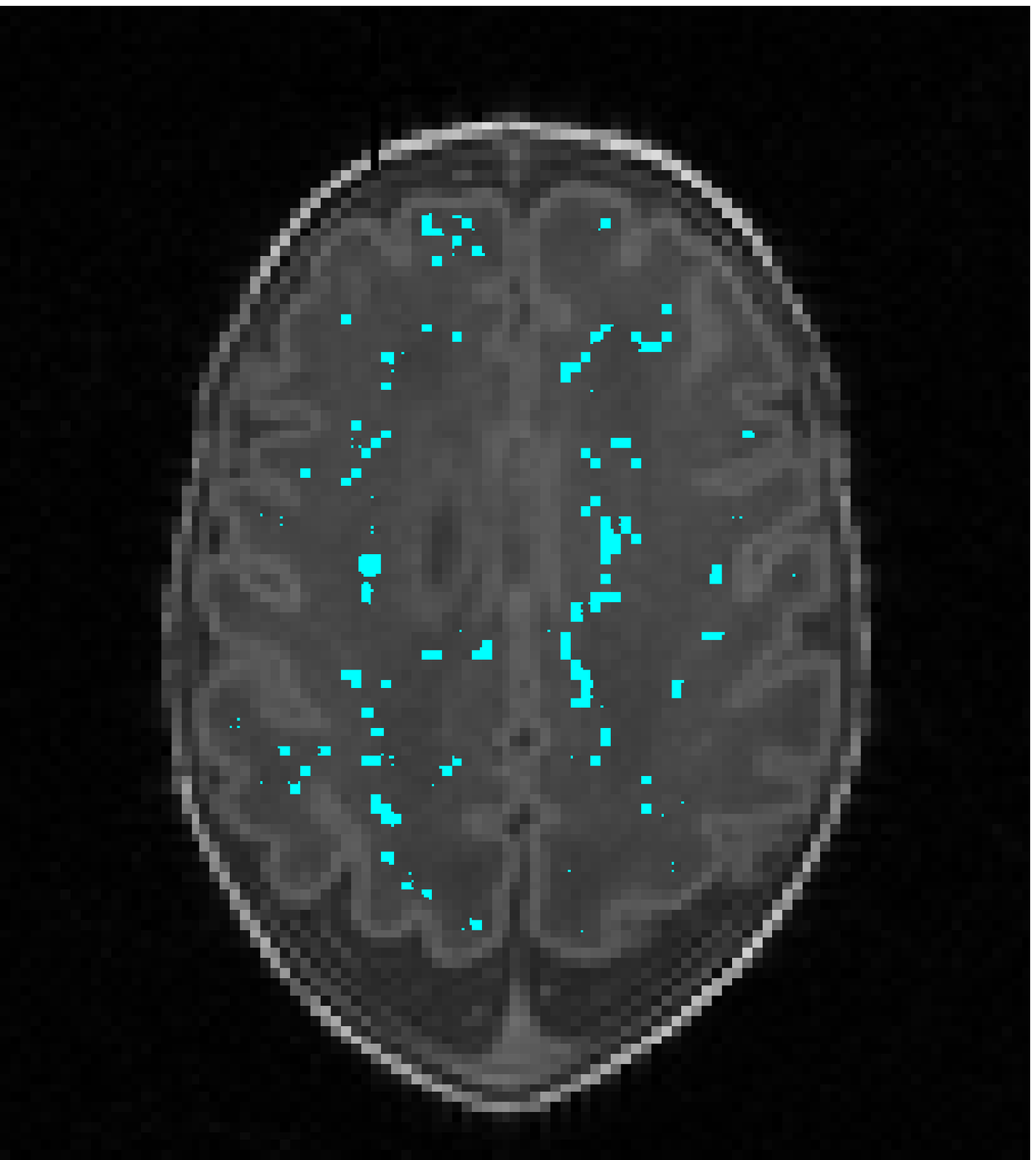}
    	\caption{$\mathcal{T}=0.045$}
    	\label{fig:Prof_42_3_045}
	\end{subfigure}
	\begin{subfigure}[b]{0.15\textwidth}
    	\centering
    	\includegraphics[width=\textwidth]{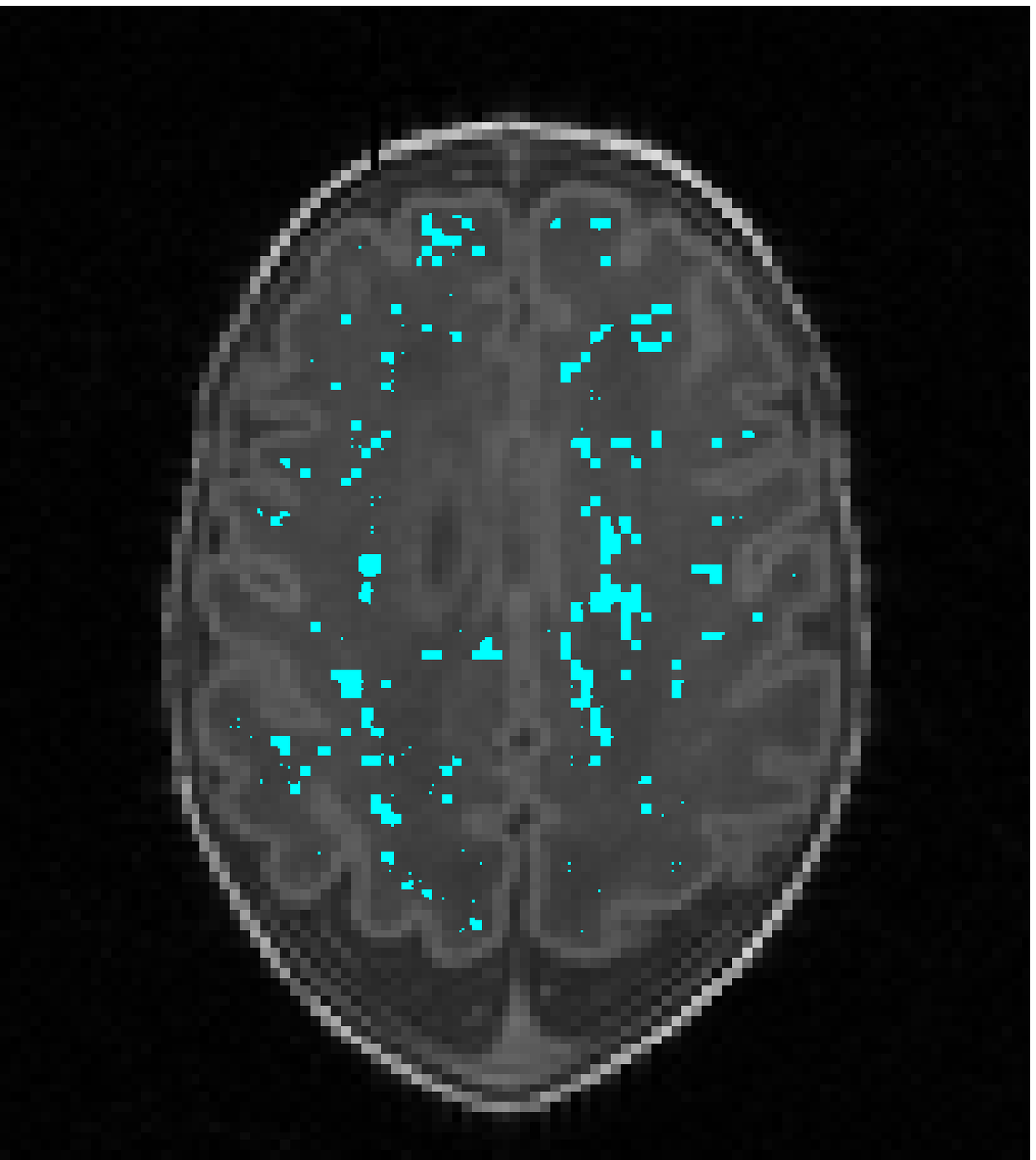}
    	\caption{$\mathcal{T}=0.055$}
    	\label{fig:Prof_42_3_055}
	\end{subfigure}
\caption{Comparison of Proposed Method and \cite{Cheng2015} using Slice 11.}
\label{fig:compare_42_3}
\end{figure}

Figs. \ref{fig:compare_24_5} through \ref{fig:compare_42_3} present a side-by-side comparison of the output from \cite{Cheng2015} and the proposed method for three representative slices belonging to the 2\textsuperscript{nd} and 3\textsuperscript{rd} subjects. It can be seen that many false positives detected in \cite{Cheng2015} for certain values of $\mathcal{T}$ are eliminated using our new method. Overall, Figs. \ref{fig:compare_24_5} through \ref{fig:compare_42_3} demonsrate how our method performs better than the method in \cite{Cheng2015}.

\section{Discussion}

From the qualitative analysis, specifically Figs. \ref{fig:compare_24_5} through \ref{fig:compare_42_3}, we observe the high dependence of the method in \cite{Cheng2015} on its parameter $\mathcal{T}$, whereas the method we propose in this paper gives consistent results. Another strength of our method is that, even when the ventricles do not appear in a particular slice, the WMI detection performance is not compromised, as can be seen in Fig. \ref{fig:compare_24_5}. Most importantly, our method does not perform segmentation of the entire WM region like most other lesion detection methods in general. In fact, many such methods, assume that an already clearly segmented WM region will be supplied as input, so that the method only focuses on WMI detection disregarding potential segmentation errors. In any case, false positives which  are obvious can be easily identified by the human expert when reviewing the computer assisted detection (first level filtering) results. 

From the quantitative analysis, we can see that the enforcement of size and distance constraints clearly increases the specificity, and decreases the sensitivity. Specificity increases as false positives WMI detections are eliminated by the enforcement of the constraints. Sensitivity decreases because the constraints erroneously eliminate some true positive WMI detections. Thus, this is an area which can be further developed in future work, to minimize such erroneous eliminations. We believe that additional criteria based on prior clinical knowledge is needed in order to selectively retain the true positive WMI detections while discarding the false positives. This is especially true for situations in which the true positive WMIs are difficult to perceive visually. On the other hand, a specificity of 100 with no constraint implies that our WM hyper-intensity detection approach is robust enough to detect all true positive WMIs. Among the two constraints, the distance constraint alone performs as well as its combination with the size constraint. This is because, as we move from the periphery of the brain towards the vicinity of the ventricles, the size of the connected components of detected WM hyper-intensities keeps decreasing, as can be seen in Fig. \ref{fig:hyper}. The periphery has larger connected components, like skull and cortical ribbon, whereas those nearer to the ventricles are more likely to be true WMI. Thus, in this respect, the distance criteria subsumes the size criteria for WMI detection.

Furthermore, it should be borne in mind that even a small increase in the number of true positive WMI detentions results in a large increase in the sensitivity score, and the opposite is true for false positives and specificity. The reason is that, in our preterm brain WMI detection based on the given ground truth, the total number of positives (WMIs) is very small compared to the total number of negatives (healthy brain tissue). Thus, the numerical values of sensitivity and specificity should be interpreted in light of the actual number of true positives and false positives while assessing the relative performance of the proposed method.

\subsection{Slice Thickness}

The inter-slice distance (thickness) in our test dataset is 1-mm and the magnetic strength of the MRI scans is 1.5 Tesla. It has been shown by Savlo et al. \cite{Savio2010} that if we vary the slice thickness between 1-mm and 3-mm, the texture features used to detect lesions remain visible. While very thin slices would reduce the signal-to-noise ratio (SNR), resulting in unreliable texture patterns. On the other hand, very thick slices would compromise the texture detail. %Seemingly, the range from 1 mm to 3 mm does not correspond to either of these cases, as established by Savlo et al. \cite{Savio2010}.

\subsection{Applicability of general assumptions regarding WM lesion characteristics}

Multiple sclerosis lesion detection by thresholding FLAIR images, and subsequent refinement of the threshold mask in order to differentiate lesion regions from normal tissue, has been performed in \cite{Cabezas2014}. The authors define a set of rules, like ``Lesions are mostly surrounded by WM voxels'' and ``Lesions should not be present between the ventricles,'' which are true in our test dataset as well. However, other rules like ``Lesions (targets) should be of a minimum size'' might not be true in our case. In their work, the authors set this ``minimum size'' as 10 voxels representing 30 $mm^3$, which approximately represent a cube with 3 mm edges. They argue that due to inherent noise and intensity inhomogeneities in MR images, some voxels with random high intensities may persist even after the preprocessing stage. Thus, to remove these small outliers, they discard all lesion regions that do not have the defined ``minimum size.'' In terms of preterm WMI detection, the size of preterm neonate brains is very small, the MRIs themselves are of very low resolution and have a very low contrast-to-noise ratio. Thus, it is inaccurate to discard random high intensities assuming they are noise, because true WMIs can span just a small subset of voxels in each slice, as shown in our expert-annotated ground truth dataset. This observation is verified by our experimental results presented in Table \ref{tab:results_md}, which shows that a minimum size cannot be imposed on preterm neonate WMI. It can be seen that by increasing the ``minimum'' size (in terms of number of voxels), there is an increasing number of missing detections, which lowers the sensitivity score. By the time we increase the minimum size to 250 voxels, all slices having correct targets detected in an unconstrained environment would have missed all correct detections. For this reason, the ``size criteria'' for eliminating false-positive in our proposed method does not impose a ``minimum'' acceptable WMI size. The specificity score increases as a result of increasing the minimum allowed lesion size, because fewer lesions are detected as such which also decreases the chance of false positive detections, as seen in Table \ref{tab:results_md}. However, clearly, this cannot be regarded as an advantage at the cost of severely increasing missed true detections.

\section{Conclusion}

We presented a robust and efficient method for the detection of white matter injury in preterm neonate brain MRI scans. We introduced a fast, automatic, unsupervised and atlas-free WMI detection approach, which avoids the WM segmentation step. We apply GA-based image patch classification to sample WM intensities and subsequently eliminate false positives using size-based and distance-based criteria. We also use adjacent slice validation. The proposed method is an effective technique for WMI detection in preterm neonates. Experimental results show that our method outperforms related work. However, there exist challenging MR scans, where the WMI cannot be identified purely relying on visual cues, leading to the failure of all methods. This opens up on-going research to improve existing algorithms by incorporating demographic and other clinical information of patients to identify candidate targets. Our future work will also include monitoring dynamic change of WMI over time and discover the impact of preterm neonate WMI on developmental deficits.

\section{Compliance with Ethical Standards}

Informed consent: Informed consent was obtained from all individual participants included in the study.

Ethical approval: All procedures performed in studies involving human participants were in accordance with the ethical standards of the institutional and/or national research committee and with the 1964 Helsinki declaration and its later amendments or comparable ethical standards.

Conflict of Interest: The authors declare that they have no conflict of interest.

The images provided were from a research project for which parents provided consent.

We followed the ``Standard Protocol Approvals, Registration, and Patient Consents'' at the BC Children's Hospital in Vancouver. A written informed consent from the legal guardian of each participating neonate was obtained. This study was reviewed and approved by the Clinical Research Ethics Board at the University of British Columbia and BC Children's and Women's Hospitals.

\begin{acknowledgements}

Financial support from CIHR, NeuroDevNet, Alberta Innovates (iCORE) Research Chair program, and NSERC in conducting this research is gratefully acknowledged.

\end{acknowledgements}

\section*{Biographies}

\textbf{Subhayan Mukherjee} did his Master of Technology (by Research) on computer vision from National Institute of Technology Karnataka, India. He worked on HDR imaging (patent pending) in Dolby Labs, Sunnyvale. His PhD work on visual saliency was published in IEEE ICIP 2016.

\textbf{Irene Cheng} is the Scientific Director of the Multimedia Research Group, and Director of the Master with Specialization in Multimedia Program, at University of Alberta, Canada. Her research interest focuses on perceptually motivated multimedia computing.

\textbf{Steven Miller} is Head of the Division of Neurology and a Senior Scientist at The Hospital for Sick Children, Toronto. He is a Professor in the Department of Paediatrics at the University of Toronto and holds the Bloorview Chair in Paediatric Neuroscience.

\textbf{Jessie Guo} is a Research Associate for brain image analysis in high-risk neonatal populations. Her PhD (Biomedical Engineering) is from Western University, Canada. She conducted her postdoctoral training at The Hospital for Sick Children, Toronto.

\textbf{Vann Chau} is a Paediatric Neurologist at The Hospital for Sick Children. He is the Principle Investigator of the Paediatric Cardiac and Neurological Registry study. He investigates brain injury and its effect on long term outcome in premature infants.

\textbf{Anup Basu} is fellow of American Neurological Association and AITF Industrial Chair in Multimedia. His Ph.D (Computing Science) is from University of Maryland, College Park and MS from Biostatistics department, Strong Memorial Hospital, Rochester.

\bibliographystyle{plain}
\bibliography{journal}

\end{document}